\documentclass[sigconf]{acmart}

\usepackage{array}
\usepackage{balance}
\usepackage{multirow}
\usepackage{subcaption}
\usepackage{times}
\usepackage{tikz}

\usepackage{graphicx}%
\usepackage{multirow}%
\usepackage{amsmath,amsfonts}%
\usepackage{amsthm}%
\usepackage{mathrsfs}%
\usepackage[title]{appendix}%
\usepackage{xcolor}%
\usepackage{textcomp}%
\usepackage{manyfoot}%
\usepackage{booktabs}%
\usepackage{algorithm}%
\usepackage{algorithmicx}%
\usepackage{algpseudocode}%
\usepackage{listings}%

\newfloat{listing}{tb}{lst}{}
\floatname{listing}{Listing}

\AtBeginDocument{%
  }

\copyrightyear{2025} 
\acmYear{2025} 
\setcopyright{cc}
\setcctype{by}
\acmConference[KDD '25]{Proceedings of the 31st ACM SIGKDD Conference on Knowledge Discovery and Data Mining V.2}{August 3--7, 2025}{Toronto, ON, Canada}
\acmBooktitle{Proceedings of the 31st ACM SIGKDD Conference on Knowledge Discovery and Data Mining V.2 (KDD '25), August 3--7, 2025, Toronto, ON, Canada}
\acmDOI{10.1145/3711896.3736818}
\acmISBN{979-8-4007-1454-2/2025/08}

\begin{document}

\title[``Oh LLM, I'm Asking Thee, Please Give Me a Decision Tree'']{``Oh LLM, I'm Asking Thee, Please Give Me a Decision Tree'': Zero-Shot Decision Tree Induction and Embedding with Large Language Models}

\author{Ricardo Knauer}
\email{ricardo.knauer@htw-berlin.de}
\affiliation{%
  \institution{KI-Werkstatt, University of Applied Sciences Berlin}
  \city{Berlin}
  \country{Germany}
}

\author{Mario Koddenbrock}
\email{mario.koddenbrock@htw-berlin.de}
\affiliation{%
  \institution{KI-Werkstatt, University of Applied Sciences Berlin}
  \city{Berlin}
  \country{Germany}
}

\author{Raphael Wallsberger}
\email{raphael.wallsberger@htw-berlin.de}
\affiliation{%
  \institution{KI-Werkstatt, University of Applied Sciences Berlin}
  \city{Berlin}
  \country{Germany}
}

\author{Nicholas M. Brisson}
\email{nicholas.brisson@charite.de}
\affiliation{%
  \institution{Julius Wolff Institute, Berlin Institute of Health at Charité - Universitätsmedizin Berlin}
  \city{Berlin}
  \country{Germany}
}

\author{Georg N. Duda}
\email{georg.duda@charite.de}
\affiliation{%
  \institution{Julius Wolff Institute, Berlin Institute of Health at Charité - Universitätsmedizin Berlin}
  \city{Berlin}
  \country{Germany}
}

\author{Deborah Falla}
\email{d.falla@bham.ac.uk}
\affiliation{%
  \institution{Centre of Precision Rehabilitation for Spinal Pain, School of Sport, Exercise and Rehabilitation Sciences at University of Birmingham}
  \city{Birmingham}
  \country{United Kingdom}
}

\author{David W. Evans}
\email{d.w.evans@bham.ac.uk}
\affiliation{%
  \institution{Centre of Precision Rehabilitation for Spinal Pain, School of Sport, Exercise and Rehabilitation Sciences at University of Birmingham}
  \city{Birmingham}
  \country{United Kingdom}
}

\author{Erik Rodner}
\email{erik.rodner@htw-berlin.de}
\affiliation{%
  \institution{KI-Werkstatt, University of Applied Sciences Berlin}
  \city{Berlin}
  \country{Germany}
}

\renewcommand{\shortauthors}{Knauer et al.}

\begin{abstract}
Large language models (LLMs) provide powerful means to leverage prior knowledge for predictive modeling when data is limited. In this work, we demonstrate how LLMs can use their compressed world knowledge to generate intrinsically interpretable machine learning models, \textit{i.e.}, decision trees, \textit{without any training data}. We find that these zero-shot decision trees can even surpass data-driven trees on some small-sized tabular datasets and that embeddings derived from these trees perform better than data-driven tree-based embeddings on average. Our decision tree induction and embedding approaches can therefore serve as new knowledge-driven baselines for data-driven machine learning methods in the low-data regime. Furthermore, they offer ways to harness the rich world knowledge within LLMs for tabular machine learning tasks. Our code and results are available at \texttt{https://github.com/ml-lab-htw/\allowbreak llm-trees}.
\end{abstract}



\received{23 January 2025}
\received[accepted]{14 May 2025}

\maketitle

\begingroup\small\noindent\raggedright\textbf{KDD Availability Link:}\\
The source code of this paper has been made publicly available at \texttt{https:\allowbreak //doi.org/10.5281/zenodo.15496002}.
\endgroup

\section{Introduction}

Machine learning algorithms are data-hungry \citep{banko2001scaling, halevy2009unreasonable, knauer2024squeezing}, including intrinsically interpretable models like decision trees \citep{van2014modern}. In many domains, though, large- or medium-sized datasets are not easily available. In the healthcare sector, for instance, data is often limited for diagnostic or prognostic modeling when medical conditions are rare or dropout rates at follow-up assessments are high, respectively \citep{moons2015transparent, moons2019probast, steyerberg2019clinical}. In such a low-data regime, training predictive models from scratch is challenging and it may be necessary to make extensive use of prior knowledge for valid and reliable inferences \citep{harrell2015regression, heinze2018variable, knauer2023cost,wallsberger2024explainable}. In recent years, advances in natural language processing and computing have granted practitioners and researchers access to the condensed world knowledge from much of the web in the form of pretrained deep learning models, \textit{i.e.}, large language models (LLMs) \citep{anthropic2024claude, google2024gemini, openai2024gpt4o, openai2024reasoning}. Although LLMs provide powerful means to leverage prior knowledge for predictive modeling when data is scarce, interpreting their decisions remains an open challenge \citep{longo2024explainable}. Additionally, state-of-the-art LLMs are proprietary \citep{chiang2024chatbotarena} and therefore cannot be readily used with sensitive data \citep{anthropic2024sensitive, google2024sensitive, openai2024sensitive}. This limits their applicability for domains where privacy is critical such as the clinical setting.

In this work, our contributions are as follows:
\begin{enumerate}
\item We show how \textbf{state-of-the-art LLMs can be used for building intrinsically interpretable machine learning models, \textit{i.e.}, decision trees, without access to pretrained model weights and \textit{without any training data}} (Sect.~\ref{sec:induction}). The zero-shot setting naturally preserves the data privacy and thus broadens the applicability of LLMs for predictive modeling across industry verticals.

\item We demonstrate how our \textbf{zero-shot decision trees can also serve as feature representations for downstream models} (Sect.~\ref{sec:embedding}). This provides a powerful way to leverage the rich world knowledge within LLMs for tabular machine learning tasks.

\item We offer a \textbf{systematic comparison of our decision tree induction and embedding approaches with state-of-the-art machine learning methods} on 13 public and 2 private tabular classification datasets in the low-data regime. We show that our knowledge-driven trees achieve a better performance than data-driven trees on 27\% of the datasets and that our zero-shot representations are statistically significantly better than data-driven tree-based embeddings (Sect.~\ref{sec:results}). Therefore, we argue that our zero-shot induction and embedding approaches can serve as new knowledge-driven baselines that data-driven machine learning methods should surpass to show their efficacy. The code and results for our zero-shot induction and embedding pipelines are openly available at \texttt{https:\allowbreak //github.com/ml-lab-htw/llm-trees}.

\end{enumerate}

\section{Related Work}

Transferring knowledge from one task with more data to another task with less data, \textit{i.e.}, transfer learning, has a long history \citep{elhoseiny2013write,pan2009survey}. However, the compressed world knowledge within LLMs has opened up new dimensions to augment data-driven machine learning methods with prior knowledge, particularly through embeddings, fine-tuning, and in-context learning. Embedding approaches extract feature representations from LLMs and apply them as inputs to downstream models \citep{peters2018contextualized}. Fine-tuning approaches update the LLM parameters for downstream tasks \citep{howard2018universal}. In-context learning is an emergent ability in LLMs and enables them to perform new tasks without any additional parameter training or updates. Instead, training examples are provided to the LLM via prompts \citep{brown2020language}. This allows for rapid prototyping \citep{hollmann2023tabpfn,sun2022black} and facilitates the integration of additional knowledge by simply changing the prompt \citep{kojima2022large,wei2022chain}. Although the exact working mechanism of in-context learning is not yet fully studied \citep{fu2024transformers,shen2024pretrained,xie2022explanation}, it has been shown that deep learning architectures that power modern LLMs, \textit{i.e.}, transformers and state space models, can in-context learn (sparse) linear functions, neural networks, and decision trees, among others \citep{garg2022can,grazzi2024mamba}. With $\geq$ 1 training examples, decision rules or trees can thus be extracted not only from
machine learning models more generally using explainable artificial intelligence (XAI) approaches \citep{molnar2022,ribeiro2016should,ribeiro2018anchors}, but also from LLMs more specifically using in-context learning \citep{han2024large,li2023tree,wang2025largelanguagemodels}.

In contrast to settings with $\geq$ 1 training examples, deriving decision rules or trees from LLMs \textit{without any training data} is an underexplored research area, but critical in settings where sensitive data cannot be readily shared. Most related to our work, \citet{nam2024featuregeneration} recently leveraged LLMs to generate features for tabular data. First, they used an LLM to propose a new feature name and to find a rule for generating the new feature values using only a task description and the feature names, but without any training examples. Second, they augmented the data with the new feature, trained a decision tree on the new data, and obtained a validation score. Finally, they used the LLM to iteratively improve the feature generation rule given a task description and the feature names as well as previous rules, trained decision trees, and validation scores. However, the focus of our research is not on feature generation, which can be detrimental when data is scarce \citep{harrell2015regression,heinze2018variable,knauer2023cost,wallsberger2024explainable}. Rather, we present the first approach to apply state-of-the-art LLMs for zero-shot model generation using in-context learning, \textit{i.e.}, we show how LLMs can build intrinsically interpretable trees without access to pretrained model weights and without any training data (Sect.~\ref{sec:induction}). Furthermore, we demonstrate how our zero-shot trees can also serve as effective feature representations, \textit{i.e.}, embeddings, for downstream models (Sect.~\ref{sec:embedding}).

\section{LLMs as Zero-Shot Model Generators}

LLMs are pretrained on massive text corpora and datasets crawled from the web and thus possess an extensive repository of prior knowledge about the world \citep{bordt2024elephants1,bordt2024elephants2}. In the following, we show that this prior knowledge can be used to guide data-driven predictive model development, especially in the low-data regime. When data is limited, the rich world knowledge within LLMs may effectively compensate for data scarcity and increase the predictive validity and reliability compared to purely data-driven approaches \citep{harrell2015regression, heinze2018variable, knauer2023cost,wallsberger2024explainable}. Additionally, the integration of knowledge- and data-driven methods may be particularly valuable in fields like healthcare that require both a deep contextual understanding and empirical insights for effective decision-making \citep{moons2015transparent,moons2019probast,steyerberg2019clinical}.

\subsection{Zero-Shot Decision Tree Induction} \label{sec:induction}

In this section, we present how state-of-the-art LLMs can be used to generate decision trees even \textit{without any training data}. We therefore go beyond prior research that has employed LLMs to generate new features for machine learning models \citep{nam2024featuregeneration}. Our approach does not only allow us to leverage the prior knowledge within LLMs for predictive modeling, but also to extract interpretable decision rules from LLMs and to naturally respect the data privacy. This broadens the applicability of LLMs for domains where data is frequently scarce, interpretability is desirable, or data protection is mandatory, such as in the healthcare sector \citep{longo2024explainable,moons2015transparent,moons2019probast,steyerberg2019clinical}.

Listing~\ref{lst:prompt_template} shows our prompting template for building a zero-shot decision tree $T$. We start each prompt by presenting some background context, followed by the actual task description. The prediction target $p\in\mathcal{P}$ and the tree's maximum depth $d\in\mathbb{N}^+$ need to be set depending on the task. Most importantly, we employ in-context learning to provide the LLM with an output indicator, \textit{i.e.}, we encourage the LLM to output the decision tree in a textual format.\footnote{Our example in Listing~\ref{lst:prompt_template} is taken from \texttt{https://scikit-learn.org/stable/\allowbreak modules/generated/sklearn.tree.export\_text.html}.} Finally, we pass the $k\in\mathbb{N}^+$ feature names $f=[f_1, ..., f_k]\in\mathcal{F}$ to the LLM including, if applicable, the categorical values or measurement units in brackets. Since state-of-the-art LLMs have seen massive text corpora and datasets from the web during pretraining \citep{bordt2024elephants1,bordt2024elephants2}, they can leverage their condensed world knowledge to transform feature names into decision trees, given that the feature names are meaningful (Sect.~\ref{sec:datasets}). This allows the LLM to derive a mapping $\phi(p, f_1, ..., f_k, d)\mapsto T$ (Sect.~\ref{sec:results_induction}).

\begin{listing}[t]
\caption{Prompt template for our zero-shot tree
induction}%
\label{lst:prompt_template}%
\begin{lstlisting}[language=sh]
[User] 
I want you to induce a decision tree classifier based on features. I first give an example below. Then, I provide you with new features and want you to build a decision tree with a maximum depth of d using the most important features. 
The tree should classify p.

Features:
sepal length (cm), sepal width (cm), petal length (cm), petal width (cm)

Decision tree:
|- petal width (cm) <= 0.80
| |- class: setosa
|- petal width (cm) > 0.80
| |- petal width (cm) <= 1.75
| | |- class: versicolor
| |- petal width (cm) > 1.75
| | |- class: virginica

Features:
f_1, ..., f_k

Decision tree:
\end{lstlisting}
\end{listing}

\subsection{Zero-Shot Decision Tree Embedding} \label{sec:embedding}

Zero-shot decision trees encode rich structural dependencies between the input features and prediction target. In this section, we show how our knowledge-driven trees can be used as feature representations, \textit{i.e.}, embeddings, for downstream models. This allows downstream models such as artificial neural networks to reap the benefits of tree-based methods, which excel at selecting important features, handling nominal features, and capturing data irregularities, among others \citep{borisov_deeptlf_2023,grinsztajn2022tree,mcelfresh2023neural,moosmann2006fast}. While prior work has focused on generating tree-based embeddings in a data-driven way \citep{borisov_deeptlf_2023,moosmann2006fast}, our approach allows us to extract task-specific embeddings from a pretrained LLM without any training data and augment data-driven machine learning models with the LLM's compressed knowledge about the world.

\subsubsection{Knowledge Distillation}
Decision tree ensembles achieve state-of-the-art results on tabular data \citep{ye2024closerlook}, so we distill the knowledge from an LLM into $\geq$ 1 zero-shot decision trees for our embedding approach. Since employing similar trees within a single representation would not add much information, we leverage the LLM's inherent stochasticity (set by the temperature) and further increase its degrees of freedom by letting it determine the maximum tree depths on its own. Our mapping $\phi$ becomes a sampling from a probability distribution $T \sim \phi(p, f_1, ..., f_k)$. The sampling is followed by a transformation to an embedding that can be easily processed by downstream models such as artificial neural networks. 

\subsubsection{Embedding Transformation}
Let $\mathcal{T}_n$ represent the set of decision trees with $n$ inner nodes and $\mathbb{X}$ the feature space. Then the mapping $\chi_n$ of the truth values of the inner nodes to a binary vector of length $n$ can be expressed as $\chi_{n}: \mathcal{T}_n \times \mathbb{X} \to \{0,1\}^n$. The entries of the resulting binary vector are $1$ if the condition of the corresponding inner node is satisfied for a sample $x\in \mathbb{X}$, and $0$ otherwise \citep{borisov_deeptlf_2023}. Figure~\ref{fig:embedding_mapping} shows this mapping for the \textit{iris} data decision tree from Listing~\ref{lst:prompt_template} and a sample with a \textit{petal width} of 1.70. To embed a decision forest $(T_1, \ldots, T_m) \in \mathcal{T}_{n_1} \times ... \times \mathcal{T}_{n_m}$ consisting of $m$ decision trees  with $n_1, \ldots, n_m$ inner nodes, we define the mapping $\chi$ as follows:

\begin{displaymath}
\chi: \mathcal{T}_{n_1} \times \ldots \times \mathcal{T}_{n_m} \times \mathbb{X} \to \{0, 1\}^{n_1 + \cdots + n_m}.
\end{displaymath}

This mapping concatenates the individual binary vectors $\chi_{n_1}(T_1, x),$ $\ldots, \chi_{n_m}(T_m, x)$ obtained from each zero-shot tree to form our final task-specific embedding, which can either extend or replace the original feature vector. Such a binary encoding can enhance the computational efficiency compared to higher-precision encodings, which is particularly beneficial in resource-constrained environments such as edge devices, and is indeed a very natural representation for many optimization methods \citep{eiben2015introduction,ryan2018handbook}.

\begin{figure}[t]
\centering
\begin{tikzpicture}[
    edge from parent/.style={draw, -{latex}},
    sibling distance=10em,
    level distance=5em,
    every node/.style={shape=rectangle, rounded corners, draw, align=center}
    ]
    \node {petal width $\leq 0.80$}
        child { node {setosa} }
        child { node {petal width $\leq 1.75$}
            child { node {versicolor} }
            child { node {virginica} }
        };

\draw[->, thick] (3.4, -1.5) to[out=20, in=160] (4.8, -1.5);
\node at (5.5, -1.5)[draw=none] {\large \(\begin{pmatrix} 0 \\ 1 \end{pmatrix}\)};
\node at (4.2, -1)[draw=none] {\large $\chi_2(T, 1.70)$};
\node at (0, -1)[draw=none] {\large $T$};
\node at (1.4, -0.7)[draw=none] {\large $False$};
\node at (2.9, -2.3)[draw=none] {\large $True$};
\end{tikzpicture}
\caption{Embedding transformation $\chi_2$ using the \textit{iris} data decision tree from Listing~\ref{lst:prompt_template} and a sample with a \textit{petal width} of 1.70. The 2 inner nodes of the structured tree $T$ are mapped to a binary embedding vector of dimension 2. Thus, information from the leaf nodes is indirectly embedded.}
\Description{Decision tree to embedding transformation.}
\label{fig:embedding_mapping}
\end{figure}

\subsection{Output Formatting} \label{sec:formatting}

Our LLM-based zero-shot decision trees are generated in a textual format, which makes it very easy for domain experts to interpret them. To efficiently use our induction and embedding approaches within machine learning pipelines, though, we need to convert our generated trees into programs that yield both predictions (Sect.~\ref{sec:induction}) and binary embedding vectors (Sect.~\ref{sec:embedding}). To this end, we use the LLM to not only generate zero-shot decision trees, but also to reformat the textual decision tree output into a structured function.

Concretely, we prompt the LLM to generate a Python function that matches the textual description of the decision tree, accepts a sample, and returns a numerical prediction (Sect.~\ref{sec:induction}) and a list representing the truth values of the tree's inner nodes (Sect.~\ref{sec:embedding}). We also provide the LLM with an example of what the function should look like. We thus employ a two-step process that decouples the decision tree generation from the function format adherence \citep{tam2024letspeakfreely}. This way, the zero-shot trees are not only easily interpretable by domain experts (in their textual format), but also easily processable within downstream pipelines (in their function format).

In the next section, we thoroughly evaluate our zero-shot decision tree induction and embedding approaches against state-of-the-art machine learning methods.

\section{Experimental Setup}

\begin{table*}[t]
    \caption{Median test F1-score at 67\%/33\% train/test splits for our LLM-based zero-shot decision tree induction approach compared to the machine learning baselines. The best score among the zero-shot methods is highlighted. *BSS via Interpretable AI 3.1.1 does not support multiclass classifications.}
    \fontsize{8}{8}\selectfont
    \centering
    \begin{tabular}{p{1.7cm} p{0.9cm} p{0.8cm} p{0.8cm} p{0.8cm} | p{0.7cm} p{0.7cm} p{0.7cm} p{0.7cm} p{0.7cm}}
    \toprule \\
    \textbf{} & \multicolumn{4}{c}{\textbf{Ours}} & \multicolumn{2}{|c}{\parbox[t]{1.4cm}{\textbf{Interpretable models}}} & \multicolumn{3}{c}{\textbf{AutoML and deep learning}} \\
    \\
        \textbf{Dataset} & \textbf{Claude 3.5 Sonnet} & \textbf{Gemini 1.5 Pro} & \textbf{GPT-4o} & \textbf{GPT-o1} & \textbf{BSS} & \textbf{OCTs} & \textbf{Auto-Gluon} & \textbf{Auto-Prog-nosis} & \textbf{TabPFN} \\ 
        \\ \midrule \\
        \multicolumn{10}{c}{\textbf{Public datasets with evidence of knowledge or memorization by at least 1 LLM}} \\ \hline \\
        boxing1 & 0.44 & 0.47 & 0.50 & \textbf{0.57} & 0.54 & 0.43 & 0.70 & 0.72 & 0.68 \\
        boxing2 & 0.54 & 0.45 & \textbf{0.64} & 0.62 & 0.70 & 0.70 & 0.68 & 0.78 & 0.61 \\
        japansolvent & \textbf{0.78} & 0.56 & 0.70 & 0.70 & 0.77 & 0.67 & 0.83 & 0.71 & 0.88 \\
        colic & 0.57 & 0.52 & 0.59 & \textbf{0.61} & 0.81 & 0.82 & 0.82 & 0.83 & 0.80 \\
        heart\_h & 0.25 & \textbf{0.64} & 0.39 & 0.27 & 0.80 & 0.78 & 0.79 & 0.80 & 0.77 \\
        hepatitis & 0.37 & \textbf{0.57} & 0.54 & \textbf{0.57} & 0.68 & 0.61 & 0.80 & 0.71 & 0.68 \\
        house\_votes\_84 & 0.11 & 0.51 & 0.43 & \textbf{0.87} & 0.94 & 0.94 & 0.96 & 0.95 & 0.95 \\
        labor & 0.63 & \textbf{0.66} & 0.42 & \textbf{0.66} & 0.79 & 0.77 & 0.88 & 0.82 & 0.89 \\
        penguins & \textbf{0.94} & 0.83 & 0.86 & 0.87 & * & 0.97 & 1.00 & 1.00 & 0.99 \\
        vote & 0.39 & \textbf{0.43} & 0.36 & 0.31 & 0.42 & 0.45 & 0.57 & 0.57 & 0.41 \\
        \textbf{median} & 0.49 & 0.54 & 0.52 & \textbf{0.62} & 0.77 & 0.74 & 0.81 & 0.79 & 0.79 \\
        \\ \hline \\
        \multicolumn{10}{c}{\textbf{Public datasets with no clear evidence of knowledge or memorization}} \\ \hline \\
        bankruptcy & 0.59 & 0.70 & \textbf{0.87} & 0.80 & 0.87 & 0.82 & 0.87 & 0.87 & 0.87 \\
        creditscore & 0.55 & \textbf{0.73} & 0.49 & 0.45 & 0.65 & 1.00 & 0.96 & 1.00 & 0.90 \\
        irish & 0.55 & \textbf{0.82} & 0.52 & 0.80 & 0.74 & 0.99 & 0.97 & 0.99 & 0.98 \\
        \textbf{median} & 0.55 & 0.73 & 0.52 & \textbf{0.80} & 0.74 & 0.99 & 0.96 & 0.99 & 0.90 \\
        \\ \hline \\
        \multicolumn{10}{c}{\textbf{Private datasets}} \\ \hline \\
        ACL injury & 0.44 & 0.44 & \textbf{0.66} & 0.49 & 0.49 & 0.49 & 0.33 & 0.44 & 0.49 \\
        post-trauma & \textbf{0.62} & 0.61 & 0.59 & \textbf{0.62} & 0.59 & 0.45 & 0.46 & 0.63 & 0.76 \\
        \textbf{median} & 0.53 & 0.52 & \textbf{0.62} & 0.55 & 0.54 & 0.47 & 0.40 & 0.54 & 0.62 \\
        \\ \bottomrule
    \end{tabular}
\label{table:results_induction_f1}
\end{table*}

\begin{figure*}[t]
\centering
\begin{subfigure}{\textwidth}
    \centering
    \includegraphics[width=0.7\textwidth]{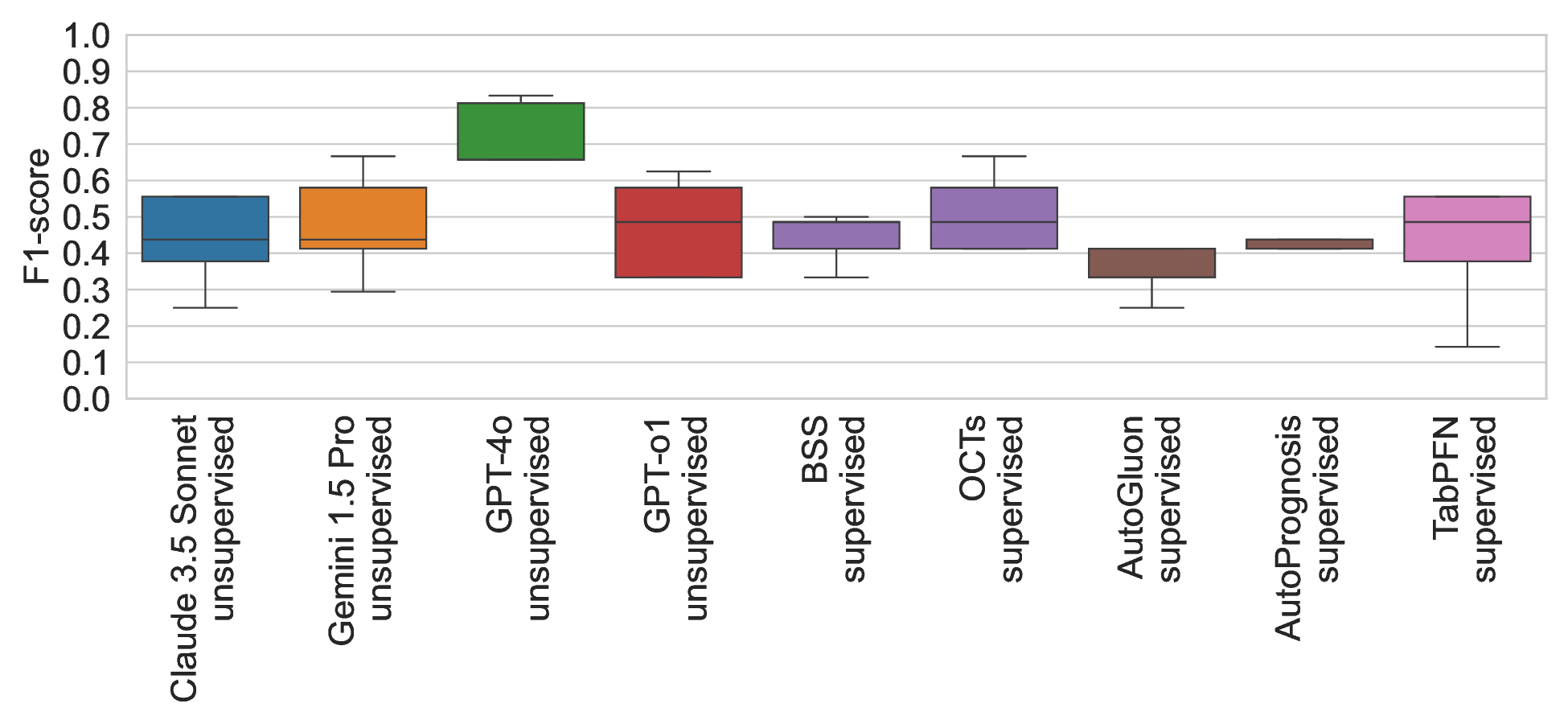}
    \caption{Test F1-score on ACL injury data.}
    \label{fig:boxplot_induction_acl_f1}
\end{subfigure}
\vfill
\begin{subfigure}{\textwidth}
    \centering
    \includegraphics[width=0.7\textwidth]{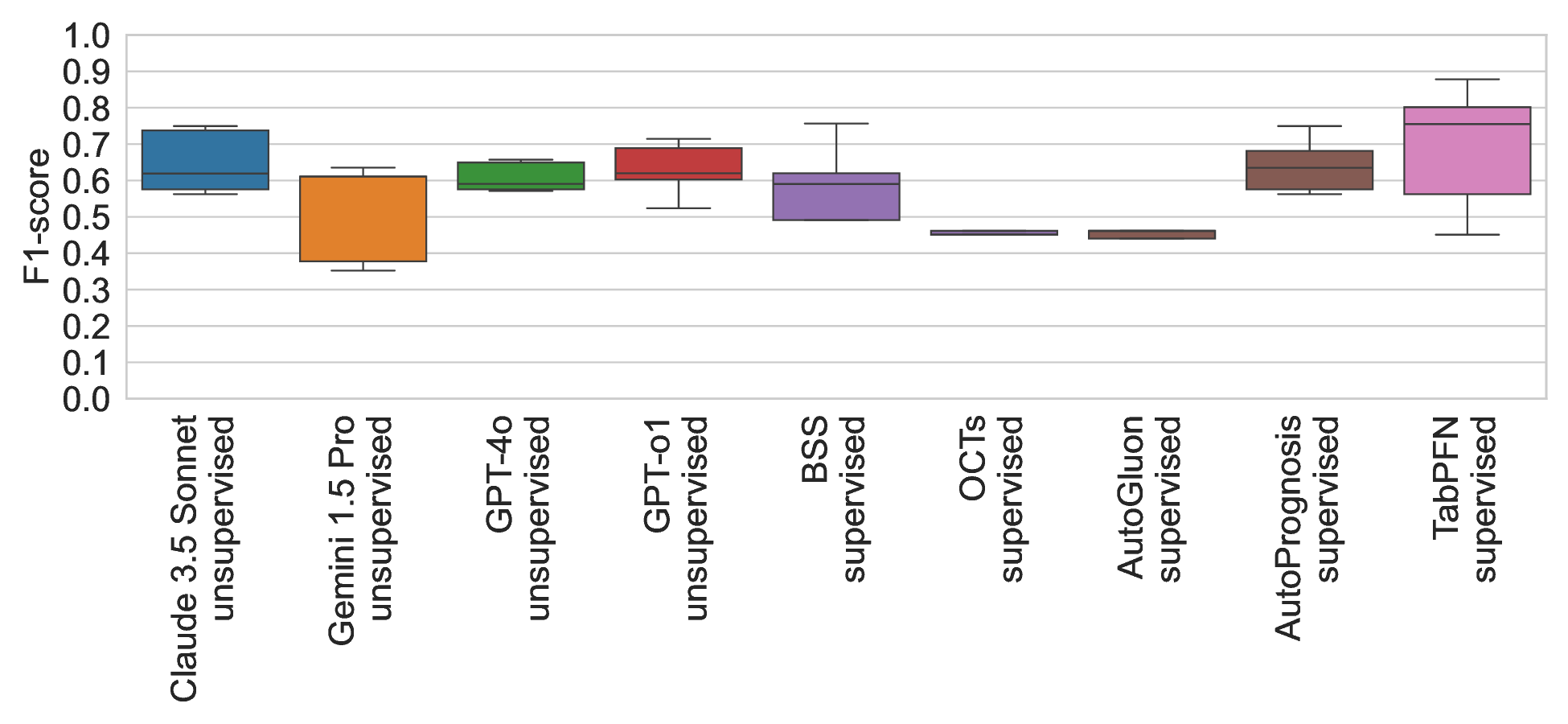}
    \caption{Test F1-score on post-trauma pain data.}
    \label{fig:boxplot_induction_posttrauma_f1}
\end{subfigure}
\caption{Test F1-score at 67\%/33\% train/test splits for our LLM-based zero-shot decision tree induction approach compared to the machine learning baselines on our private (a) ACL injury data and (b) post-trauma pain data.}
\Description{Bar charts showing test F1-scores for our induction experiments for the private data.}
\label{fig:boxplot_induction_acl_posttrauma_f1}
\end{figure*}

\begin{table*}
  \caption{Median test F1-score at 67\%/33\% train/test splits of a multi-layer perceptron without embeddings (first column). Subsequent columns display the performance difference of our LLM-based zero-shot decision tree embedding approach as well as the unsupervised, self-supervised, and supervised embedding baselines relative to the first column. The best score is highlighted.}
  \fontsize{8}{8}\selectfont
  \centering
  \begin{tabular}{p{1.7cm} p{0.9cm} | >{\raggedleft\arraybackslash}p{0.9cm} >{\raggedleft\arraybackslash}p{0.8cm} >{\raggedleft\arraybackslash}p{0.8cm} >
  {\raggedleft\arraybackslash}p{0.8cm} >{\raggedleft\arraybackslash}p{0.8cm} >{\raggedleft\arraybackslash}p{0.8cm} >{\raggedleft\arraybackslash}p{0.8cm} >{\raggedleft\arraybackslash}p{0.8cm} >{\raggedleft\arraybackslash}p{0.8cm} >{\raggedleft\arraybackslash}p{0.8cm} >{\raggedleft\arraybackslash}p{0.8cm}}
  \toprule \\
\textbf{} & \textbf{} & \multicolumn{4}{|c}{\textbf{Ours}} & \multicolumn{1}{c}{\parbox[t]{0.8cm}{\raggedleft\textbf{Un\-su\-per\-vised}}} & \multicolumn{3}{c}{\textbf{Self-supervised}} & \multicolumn{3}{c}{\textbf{Supervised}} \\
\\
    \textbf{Dataset} & \textbf{No embedding} & \textbf{Claude 3.5 Sonnet} & \textbf{Gemini 1.5 Pro} & \textbf{GPT-4o} & \textbf{GPT-o1} & \textbf{Ran-dom trees} & \textbf{Extra trees} & \textbf{Ran-dom forest} & \textbf{XG-Boost} & \textbf{Extra trees} & \textbf{Ran-dom forest} & \textbf{XG-Boost} \\
    \\ \midrule \\
    \multicolumn{13}{c}{\textbf{Public datasets with evidence of knowledge or memorization by at least 1 LLM}} \\ \hline \\
    boxing1 & 0.40 & +0.17 & +0.18 & +0.16 & +0.18 & \textbf{+0.28} & +0.00 & +0.00 & +0.16 & +0.00 & +0.00 & -0.01 \\
    boxing2 & 0.62 & +0.05 & -0.01 & +0.05 & +0.07 & +0.06 & -0.01 & -0.01 & -0.06 & -0.07 & +0.03 & \textbf{+0.12} \\
    japansolvent & 0.48 & +0.35 & +0.35 & +0.35 & \textbf{+0.41} & +0.30 & -0.02 & -0.09 & +0.23 & +0.24 & +0.29 & +0.29 \\
    colic & 0.81 & +0.01 & +0.00 & +0.02 & +0.02 & -0.05 & -0.43 & -0.43 & \textbf{+0.03} & -0.13 & -0.14 & -0.06 \\
    heart\_h & \textbf{0.80} & -0.01 & -0.01 & -0.05 & -0.03 & -0.03 & -0.41 & -0.41 & -0.02 & -0.02 & -0.06 & -0.38 \\
    hepatitis & 0.75 & \textbf{+0.02} & \textbf{+0.02} & \textbf{+0.02} & \textbf{+0.02} & -0.21 & -0.31 & -0.31 & -0.12 & -0.31 & -0.31 & -0.07 \\
    house\_votes\_84 & 0.95 & +0.01 & \textbf{+0.02} & +0.01 & \textbf{+0.02} & -0.02 & -0.22 & -0.04 & +0.00 & -0.04 & -0.05 & \textbf{+0.02} \\
    labor & 0.83 & +0.05 & \textbf{+0.10} & +0.01 & +0.04 & +0.00 & -0.15 & -0.15 & +0.05 & -0.06 & -0.03 & -0.20 \\
    penguins & \textbf{1.00} & -0.04 & -0.04 & -0.01 & -0.03 & -0.03 & -0.41 & -0.46 & -0.40 & -0.28 & -0.08 & -0.04 \\
    vote & 0.43 & -0.02 & +0.01 & +0.02 & -0.03 & -0.01 & -0.04 & -0.05 & \textbf{+0.05} & -0.04 & -0.03 & -0.03 \\
    \textbf{median} & 0.78 & \textbf{+0.02} & \textbf{+0.02} & \textbf{+0.02} & \textbf{+0.02} & -0.01 & -0.19 & -0.12 & +0.01 & -0.05 & -0.04 & -0.03 \\
    \\ \hline \\
    \multicolumn{13}{c}{\textbf{Public datasets with no clear evidence of knowledge or memorization}} \\ \hline \\
    bankruptcy & 0.80 & +0.06 & \textbf{+0.08} & \textbf{+0.08} & +0.02 & +0.01 & +0.01 & +0.01 & +0.07 & +0.00 & -0.03 & -0.06 \\
    creditscore & 0.43 & +0.22 & \textbf{+0.57} & +0.22 & +0.19 & +0.21 & +0.46 & +0.45 & +0.04 & +0.01 & \textbf{+0.57} & +0.54 \\
    irish & 0.99 & -0.01 & -0.01 & -0.01 & -0.01 & -0.02 & -0.29 & -0.28 & -0.06 & -0.12 & -0.10 & \textbf{+0.01} \\
    \textbf{median} & 0.80 & +0.06 & \textbf{+0.08} & \textbf{+0.08} & +0.02 & +0.01 & +0.01 & +0.01 & +0.04 & +0.00 & -0.03 & +0.01 \\
    \\ \hline \\
    \multicolumn{13}{c}{\textbf{Private datasets}} \\ \hline \\
    ACL injury & 0.38 & +0.11 & +0.03 & +0.20 & \textbf{+0.28} & +0.20 & +0.00 & -0.01 & +0.02 & +0.18 & +0.18 & \textbf{+0.28} \\
    post-trauma & 0.55 & +0.07 & \textbf{+0.10} & +0.07 & +0.09 & -0.03 & -0.14 & -0.14 & +0.09 & -0.12 & -0.14 & +0.00 \\
    \textbf{median} & 0.46 & +0.09 & +0.07 & +0.14 & \textbf{+0.18} & +0.09 & -0.07 & -0.08 & +0.05 & +0.03 & +0.02 & +0.14 \\
  \\ \bottomrule
  \end{tabular}
  \label{tab:results_embeddings_f1}
\end{table*}

\begin{figure*}[t]
\centering
\begin{subfigure}{\textwidth}
    \centering
    \includegraphics[width=0.7\textwidth]{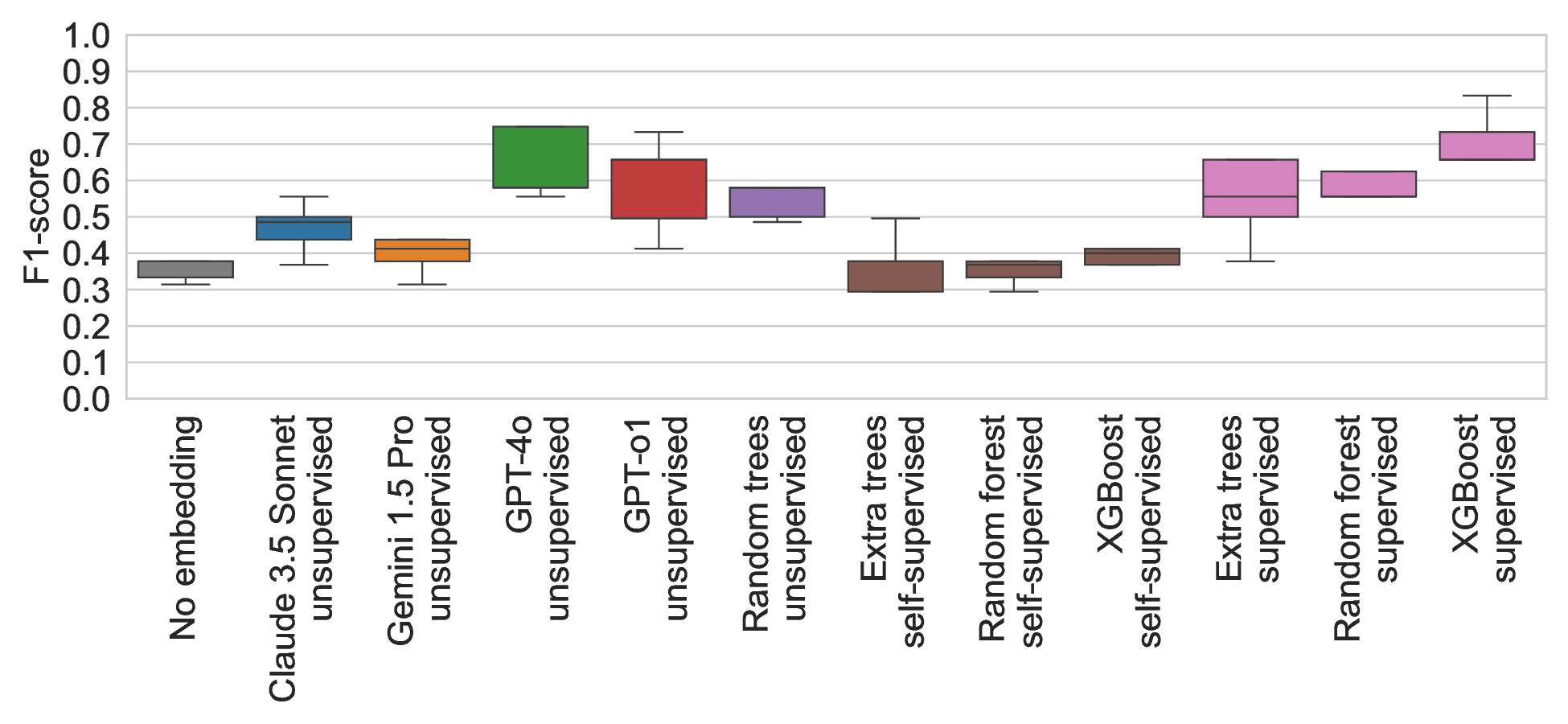}
    \caption{Test F1-score on ACL injury data.}
    \label{fig:boxplot_embedding_acl_f1}
\end{subfigure}
\vfill
\begin{subfigure}{\textwidth}
    \centering
    \includegraphics[width=0.7\textwidth]{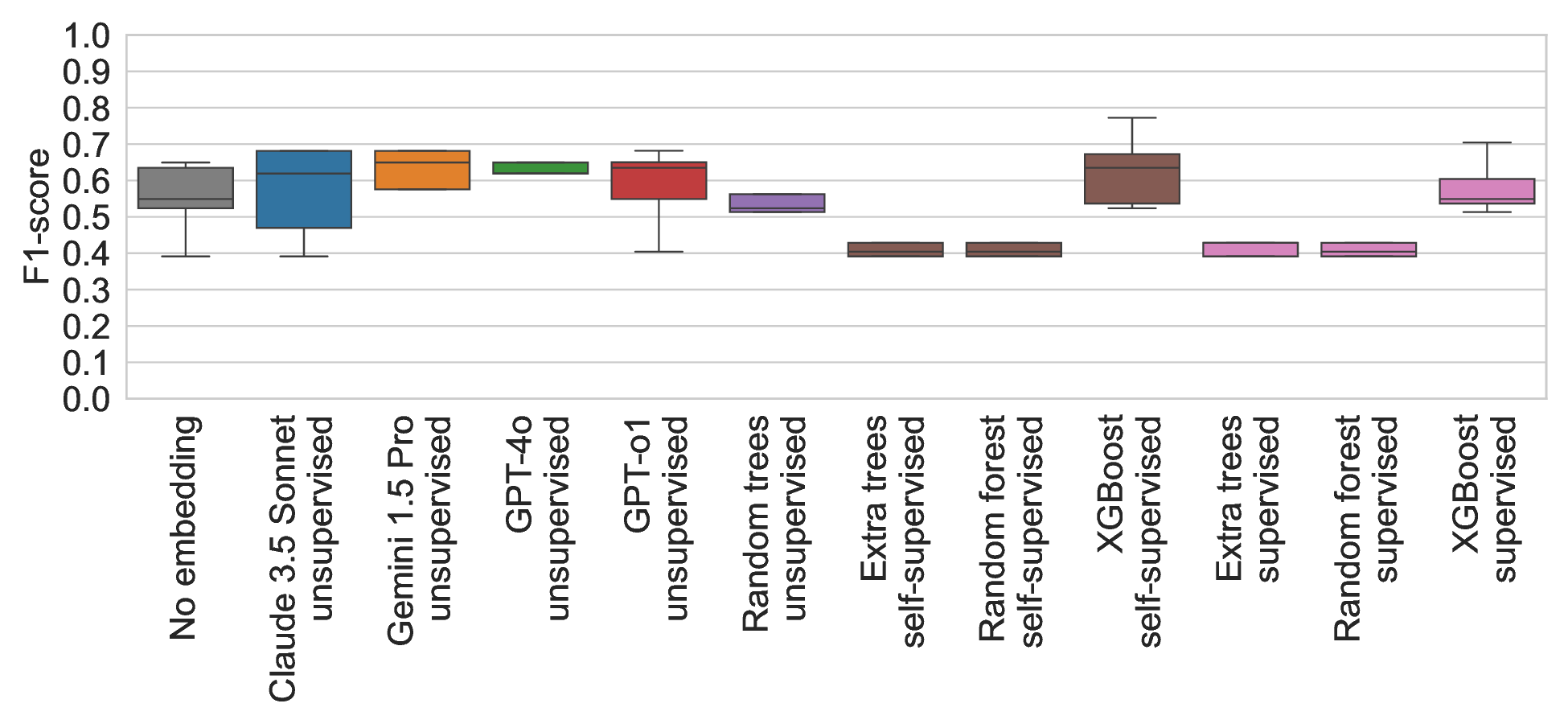}
    \caption{Test F1-score on post-trauma pain data.}
    \label{fig:boxplot_embedding_posttrauma_f1}
\end{subfigure}
\caption{Test F1-score at 67\%/33\% train/test splits of a multi-layer perceptron without embeddings, with our LLM-based zero-shot decision tree embeddings, as well as with unsupervised, self-supervised, and supervised embedding baselines on our private (a) ACL injury data and (b) post-trauma pain data.}
\Description{Bar charts showing test F1-scores for our embedding experiments for the private data.}
\label{fig:boxplot_embedding_acl_posttrauma_f1}
\end{figure*}

In the following, we provide details on the experimental setup to assess our zero-shot decision tree induction (Sect.~\ref{sec:induction}) and embedding (Sect.~\ref{sec:embedding}), including the datasets, preprocessing, machine learning baselines, and performance metrics.

\subsection{Datasets}\label{sec:datasets}

To evaluate our induction and embedding methods, we selected all small-sized tabular datasets from the public Penn Machine Learning Benchmarks (PMLB) \citep{olson2017pmlb,romano2022pmlb} that fulfilled the following criteria:
\begin{itemize}
    \item The sample size was $\leq$ 500. This allowed us to test our zero-shot decision trees in a setting where data-driven machine learning methods often struggle \citep{christodoulou2019systematic,knauer2024pmlbmini} and knowledge-driven or -guided approaches show great promise for predictive modeling \citep{harrell2015regression, heinze2018variable, knauer2023cost,wallsberger2024explainable}.
    \item The dataset covered a classification problem. Regression problems (that are not piecewise linear) typically require deeper trees that are much less interpretable, removing one of \textit{the} key advantages that decision trees have over other machine learning methods.
    \item A data-driven decision tree classifier \citep{bertsimas2017optimal} with a maximum depth of up to 2 could achieve a training F1-score $\geq$ 0.8 on the dataset. This ensured that a shallow decision tree was indeed a reasonable model for the data.
    \item The feature names were informative. As our induction and embedding approaches rely on the feature names to derive decision trees, names like ``xs'' and ``ys''\footnote{``xs'' and ``ys'' are used as feature names in \texttt{https:\allowbreak //epistasislab.github.io/pmlb/profile/prnn\_synth.html}.} cannot be expected to yield meaningful zero-shot models.
\end{itemize}

With these criteria, we arrived at 14 public datasets. We excluded the \textit{iris} dataset, though, because we employed it in our prompt template (Listing~\ref{lst:prompt_template}), yielding a total of 13 public datasets for our evaluation. Please refer to Appendix~\ref{sec:appendix_dataset} for more details about the PMLB datasets, including tests evaluating whether state-of-the-art LLMs know or have memorized these public datasets \citep{bordt2024elephants1,bordt2024elephants2}. Overall, we found evidence of knowledge (54\%) or partial memorization (62\%) of the public datasets by at least 1 state-of-the-art LLM. Furthermore, negative test results do not guarantee that an LLM has not seen or memorized a dataset. Our performance estimates on the public datasets may therefore be overoptimistic. To obtain unbiased performance estimates, we also included 2 small-sized tabular datasets that are private. The private datasets are not publicly available and hence cannot have been seen during the LLMs' pretraining. The prediction target for these datasets was to classify whether the prognosis after musculoskeletal trauma or anterior cruciate ligament (ACL) injury was good or bad. Additional details can be found in Appendix~\ref{sec:appendix_dataset}. 

\subsection{Decision Tree Induction Setup}

\subsubsection{Methods}
For our decision tree induction experiments, we used 4 state-of-the-art LLMs as well as 5 machine learning baselines, \textit{i.e.}, 2 data-driven intrinsically interpretable predictive models, 2 automated machine learning (AutoML) frameworks, and 1 off-the-shelf pretrained deep neural network.

Our zero-shot decision trees were generated with the top 3 LLMs on the Chatbot Arena Leaderboard from July 1st, 2024 \citep{chiang2024chatbotarena}: GPT-4o-2024-05-13 \citep{openai2024gpt4o}, Claude 3.5 Sonnet \citep{anthropic2024claude}, and Gemini-1.5-Pro-API-0514 \citep{google2024gemini}. Additionally, we used GPT-o1-preview, introduced in September 2024, due to its advanced reasoning capabilities \citep{openai2024reasoning}.\footnote{Since our 2 private datasets and 2 of our public datasets come from the healthcare sector (Appendix~\ref{sec:appendix_dataset}), we intended to employ clinical LLMs as well. However, state-of-the-art clinical LLMs are not yet publicly available \citep{matias2023medlm, saab2024capabilities, yang2024advancing}. Even the top 9 models from the Open Medical-LLM Leaderboard \citep{pal2024medicalllm} were not accessible on the Hugging Face Hub on July 1st, 2024, and the next best model was not documented and not robust to synonym substitutions \citep{gallifant2024languagemodels}. We therefore focused our evaluation on general-purpose instead of special-purpose models \citep{nori2023generalist}.} For all LLMs, we set the temperature to 0 for more deterministic outputs and fixed the maximum depth $d$ to 2 for more interpretable trees. Please refer to Sect.~\ref{sec:ablations} for ablations regarding these hyperparameter choices. Furthermore, missing values were imputed using an optimal 10-nearest-neighbors-based imputation \citep{bertsimas2018predictive} before feeding test data to our trees.

As data-driven interpretable baselines, we applied cardinality-constrained best subset selection for logistic regression (BSS) \citep{bertsimas2021sparse} and optimal classification trees (OCTs) \citep{bertsimas2017optimal} using Interpretable AI 3.1.1 with Gurobi 11.0.1. For preprocessing, we again imputed missing values. We then passed nominal feature indicators to both methods and used a 3-fold cross-validation with an F1-score validation criterion to find the best hyperparameters for BSS (\textit{i.e.}, the cardinality from the grid [1, 2, 3, 4] \citep{evans2022estimating} and the regularization strength from the grid [0.1, 0.02, 0.004] \citep{knauer2023cost}). The maximum depth for the OCTs was chosen from the grid [1, 2].

To obtain an upper bound on the predictive performance, we additionally evaluated 2 AutoML frameworks and 1 off-the-shelf pretrained deep neural network: AutoGluon 1.1.1 \citep{erickson2020autogluon, salinas2024tabrepo}, AutoPrognosis 0.1.21 \citep{alaa2018autoprognosis, imrie2023autoprognosis}, and TabPFN 0.1.10 \citep{hollmann2023tabpfn,hollmann2025accurate}. AutoGluon, AutoPrognosis, and TabPFN have recently been shown to achieve a better mean rank in terms of discriminative performance than a simple logistic regression baseline in the low-data regime \citep{knauer2024squeezing}. We used 32 ensemble members and one-hot encoded nominal features for TabPFN \citep{hollmann2023tabpfn}, selected the ``best quality'' preset for AutoGluon, passed nominal feature indicators to TabPFN and AutoGluon, and optimized AutoGluon and AutoPrognosis with an F1-score validation criterion. The runtime limit was 1h on an internal cluster (Rocky Linux 8.6) with 8 vCPU cores and 32 GiB memory (AMD Epyc 7352).

\subsubsection{Evaluation Metrics}
Our primary performance metrics were the test F1-score, \textit{i.e.}, the harmonic mean of precision and sensitivity (or recall) with a macro average over classes, and the test balanced accuracy at a 67\%/33\% train/test split. For binary prediction targets, the balanced accuracy is equal to the arithmetic mean of specificity and sensitivity, or the area under the receiver operating characteristic curve with binary predictions rather than scores. The F1-score and the balanced accuracy are suitable for both binary and multiclass problems as well as balanced and imbalanced data \citep{he2013imbalanced}. Additionally, we evaluated the test F1-score and the test balanced accuracy at 50\%/50\% and 33\%/67\% train/test splits to assess the effect of reducing the training set size for our data-driven baselines. Each split was repeated 5 times to account for randomness in the split, the preprocessing, and the methods themselves. As there is no guarantee that syntactically correct zero-shot trees are generated, we performed API calls until 5 valid trees were found (which required at most 5 additional attempts per tree).

\subsection{Decision Tree Embedding Setup}

\subsubsection{Methods}
For our decision tree embedding experiments, we used the same 4 state-of-the-art LLMs as in our induction setup. To introduce greater variability in our tree generation, though, we set the LLMs' temperature to the default value of 1.0, fostering more diverse and creative outputs (Sect.~\ref{sec:embedding}). We generated 5 decision trees for each LLM to form our decision forest for the embedding and concatenated the embedding with the original feature vector. We then applied our embeddings as inputs to a simple multi-layer perceptron classifier \citep{borisov_deeptlf_2023}. We trained the artificial neural network with a single hidden layer and optimized its size from the grid [10, 25, 50, 75, 100] along with the regularization strength from [0.0001, 0.001, 0.01, 0.1, 1.0], using a 3-fold cross-validation with an F1-score validation criterion. Please refer to Sect.~\ref{sec:ablations} for ablations regarding these hyperparameter choices.

We compared our zero-shot embeddings to unsupervised random trees embeddings \citep{moosmann2006fast} with 5 trees. Furthermore, we fitted random forests \citep{breiman2001random}, extremely randomized trees \citep{geurts2006extremely}, and XGBoost \citep{chen2016xgboost} with 5 trees on our design matrices or labels to derive self-supervised or supervised tree embeddings \citep{borisov_deeptlf_2023}. A multi-layer perceptron classifier without any embedding served as a baseline for our experiments. The random trees embedding, random forest, extremely randomized trees, and multi-layer perceptron were run from scikit-learn 1.5.1, XGBoost from the XGBoost Python package 2.1.0. For all methods, missing values were imputed, nominal features one-hot encoded, and numerical features ``min-max'' scaled.

\subsubsection{Evaluation Metrics}
We assessed the predictive performance of each method with the test F1-score and balanced accuracy at a 67\%/33\% train/test split. We again repeated the split 5 times to account for randomness. To gain a deeper understanding of our LLM-based embedding approach, we evaluated the dimensionality of our embeddings (an indirect measure of the maximum tree depths chosen by the LLMs) and the selected feature variability (an indirect measure of the LLMs' output diversity) per LLM and dataset.

\section{Experimental Results} \label{sec:results}

In this section, we present the experimental results for our zero-shot decision tree induction (Sect.~\ref{sec:induction}) and embedding (Sect.~\ref{sec:embedding}). We noted that the LLMs often generated explanations in addition to the decision trees, which can provide valuable insights into their reasoning processes.\footnote{On our post-trauma pain data, for example, the following explanation was provided by Claude 3.5 Sonnet: ``These features [average pain intensity, Short Form 36 (SF-36) physical component summary, pain self-efficacy questionnaire (PSEQ)] were chosen because they directly relate to physical functioning, pain management, and pain intensity, which are crucial factors in determining the outcome of pain and disability following musculoskeletal trauma.''} In terms of predictive performance, we found that LLM-based zero-shot decision tree generations showed a better average performance than the data-driven OCTs on 27\% of our evaluated datasets, and that the zero-shot decision tree embeddings generated by Gemini 1.5 Pro were statistically significantly better than data-driven tree-based embeddings. We therefore argue that our induction and embedding approaches can serve as new knowledge-driven baselines for data-driven machine learning methods in the low-data regime.

\subsection{Decision Tree Induction Results} \label{sec:results_induction}

The results for the median test F1-score for our induction experiments can be found in Table~\ref{table:results_induction_f1} and the balanced accuracy results in Table~\ref{table:results_induction_acc} in Appendix~\ref{sec:appendix_results}. Figure~\ref{fig:boxplot_induction_acl_posttrauma_f1} and Figure~\ref{fig:boxplot_induction_acl_posttrauma_acc} in Appendix~\ref{sec:appendix_results} show test F1-score and balanced accuracy boxplots for our private datasets, respectively. Figure~\ref{fig:results_induction_cdd} in Appendix~\ref{sec:appendix_results} shows the statistical significance of our findings.

Our zero-shot trees showed mixed results in the induction experiments. On average, they achieved a statistically significantly worse performance than data-driven trees (Figure~\ref{fig:results_induction_cdd}), with median test F1-score / balanced accuracy differences of -0.22 / -0.19, -0.20 / -0.16, -0.18 / -0.14, and -0.15 / -0.12 for Claude 3.5 Sonnet, Gemini 1.5 Pro, GPT-4o, and GPT-o1. However, on 27\% of the datasets, knowledge-driven trees reached an appreciably better performance than data-driven trees, with median test F1-score / balanced accuracy differences up to +0.17 / +0.17, +0.16 / +0.11, +0.17 / +0.19, and +0.17 / +0.13 for Claude 3.5 Sonnet, Gemini 1.5 Pro, GPT-4o, and GPT-o1. On our ACL injury data, GPT-4o achieved an appreciably better performance than all induction baselines. Compared to AutoGluon, for instance, the median test F1-score / balanced accuracy differences were +0.33 / +0.19. On our post-trauma data, Claude 3.5 Sonnet, Gemini 1.5 Pro, and GPT-o1 reached an appreciably better performance than data-driven trees and AutoGluon, with median test F1-score / balanced accuracy differences up to +0.17 / +0.17, +0.16 / +0.11, and +0.17 / +0.13. Please refer to Appendix~\ref{sec:appendix_results} for results on the predictive performance at train/test splits other than 67\%/33\% and for details on the best trees for our private data.

\subsection{Decision Tree Embedding Results}

The median test F1-score results for our embedding experiments can be found in Table~\ref{tab:results_embeddings_f1} and the balanced accuracy results in Table~\ref{tab:results_embeddings_acc} in Appendix~\ref{sec:appendix_results}. Figure~\ref{fig:boxplot_embedding_acl_posttrauma_f1} and Figure~\ref{fig:boxplot_embedding_acl_posttrauma_acc} in Appendix~\ref{sec:appendix_results} display test F1-score and balanced accuracy boxplots for our private datasets, respectively. Figure~\ref{fig:results_embedding_cdd} in Appendix~\ref{sec:appendix_results} shows the statistical significance of our findings.

Our zero-shot embeddings showed impressive results compared to the data-driven baselines. Gemini 1.5 Pro achieved a performance that was on par with best induction baseline, \textit{i.e.}, AutoPrognosis, with a median test F1-score / balanced accuracy difference of -0.01 / -0.02. Compared to no embedding and the embedding baselines, the median test F1-score / balanced accuracy differences were up to +0.32 / +0.28 and statistically significant (Figure~\ref{fig:results_embedding_cdd}). On our ACL data, GPT-o1 reached the same performance as the best induction and embedding methods for this dataset, \textit{i.e.}, GPT-4o and supervised XGBoost, with median test F1-score / balanced accuracy differences of 0.00 / 0.00. On our post-trauma data, Gemini 1.5 Pro achieved a similar performance to the best interpretable and embedding methods for this dataset, \textit{i.e.}, Claude 3.5 Sonnet and self-supervised XGBoost, with median test F1-score / balanced accuracy differences of +0.03 / -0.03 and +0.01 / +0.02. Please refer to Appendix~\ref{sec:appendix_results} for more insights regarding the dimensionality of our embeddings and the selected feature variability.

\subsection{Ablations} \label{sec:ablations}

To validate our experimental design choices, we performed an extensive ablation study for our zero-shot decision tree inductions and embeddings. For our induction experiments, we analyzed a broader range of maximum tree depths ([1, 3, 4, 5] instead of 2), temperatures ([0.5, 1.0] instead of 0), and in-context examples ([2, 3] instead of 1). Additionally, we evaluated the effect of including a dataset (and feature) description for \textit{bankruptcy}, for which an extensive description is provided in \citet{simonoff2003analyzing}, as well as the effect of excluding our decision tree output indicator. For our embedding experiments, we considered different maximum tree depths ([1, 2, 3, 4, 5] instead of no restriction), temperatures ([0, 0.5] instead of 1.0), in-context examples ([2, 3] instead of 1), and number of generated trees ([1, 2, 3, 4] instead of 5). We assessed the effect of including a dataset (and feature) description for \textit{bankruptcy} as in our induction setup \citep{simonoff2003analyzing}, the effect of replacing instead of extending the original feature vectors with our embeddings, as well as the effect of using gradient-boosted decision trees or logistic regression instead of multi-layer perceptrons as downstream classifiers. Overall, we generated close to 5000 trees for our ablations.

The results for our induction and embedding experiments are presented in Table~\ref{tab:ablations_induction} and Table~\ref{tab:ablations_embedding} in Appendix~\ref{sec:appendix_results}. The performance of the ablation experiments was generally comparable to or worse than the performance of our experimental defaults. The only exception was the inclusion of the dataset (and feature) description for \textit{bankruptcy}, with test F1-score / balanced accuracy improvements up to +0.13 / 0.17, +0.11 / +0.13, and +0.18 / +0.15 for Gemini 1.5 Pro, GPT-4o, and GPT-o1. This suggests that our LLM-based decision tree induction and embedding approaches can profit from additional background context, for instance when domain- or task-specific knowledge is limited, and highlights opportunities to refine our prompting template for better performance in the future \citep{dong2024survey,nori2023generalist,schulhoff2024promptreport,yuksekgonul2025optimizing}.

\section{Conclusion}

LLMs provide powerful means to leverage prior knowledge in the low-data regime. In this work, we presented how we can use the condensed world knowledge within state-of-the-art LLMs to induce decision trees \textit{without any training data}. We showed that these zero-shot trees can even surpass data-driven trees on some small-sized tabular datasets, while remaining intrinsically interpretable and privacy-preserving. Additionally, we demonstrated how we can generate knowledge-driven embeddings using our zero-shot trees and combine them with downstream models to infuse prior knowledge into data-driven approaches, with a predictive performance that is better than data-driven tree-based embeddings on average. We therefore argue that our tree induction and embedding approaches can serve as new knowledge-driven baselines for data-driven machine learning methods when data is scarce.

Nevertheless, we want to point out that our conclusions are so far based on small-sized tabular classification datasets and do not necessarily extend to other settings. We also focused on a simple prompting template that could be adapted for better performance \citep{dong2024survey,nori2023generalist,schulhoff2024promptreport,yuksekgonul2025optimizing}. We also expect further performance boosts with the rise of even more powerful LLMs \citep{google2024gemini2,openai2024pro} and by iteratively improving our zero-shot trees with training data \citep{nam2024featuregeneration,snell2025testtimecompute,yuksekgonul2025optimizing}, \textit{e.g.}, via evolutionary algorithms \citep{chao2024llmsmeet,eiben2015introduction,lee2025evolvingdeeperllmthinking,liu2025decision,ryan2018handbook}. Alternatively, our trees could be used to obtain probabilistic outputs, \textit{e.g.}, by counting the fraction of training examples in each leaf. Even different probabilistic classifiers like logistic regression could potentially be generated following our general template \citep{garg2022can,grazzi2024mamba}. Overall, we believe that leveraging LLMs as zero-shot model generators opens up a new toolbox for practitioners and researchers to tailor LLM-based machine learning models to their needs.

\section*{Ethical Use of Data}
The public data used in this article was sourced from PMLB \citep{olson2017pmlb,romano2022pmlb}. The private data \citep{evans2022estimating,brisson2020cartilage,brisson2024cartilage} contains sensitive information and therefore cannot be shared publicly due to privacy restrictions. Ethical approval was obtained from the relevant Institutional Ethics Review Boards to ensure compliance with ethical standards and the protection of participants' rights and privacy. Please also note that the proposed methods and prognostic models derived from the private data are not validated for clinical use.

\begin{acks}
The work in this article was funded by the Bundesministerium für Bildung und Forschung (16DHBKI071), the Deutsche Forschungsgemeinschaft (CRC 1444, BR 6698/1-1, 528483508), the European Research Council (101054501), and the National Institute for Health Research (501100000272).
\end{acks}

\bibliographystyle{ACM-Reference-Format}
\balance
\bibliography{sample-base}


\begin{thebibliography}{97}


\ifx \showCODEN    \undefined \def \showCODEN     #1{\unskip}     \fi
\ifx \showISBNx    \undefined \def \showISBNx     #1{\unskip}     \fi
\ifx \showISBNxiii \undefined \def \showISBNxiii  #1{\unskip}     \fi
\ifx \showISSN     \undefined \def \showISSN      #1{\unskip}     \fi
\ifx \showLCCN     \undefined \def \showLCCN      #1{\unskip}     \fi
\ifx \shownote     \undefined \def \shownote      #1{#1}          \fi
\ifx \showarticletitle \undefined \def \showarticletitle #1{#1}   \fi
\ifx \showURL      \undefined \def \showURL       {\relax}        \fi
\providecommand\bibfield[2]{#2}
\providecommand\bibinfo[2]{#2}
\providecommand\natexlab[1]{#1}
\providecommand\showeprint[2][]{arXiv:#2}

\bibitem[Alaa and van~der Schaar(2018)]%
        {alaa2018autoprognosis}
\bibfield{author}{\bibinfo{person}{Ahmed Alaa} {and} \bibinfo{person}{Mihaela van~der Schaar}.} \bibinfo{year}{2018}\natexlab{}.
\newblock \showarticletitle{Autoprognosis: Automated clinical prognostic modeling via bayesian optimization with structured kernel learning}. In \bibinfo{booktitle}{\emph{Proceedings of the 35th International Conference on Machine Learning (ICML)}}. \bibinfo{publisher}{PMLR}, \bibinfo{address}{Stockholm, Sweden}, \bibinfo{pages}{139--148}.
\newblock


\bibitem[Anthropic(2024a)]%
        {anthropic2024claude}
\bibfield{author}{\bibinfo{person}{Anthropic}.} \bibinfo{year}{2024}\natexlab{a}.
\newblock \bibinfo{title}{The Claude 3 model family: Opus, Sonnet, Haiku}.
\newblock \bibinfo{howpublished}{https://assets.anthropic.com/m/61e7d27f8c8f5919/original/Claude-3-Model-Card.pdf}.
\newblock
\newblock
\shownote{Accessed: 2024-07-01}.


\bibitem[Anthropic(2024b)]%
        {anthropic2024sensitive}
\bibfield{author}{\bibinfo{person}{Anthropic}.} \bibinfo{year}{2024}\natexlab{b}.
\newblock \bibinfo{title}{Data Processing Addendum}.
\newblock \bibinfo{howpublished}{https://www.anthropic.com/legal/commercial-terms}.
\newblock
\newblock
\shownote{Accessed: 2024-07-01}.


\bibitem[Banko and Brill(2001)]%
        {banko2001scaling}
\bibfield{author}{\bibinfo{person}{Michele Banko} {and} \bibinfo{person}{Eric Brill}.} \bibinfo{year}{2001}\natexlab{}.
\newblock \showarticletitle{Scaling to very very large corpora for natural language disambiguation}. In \bibinfo{booktitle}{\emph{Proceedings of the 39th Annual Meeting of the Association for Computational Linguistics}}. \bibinfo{publisher}{Association for Computational Linguistics (ACL)}, \bibinfo{address}{Toulouse, France}, \bibinfo{pages}{26--33}.
\newblock


\bibitem[Benavoli et~al\mbox{.}(2016)]%
        {benavoli2016should}
\bibfield{author}{\bibinfo{person}{Alessio Benavoli}, \bibinfo{person}{Giorgio Corani}, {and} \bibinfo{person}{Francesca Mangili}.} \bibinfo{year}{2016}\natexlab{}.
\newblock \showarticletitle{Should we really use post-hoc tests based on mean-ranks?}
\newblock \bibinfo{journal}{\emph{The Journal of Machine Learning Research}} \bibinfo{volume}{17}, \bibinfo{number}{1} (\bibinfo{year}{2016}), \bibinfo{pages}{152--161}.
\newblock
\urldef\tempurl%
\url{http://jmlr.org/papers/v17/benavoli16a.html}
\showURL{%
\tempurl}


\bibitem[Bertsimas and Dunn(2017)]%
        {bertsimas2017optimal}
\bibfield{author}{\bibinfo{person}{Dimitris Bertsimas} {and} \bibinfo{person}{Jack Dunn}.} \bibinfo{year}{2017}\natexlab{}.
\newblock \showarticletitle{Optimal classification trees}.
\newblock \bibinfo{journal}{\emph{Machine Learning}}  \bibinfo{volume}{106} (\bibinfo{year}{2017}), \bibinfo{pages}{1039--1082}.
\newblock
\href{https://doi.org/10.1007/s10994-017-5633-9}{doi:\nolinkurl{10.1007/s10994-017-5633-9}}


\bibitem[Bertsimas et~al\mbox{.}(2021)]%
        {bertsimas2021sparse}
\bibfield{author}{\bibinfo{person}{Dimitris Bertsimas}, \bibinfo{person}{Jean Pauphilet}, {and} \bibinfo{person}{Bart Van~Parys}.} \bibinfo{year}{2021}\natexlab{}.
\newblock \showarticletitle{Sparse classification: a scalable discrete optimization perspective}.
\newblock \bibinfo{journal}{\emph{Machine Learning}}  \bibinfo{volume}{110} (\bibinfo{year}{2021}), \bibinfo{pages}{3177--3209}.
\newblock
\href{https://doi.org/10.1007/s10994-021-06085-5}{doi:\nolinkurl{10.1007/s10994-021-06085-5}}


\bibitem[Bertsimas et~al\mbox{.}(2018)]%
        {bertsimas2018predictive}
\bibfield{author}{\bibinfo{person}{Dimitris Bertsimas}, \bibinfo{person}{Colin Pawlowski}, {and} \bibinfo{person}{Ying~Daisy Zhuo}.} \bibinfo{year}{2018}\natexlab{}.
\newblock \showarticletitle{From predictive methods to missing data imputation: an optimization approach}.
\newblock \bibinfo{journal}{\emph{Journal of Machine Learning Research}} \bibinfo{volume}{18}, \bibinfo{number}{196} (\bibinfo{year}{2018}), \bibinfo{pages}{1--39}.
\newblock
\urldef\tempurl%
\url{http://jmlr.org/papers/v18/17-073.html}
\showURL{%
\tempurl}


\bibitem[Biswal et~al\mbox{.}(2002)]%
        {biswal2002risk}
\bibfield{author}{\bibinfo{person}{Sandip Biswal}, \bibinfo{person}{Trevor Hastie}, \bibinfo{person}{Thomas~P Andriacchi}, \bibinfo{person}{Gabrielle~A Bergman}, \bibinfo{person}{Michael~F Dillingham}, {and} \bibinfo{person}{Philipp Lang}.} \bibinfo{year}{2002}\natexlab{}.
\newblock \showarticletitle{Risk factors for progressive cartilage loss in the knee: a longitudinal magnetic resonance imaging study in forty-three patients}.
\newblock \bibinfo{journal}{\emph{Arthritis \& Rheumatism: Official Journal of the American College of Rheumatology}} \bibinfo{volume}{46}, \bibinfo{number}{11} (\bibinfo{year}{2002}), \bibinfo{pages}{2884--2892}.
\newblock
\href{https://doi.org/10.1002/art.10573}{doi:\nolinkurl{10.1002/art.10573}}


\bibitem[Bordt et~al\mbox{.}(2024a)]%
        {bordt2024elephants1}
\bibfield{author}{\bibinfo{person}{Sebastian Bordt}, \bibinfo{person}{Harsha Nori}, {and} \bibinfo{person}{Rich Caruana}.} \bibinfo{year}{2024}\natexlab{a}.
\newblock \bibinfo{title}{Elephants Never Forget: Testing Language Models for Memorization of Tabular Data}.
\newblock
\showeprint[arxiv]{2403.06644}~[cs.LG]
\urldef\tempurl%
\url{https://arxiv.org/abs/2403.06644}
\showURL{%
\tempurl}


\bibitem[Bordt et~al\mbox{.}(2024b)]%
        {bordt2024elephants2}
\bibfield{author}{\bibinfo{person}{Sebastian Bordt}, \bibinfo{person}{Harsha Nori}, \bibinfo{person}{Vanessa Rodrigues}, \bibinfo{person}{Besmira Nushi}, {and} \bibinfo{person}{Rich Caruana}.} \bibinfo{year}{2024}\natexlab{b}.
\newblock \bibinfo{title}{Elephants Never Forget: Memorization and Learning of Tabular Data in Large Language Models}.
\newblock
\showeprint[arxiv]{2404.06209}~[cs.LG]
\urldef\tempurl%
\url{https://arxiv.org/abs/2404.06209}
\showURL{%
\tempurl}


\bibitem[Borisov et~al\mbox{.}(2023)]%
        {borisov_deeptlf_2023}
\bibfield{author}{\bibinfo{person}{Vadim Borisov}, \bibinfo{person}{Klaus Broelemann}, \bibinfo{person}{Enkelejda Kasneci}, {and} \bibinfo{person}{Gjergji Kasneci}.} \bibinfo{year}{2023}\natexlab{}.
\newblock \showarticletitle{{DeepTLF}: robust deep neural networks for heterogeneous tabular data}.
\newblock \bibinfo{journal}{\emph{International Journal of Data Science and Analytics}} \bibinfo{volume}{16}, \bibinfo{number}{1} (\bibinfo{year}{2023}), \bibinfo{pages}{85--100}.
\newblock
\showISSN{2364-4168}
\href{https://doi.org/10.1007/s41060-022-00350-z}{doi:\nolinkurl{10.1007/s41060-022-00350-z}}


\bibitem[Breiman(2001)]%
        {breiman2001random}
\bibfield{author}{\bibinfo{person}{Leo Breiman}.} \bibinfo{year}{2001}\natexlab{}.
\newblock \showarticletitle{Random forests}.
\newblock \bibinfo{journal}{\emph{Machine learning}}  \bibinfo{volume}{45} (\bibinfo{year}{2001}), \bibinfo{pages}{5--32}.
\newblock
\href{https://doi.org/10.1023/A:1010933404324}{doi:\nolinkurl{10.1023/A:1010933404324}}


\bibitem[Brisson et~al\mbox{.}(2020)]%
        {brisson2020cartilage}
\bibfield{author}{\bibinfo{person}{N.~M. Brisson}, \bibinfo{person}{L.~A.~N. Krahl}, \bibinfo{person}{W. Wirth}, \bibinfo{person}{F. Eckstein}, {and} \bibinfo{person}{G.~N. Duda}.} \bibinfo{year}{2020}\natexlab{}.
\newblock \showarticletitle{Knee Cartilage Thickness of ACL-Injured Non-Copers \& Copers: A 1-Year Study}. In \bibinfo{booktitle}{\emph{Proceedings of the 14th International Workshop on Osteoarthritis Imaging}}. \bibinfo{publisher}{International Society of Osteoarthritis Imaging}, \bibinfo{address}{Salzburg, Austria}.
\newblock


\bibitem[Brisson et~al\mbox{.}(2024)]%
        {brisson2024cartilage}
\bibfield{author}{\bibinfo{person}{N.~M. Brisson}, \bibinfo{person}{L.~A.~N. Krahl}, \bibinfo{person}{W. Wirth}, \bibinfo{person}{F. Eckstein}, {and} \bibinfo{person}{G.~N. Duda}.} \bibinfo{year}{2024}\natexlab{}.
\newblock \showarticletitle{Knee Cartilage Transverse Relaxation Time (T2) in Patients with Anterior Cruciate Ligament Injury Treated Conservatively with Structured Physical Rehabilitation or Standard of Care}. In \bibinfo{booktitle}{\emph{Proceedings of the 70th Annual Meeting of the Orthopaedic Research Society}}. \bibinfo{publisher}{Orthopaedic Research Society}, \bibinfo{address}{Long Beach, USA}.
\newblock


\bibitem[Brown et~al\mbox{.}(2020)]%
        {brown2020language}
\bibfield{author}{\bibinfo{person}{Tom Brown}, \bibinfo{person}{Benjamin Mann}, \bibinfo{person}{Nick Ryder}, \bibinfo{person}{Melanie Subbiah}, \bibinfo{person}{Jared~D Kaplan}, \bibinfo{person}{Prafulla Dhariwal}, \bibinfo{person}{Arvind Neelakantan}, \bibinfo{person}{Pranav Shyam}, \bibinfo{person}{Girish Sastry}, \bibinfo{person}{Amanda Askell}, {et~al\mbox{.}}} \bibinfo{year}{2020}\natexlab{}.
\newblock \showarticletitle{Language models are few-shot learners}.
\newblock \bibinfo{journal}{\emph{Advances in neural information processing systems}}  \bibinfo{volume}{33} (\bibinfo{year}{2020}), \bibinfo{pages}{1877--1901}.
\newblock
\href{https://doi.org/10.48550/arXiv.2005.14165}{doi:\nolinkurl{10.48550/arXiv.2005.14165}}


\bibitem[Chao et~al\mbox{.}(2024)]%
        {chao2024llmsmeet}
\bibfield{author}{\bibinfo{person}{Wang Chao}, \bibinfo{person}{Jiaxuan Zhao}, \bibinfo{person}{Licheng Jiao}, \bibinfo{person}{Lingling Li}, \bibinfo{person}{Fang Liu}, {and} \bibinfo{person}{Shuyuan Yang}.} \bibinfo{year}{2024}\natexlab{}.
\newblock \bibinfo{title}{When large language models meet evolutionary algorithms}.
\newblock
\showeprint[arxiv]{2401.10510}~[cs.NE]
\urldef\tempurl%
\url{https://arxiv.org/abs/2401.10510}
\showURL{%
\tempurl}


\bibitem[Chen and Guestrin(2016)]%
        {chen2016xgboost}
\bibfield{author}{\bibinfo{person}{Tianqi Chen} {and} \bibinfo{person}{Carlos Guestrin}.} \bibinfo{year}{2016}\natexlab{}.
\newblock \showarticletitle{Xgboost: A scalable tree boosting system}. In \bibinfo{booktitle}{\emph{Proceedings of the 22nd ACM SIGKDD international conference on knowledge discovery and data mining}}. \bibinfo{publisher}{Association for Computing Machinery (ACM)}, \bibinfo{address}{San Francisco, USA}, \bibinfo{pages}{785--794}.
\newblock


\bibitem[Chiang et~al\mbox{.}(2024)]%
        {chiang2024chatbotarena}
\bibfield{author}{\bibinfo{person}{Wei-Lin Chiang}, \bibinfo{person}{Lianmin Zheng}, \bibinfo{person}{Ying Sheng}, \bibinfo{person}{Anastasios~N. Angelopoulos}, \bibinfo{person}{Tianle Li}, \bibinfo{person}{Dacheng Li}, \bibinfo{person}{Hao Zhang}, \bibinfo{person}{Banghua Zhu}, \bibinfo{person}{Hao Zhang}, \bibinfo{person}{Michael~I. Jordan}, \bibinfo{person}{Joseph~E. Gonzalez}, {and} \bibinfo{person}{Ion Stoica}.} \bibinfo{year}{2024}\natexlab{}.
\newblock \showarticletitle{Chatbot arena: an open platform for evaluating LLMs by human preference}. In \bibinfo{booktitle}{\emph{Proceedings of the 41st International Conference on Machine Learning (ICML)}}. \bibinfo{publisher}{PMLR}, \bibinfo{address}{Vienna, Austria}, \bibinfo{pages}{8359--8388}.
\newblock


\bibitem[Christodoulou et~al\mbox{.}(2019)]%
        {christodoulou2019systematic}
\bibfield{author}{\bibinfo{person}{Evangelia Christodoulou}, \bibinfo{person}{Jie Ma}, \bibinfo{person}{Gary~S. Collins}, \bibinfo{person}{Ewout~W. Steyerberg}, \bibinfo{person}{Jan~Y. Verbakel}, {and} \bibinfo{person}{Ben Van~Calster}.} \bibinfo{year}{2019}\natexlab{}.
\newblock \showarticletitle{A systematic review shows no performance benefit of machine learning over logistic regression for clinical prediction models}.
\newblock \bibinfo{journal}{\emph{Journal of clinical epidemiology}}  \bibinfo{volume}{110} (\bibinfo{year}{2019}), \bibinfo{pages}{12--22}.
\newblock
\href{https://doi.org/10.1016/j.jclinepi.2019.02.004}{doi:\nolinkurl{10.1016/j.jclinepi.2019.02.004}}


\bibitem[Dem{\v{s}}ar(2006)]%
        {demvsar2006statistical}
\bibfield{author}{\bibinfo{person}{Janez Dem{\v{s}}ar}.} \bibinfo{year}{2006}\natexlab{}.
\newblock \showarticletitle{Statistical comparisons of classifiers over multiple data sets}.
\newblock \bibinfo{journal}{\emph{The Journal of Machine learning research}}  \bibinfo{volume}{7} (\bibinfo{year}{2006}), \bibinfo{pages}{1--30}.
\newblock
\urldef\tempurl%
\url{http://jmlr.org/papers/v7/demsar06a.html}
\showURL{%
\tempurl}


\bibitem[Dong et~al\mbox{.}(2024)]%
        {dong2024survey}
\bibfield{author}{\bibinfo{person}{Qingxiu Dong}, \bibinfo{person}{Lei Li}, \bibinfo{person}{Damai Dai}, \bibinfo{person}{Ce Zheng}, \bibinfo{person}{Jingyuan Ma}, \bibinfo{person}{Rui Li}, \bibinfo{person}{Heming Xia}, \bibinfo{person}{Jingjing Xu}, \bibinfo{person}{Zhiyong Wu}, \bibinfo{person}{Baobao Chang}, \bibinfo{person}{Xu Sun}, \bibinfo{person}{Lei Li}, {and} \bibinfo{person}{Zhifang Sui}.} \bibinfo{year}{2024}\natexlab{}.
\newblock \showarticletitle{A Survey on In-context Learning}. In \bibinfo{booktitle}{\emph{Proceedings of the 2024 Conference on Empirical Methods in Natural Language Processing (EMNLP)}}. \bibinfo{publisher}{Association for Computational Linguistics (ACL)}, \bibinfo{address}{Miami, USA}, \bibinfo{pages}{1107--1128}.
\newblock


\bibitem[Eiben and Smith(2015)]%
        {eiben2015introduction}
\bibfield{author}{\bibinfo{person}{Agoston~E. Eiben} {and} \bibinfo{person}{James~E. Smith}.} \bibinfo{year}{2015}\natexlab{}.
\newblock \bibinfo{booktitle}{\emph{Introduction to evolutionary computing}}.
\newblock \bibinfo{publisher}{Springer}.
\newblock


\bibitem[Elhoseiny et~al\mbox{.}(2013)]%
        {elhoseiny2013write}
\bibfield{author}{\bibinfo{person}{Mohamed Elhoseiny}, \bibinfo{person}{Babak Saleh}, {and} \bibinfo{person}{Ahmed Elgammal}.} \bibinfo{year}{2013}\natexlab{}.
\newblock \showarticletitle{Write a classifier: Zero-shot learning using purely textual descriptions}. In \bibinfo{booktitle}{\emph{Proceedings of the IEEE International Conference on Computer Vision}}. \bibinfo{publisher}{Institute of Electrical and Electronics Engineers (IEEE)}, \bibinfo{address}{Sydney, Australia}, \bibinfo{pages}{2584--2591}.
\newblock


\bibitem[Erickson et~al\mbox{.}(2020)]%
        {erickson2020autogluon}
\bibfield{author}{\bibinfo{person}{Nick Erickson}, \bibinfo{person}{Jonas Mueller}, \bibinfo{person}{Alexander Shirkov}, \bibinfo{person}{Hang Zhang}, \bibinfo{person}{Pedro Larroy}, \bibinfo{person}{Mu Li}, {and} \bibinfo{person}{Alexander Smola}.} \bibinfo{year}{2020}\natexlab{}.
\newblock \bibinfo{title}{AutoGluon-Tabular: Robust and Accurate AutoML for Structured Data}.
\newblock
\showeprint[arxiv]{2003.06505}~[stat.ML]
\urldef\tempurl%
\url{https://arxiv.org/abs/2003.06505}
\showURL{%
\tempurl}


\bibitem[Evans et~al\mbox{.}(2022)]%
        {evans2022estimating}
\bibfield{author}{\bibinfo{person}{David~W. Evans}, \bibinfo{person}{Alison Rushton}, \bibinfo{person}{Nicola Middlebrook}, \bibinfo{person}{Jon Bishop}, \bibinfo{person}{Marco Barbero}, \bibinfo{person}{Jaimin Patel}, {and} \bibinfo{person}{Deborah Falla}.} \bibinfo{year}{2022}\natexlab{}.
\newblock \showarticletitle{Estimating risk of chronic pain and disability following musculoskeletal trauma in the United Kingdom}.
\newblock \bibinfo{journal}{\emph{JAMA network open}} \bibinfo{volume}{5}, \bibinfo{number}{8} (\bibinfo{year}{2022}), \bibinfo{pages}{e2228870--e2228870}.
\newblock
\href{https://doi.org/10.1001/jamanetworkopen.2022.28870}{doi:\nolinkurl{10.1001/jamanetworkopen.2022.28870}}


\bibitem[Frobell et~al\mbox{.}(2009)]%
        {frobell2009acutely}
\bibfield{author}{\bibinfo{person}{R.~B. Frobell}, \bibinfo{person}{M.~P. Le~Graverand}, \bibinfo{person}{R. Buck}, \bibinfo{person}{E.~M. Roos}, \bibinfo{person}{H.~P. Roos}, \bibinfo{person}{J. Tamez-Pena}, \bibinfo{person}{S. Totterman}, {and} \bibinfo{person}{L.~S. Lohmander}.} \bibinfo{year}{2009}\natexlab{}.
\newblock \showarticletitle{The acutely ACL injured knee assessed by MRI: changes in joint fluid, bone marrow lesions, and cartilage during the first year}.
\newblock \bibinfo{journal}{\emph{Osteoarthritis and cartilage}} \bibinfo{volume}{17}, \bibinfo{number}{2} (\bibinfo{year}{2009}), \bibinfo{pages}{161--167}.
\newblock
\href{https://doi.org/10.1016/j.joca.2008.06.020}{doi:\nolinkurl{10.1016/j.joca.2008.06.020}}


\bibitem[Fu et~al\mbox{.}(2024)]%
        {fu2024transformers}
\bibfield{author}{\bibinfo{person}{Deqing Fu}, \bibinfo{person}{Tian-Qi Chen}, \bibinfo{person}{Robin Jia}, {and} \bibinfo{person}{Vatsal Sharan}.} \bibinfo{year}{2024}\natexlab{}.
\newblock \bibinfo{title}{Transformers Learn Higher-Order Optimization Methods for In-Context Learning: A Study with Linear Models}.
\newblock
\showeprint[arxiv]{2310.17086}~[cs.LG]
\urldef\tempurl%
\url{https://arxiv.org/abs/2310.17086}
\showURL{%
\tempurl}


\bibitem[Gallifant et~al\mbox{.}(2024)]%
        {gallifant2024languagemodels}
\bibfield{author}{\bibinfo{person}{Jack Gallifant}, \bibinfo{person}{Shan Chen}, \bibinfo{person}{Pedro Jos{\'e}~Ferreira Moreira}, \bibinfo{person}{Nikolaj Munch}, \bibinfo{person}{Mingye Gao}, \bibinfo{person}{Jackson Pond}, \bibinfo{person}{Leo~Anthony Celi}, \bibinfo{person}{Hugo Aerts}, \bibinfo{person}{Thomas Hartvigsen}, {and} \bibinfo{person}{Danielle Bitterman}.} \bibinfo{year}{2024}\natexlab{}.
\newblock \showarticletitle{Language Models are Surprisingly Fragile to Drug Names in Biomedical Benchmarks}. In \bibinfo{booktitle}{\emph{Proceedings of the 2024 Conference on Empirical Methods in Natural Language Processing (EMNLP)}}. \bibinfo{publisher}{Association for Computational Linguistics (ACL)}, \bibinfo{address}{Miami, USA}, \bibinfo{pages}{12448--12465}.
\newblock


\bibitem[Garg et~al\mbox{.}(2022)]%
        {garg2022can}
\bibfield{author}{\bibinfo{person}{Shivam Garg}, \bibinfo{person}{Dimitris Tsipras}, \bibinfo{person}{Percy~S Liang}, {and} \bibinfo{person}{Gregory Valiant}.} \bibinfo{year}{2022}\natexlab{}.
\newblock \showarticletitle{What can transformers learn in-context? a case study of simple function classes}.
\newblock \bibinfo{journal}{\emph{Advances in neural information processing systems}}  \bibinfo{volume}{35} (\bibinfo{year}{2022}), \bibinfo{pages}{30583--30598}.
\newblock
\href{https://doi.org/10.48550/arXiv.2208.01066}{doi:\nolinkurl{10.48550/arXiv.2208.01066}}


\bibitem[Geurts et~al\mbox{.}(2006)]%
        {geurts2006extremely}
\bibfield{author}{\bibinfo{person}{Pierre Geurts}, \bibinfo{person}{Damien Ernst}, {and} \bibinfo{person}{Louis Wehenkel}.} \bibinfo{year}{2006}\natexlab{}.
\newblock \showarticletitle{Extremely randomized trees}.
\newblock \bibinfo{journal}{\emph{Machine learning}}  \bibinfo{volume}{63} (\bibinfo{year}{2006}), \bibinfo{pages}{3--42}.
\newblock
\href{https://doi.org/10.1007/s10994-006-6226-1}{doi:\nolinkurl{10.1007/s10994-006-6226-1}}


\bibitem[Google(2024a)]%
        {google2024sensitive}
\bibfield{author}{\bibinfo{person}{Google}.} \bibinfo{year}{2024}\natexlab{a}.
\newblock \bibinfo{title}{Cloud Data Processing Addendum}.
\newblock \bibinfo{howpublished}{https://cloud.google.com/terms/data-processing-addendum}.
\newblock
\newblock
\shownote{Accessed: 2024-07-01}.


\bibitem[Google(2024b)]%
        {google2024gemini}
\bibfield{author}{\bibinfo{person}{Google}.} \bibinfo{year}{2024}\natexlab{b}.
\newblock \bibinfo{title}{Gemini 1.5: Unlocking multimodal understanding across millions of tokens of context}.
\newblock
\showeprint[arxiv]{2403.05530}~[cs.CL]
\urldef\tempurl%
\url{https://arxiv.org/abs/2403.05530}
\showURL{%
\tempurl}


\bibitem[Google(2024c)]%
        {google2024gemini2}
\bibfield{author}{\bibinfo{person}{Google}.} \bibinfo{year}{2024}\natexlab{c}.
\newblock \bibinfo{title}{Introducing Gemini 2.0: our new AI model for the agentic era}.
\newblock \bibinfo{howpublished}{https://blog.google/technology/google-deepmind/google-gemini-ai-update-december-2024/}.
\newblock
\newblock
\shownote{Accessed: 2024-12-11}.


\bibitem[Grazzi et~al\mbox{.}(2024)]%
        {grazzi2024mamba}
\bibfield{author}{\bibinfo{person}{Riccardo Grazzi}, \bibinfo{person}{Julien~Niklas Siems}, \bibinfo{person}{Simon Schrodi}, \bibinfo{person}{Thomas Brox}, {and} \bibinfo{person}{Frank Hutter}.} \bibinfo{year}{2024}\natexlab{}.
\newblock \showarticletitle{Is Mamba Capable of In-Context Learning?}. In \bibinfo{booktitle}{\emph{Proceedings of the 3rd International Conference on Automated Machine Learning (AutoML)}}. \bibinfo{publisher}{PMLR}, \bibinfo{address}{Paris, France}, \bibinfo{pages}{1/1--26}.
\newblock


\bibitem[Grinsztajn et~al\mbox{.}(2022)]%
        {grinsztajn2022tree}
\bibfield{author}{\bibinfo{person}{L{\'e}o Grinsztajn}, \bibinfo{person}{Edouard Oyallon}, {and} \bibinfo{person}{Ga{\"e}l Varoquaux}.} \bibinfo{year}{2022}\natexlab{}.
\newblock \showarticletitle{Why do tree-based models still outperform deep learning on typical tabular data?}
\newblock \bibinfo{journal}{\emph{Advances in neural information processing systems}}  \bibinfo{volume}{35} (\bibinfo{year}{2022}), \bibinfo{pages}{507--520}.
\newblock
\href{https://doi.org/10.48550/arXiv.2207.08815}{doi:\nolinkurl{10.48550/arXiv.2207.08815}}


\bibitem[Halevy et~al\mbox{.}(2009)]%
        {halevy2009unreasonable}
\bibfield{author}{\bibinfo{person}{Alon Halevy}, \bibinfo{person}{Peter Norvig}, {and} \bibinfo{person}{Fernando Pereira}.} \bibinfo{year}{2009}\natexlab{}.
\newblock \showarticletitle{The unreasonable effectiveness of data}.
\newblock \bibinfo{journal}{\emph{IEEE intelligent systems}} \bibinfo{volume}{24}, \bibinfo{number}{2} (\bibinfo{year}{2009}), \bibinfo{pages}{8--12}.
\newblock
\href{https://doi.org/10.1109/MIS.2009.36}{doi:\nolinkurl{10.1109/MIS.2009.36}}


\bibitem[Han et~al\mbox{.}(2024)]%
        {han2024large}
\bibfield{author}{\bibinfo{person}{Sungwon Han}, \bibinfo{person}{Jinsung Yoon}, \bibinfo{person}{Sercan~Ö Arik}, {and} \bibinfo{person}{Tomas Pfister}.} \bibinfo{year}{2024}\natexlab{}.
\newblock \showarticletitle{Large language models can automatically engineer features for few-shot tabular learning}. In \bibinfo{booktitle}{\emph{Proceedings of the 41st International Conference on Machine Learning (ICML)}}. \bibinfo{publisher}{PMLR}, \bibinfo{address}{Vienna, Austria}, \bibinfo{pages}{17454--17479}.
\newblock


\bibitem[Harrell(2015)]%
        {harrell2015regression}
\bibfield{author}{\bibinfo{person}{Frank~E. Harrell}.} \bibinfo{year}{2015}\natexlab{}.
\newblock \bibinfo{booktitle}{\emph{Regression modeling strategies: with applications to linear models, logistic regression, and survival analysis}}.
\newblock \bibinfo{publisher}{Springer}.
\newblock


\bibitem[He and Ma(2013)]%
        {he2013imbalanced}
\bibfield{author}{\bibinfo{person}{Haibo He} {and} \bibinfo{person}{Yunqian Ma}.} \bibinfo{year}{2013}\natexlab{}.
\newblock \bibinfo{booktitle}{\emph{Imbalanced learning: foundations, algorithms, and applications}}.
\newblock \bibinfo{publisher}{John Wiley \& Sons}.
\newblock


\bibitem[Heinze et~al\mbox{.}(2018)]%
        {heinze2018variable}
\bibfield{author}{\bibinfo{person}{Georg Heinze}, \bibinfo{person}{Christine Wallisch}, {and} \bibinfo{person}{Daniela Dunkler}.} \bibinfo{year}{2018}\natexlab{}.
\newblock \showarticletitle{Variable selection--a review and recommendations for the practicing statistician}.
\newblock \bibinfo{journal}{\emph{Biometrical journal}} \bibinfo{volume}{60}, \bibinfo{number}{3} (\bibinfo{year}{2018}), \bibinfo{pages}{431--449}.
\newblock
\href{https://doi.org/10.1002/bimj.201700067}{doi:\nolinkurl{10.1002/bimj.201700067}}


\bibitem[Hollmann et~al\mbox{.}(2023)]%
        {hollmann2023tabpfn}
\bibfield{author}{\bibinfo{person}{Noah Hollmann}, \bibinfo{person}{Samuel M{\"u}ller}, \bibinfo{person}{Katharina Eggensperger}, {and} \bibinfo{person}{Frank Hutter}.} \bibinfo{year}{2023}\natexlab{}.
\newblock \showarticletitle{Tabpfn: A transformer that solves small tabular classification problems in a second}. In \bibinfo{booktitle}{\emph{Proceedings of the 11th International Conference on Learning Representations (ICLR)}}. \bibinfo{publisher}{International Conference on Learning Representations (ICLR)}, \bibinfo{address}{Kigali, Rwanda}.
\newblock


\bibitem[Hollmann et~al\mbox{.}(2025)]%
        {hollmann2025accurate}
\bibfield{author}{\bibinfo{person}{Noah Hollmann}, \bibinfo{person}{Samuel M{\"u}ller}, \bibinfo{person}{Lennart Purucker}, \bibinfo{person}{Arjun Krishnakumar}, \bibinfo{person}{Max K{\"o}rfer}, \bibinfo{person}{Shi~Bin Hoo}, \bibinfo{person}{Robin~Tibor Schirrmeister}, {and} \bibinfo{person}{Frank Hutter}.} \bibinfo{year}{2025}\natexlab{}.
\newblock \showarticletitle{Accurate predictions on small data with a tabular foundation model}.
\newblock \bibinfo{journal}{\emph{Nature}} \bibinfo{volume}{637}, \bibinfo{number}{8045} (\bibinfo{year}{2025}), \bibinfo{pages}{319--326}.
\newblock


\bibitem[Howard and Ruder(2018)]%
        {howard2018universal}
\bibfield{author}{\bibinfo{person}{Jeremy Howard} {and} \bibinfo{person}{Sebastian Ruder}.} \bibinfo{year}{2018}\natexlab{}.
\newblock \showarticletitle{Universal Language Model Fine-tuning for Text Classification}. In \bibinfo{booktitle}{\emph{Proceedings of the 56th Annual Meeting of the Association for Computational Linguistics}}. \bibinfo{publisher}{Association for Computational Linguistics (ACL)}, \bibinfo{address}{Melbourne, Australia}, \bibinfo{pages}{328--339}.
\newblock


\bibitem[Imrie et~al\mbox{.}(2023)]%
        {imrie2023autoprognosis}
\bibfield{author}{\bibinfo{person}{Fergus Imrie}, \bibinfo{person}{Bogdan Cebere}, \bibinfo{person}{Eoin~F McKinney}, {and} \bibinfo{person}{Mihaela van~der Schaar}.} \bibinfo{year}{2023}\natexlab{}.
\newblock \showarticletitle{AutoPrognosis 2.0: Democratizing diagnostic and prognostic modeling in healthcare with automated machine learning}.
\newblock \bibinfo{journal}{\emph{PLOS Digital Health}} \bibinfo{volume}{2}, \bibinfo{number}{6} (\bibinfo{year}{2023}), \bibinfo{pages}{e0000276}.
\newblock
\href{https://doi.org/10.1371/journal.pdig.0000276}{doi:\nolinkurl{10.1371/journal.pdig.0000276}}


\bibitem[Iriondo et~al\mbox{.}(2021)]%
        {iriondo2021towards}
\bibfield{author}{\bibinfo{person}{Claudia Iriondo}, \bibinfo{person}{Felix Liu}, \bibinfo{person}{Francesco Caliv{\`a}}, \bibinfo{person}{Sarthak Kamat}, \bibinfo{person}{Sharmila Majumdar}, {and} \bibinfo{person}{Valentina Pedoia}.} \bibinfo{year}{2021}\natexlab{}.
\newblock \showarticletitle{Towards understanding mechanistic subgroups of osteoarthritis: 8-year cartilage thickness trajectory analysis}.
\newblock \bibinfo{journal}{\emph{Journal of Orthopaedic Research{\textregistered}}} \bibinfo{volume}{39}, \bibinfo{number}{6} (\bibinfo{year}{2021}), \bibinfo{pages}{1305--1317}.
\newblock
\href{https://doi.org/10.1002/jor.24849}{doi:\nolinkurl{10.1002/jor.24849}}


\bibitem[Kaplan(2011)]%
        {kaplan2011identifying}
\bibfield{author}{\bibinfo{person}{Yonatan Kaplan}.} \bibinfo{year}{2011}\natexlab{}.
\newblock \showarticletitle{Identifying individuals with an anterior cruciate ligament-deficient knee as copers and noncopers: a narrative literature review}.
\newblock \bibinfo{journal}{\emph{Journal of orthopaedic \& sports physical therapy}} \bibinfo{volume}{41}, \bibinfo{number}{10} (\bibinfo{year}{2011}), \bibinfo{pages}{758--766}.
\newblock
\href{https://doi.org/10.2519/jospt.2011.3384}{doi:\nolinkurl{10.2519/jospt.2011.3384}}


\bibitem[Knauer et~al\mbox{.}(2024)]%
        {knauer2024pmlbmini}
\bibfield{author}{\bibinfo{person}{Ricardo Knauer}, \bibinfo{person}{Marvin Grimm}, {and} \bibinfo{person}{Erik Rodner}.} \bibinfo{year}{2024}\natexlab{}.
\newblock \bibinfo{title}{PMLBmini: A Tabular Classification Benchmark Suite for Data-Scarce Applications}.
\newblock
\showeprint[arxiv]{2409.01635}~[cs.LG]
\urldef\tempurl%
\url{https://arxiv.org/abs/2409.01635}
\showURL{%
\tempurl}


\bibitem[Knauer and Rodner(2023)]%
        {knauer2023cost}
\bibfield{author}{\bibinfo{person}{Ricardo Knauer} {and} \bibinfo{person}{Erik Rodner}.} \bibinfo{year}{2023}\natexlab{}.
\newblock \showarticletitle{Cost-Sensitive Best Subset Selection for Logistic Regression: A Mixed-Integer Conic Optimization Perspective}. In \bibinfo{booktitle}{\emph{Proceedings of the 46th German Conference on Artificial Intelligence (K{\"u}nstliche Intelligenz)}}. \bibinfo{publisher}{Springer}, \bibinfo{address}{Berlin, Germany}, \bibinfo{pages}{114--129}.
\newblock


\bibitem[Knauer and Rodner(2024)]%
        {knauer2024squeezing}
\bibfield{author}{\bibinfo{person}{Ricardo Knauer} {and} \bibinfo{person}{Erik Rodner}.} \bibinfo{year}{2024}\natexlab{}.
\newblock \bibinfo{title}{Squeezing Lemons with Hammers: An Evaluation of AutoML and Tabular Deep Learning for Data-Scarce Classification Applications}.
\newblock
\showeprint[arxiv]{2405.07662}~[cs.LG]
\urldef\tempurl%
\url{https://arxiv.org/abs/2405.07662}
\showURL{%
\tempurl}


\bibitem[Kojima et~al\mbox{.}(2022)]%
        {kojima2022large}
\bibfield{author}{\bibinfo{person}{Takeshi Kojima}, \bibinfo{person}{Shixiang~Shane Gu}, \bibinfo{person}{Machel Reid}, \bibinfo{person}{Yutaka Matsuo}, {and} \bibinfo{person}{Yusuke Iwasawa}.} \bibinfo{year}{2022}\natexlab{}.
\newblock \showarticletitle{Large language models are zero-shot reasoners}.
\newblock \bibinfo{journal}{\emph{Advances in neural information processing systems}}  \bibinfo{volume}{35} (\bibinfo{year}{2022}), \bibinfo{pages}{22199--22213}.
\newblock
\href{https://doi.org/10.48550/arXiv.2205.11916}{doi:\nolinkurl{10.48550/arXiv.2205.11916}}


\bibitem[Lee et~al\mbox{.}(2025)]%
        {lee2025evolvingdeeperllmthinking}
\bibfield{author}{\bibinfo{person}{Kuang-Huei Lee}, \bibinfo{person}{Ian Fischer}, \bibinfo{person}{Yueh-Hua Wu}, \bibinfo{person}{Dave Marwood}, \bibinfo{person}{Shumeet Baluja}, \bibinfo{person}{Dale Schuurmans}, {and} \bibinfo{person}{Xinyun Chen}.} \bibinfo{year}{2025}\natexlab{}.
\newblock \bibinfo{title}{Evolving Deeper LLM Thinking}.
\newblock
\showeprint[arxiv]{2501.09891}~[cs.AI]
\urldef\tempurl%
\url{https://arxiv.org/abs/2501.09891}
\showURL{%
\tempurl}


\bibitem[Li et~al\mbox{.}(2023)]%
        {li2023tree}
\bibfield{author}{\bibinfo{person}{Qinbin Li}, \bibinfo{person}{Yesheng Liang}, \bibinfo{person}{Yiqun Diao}, \bibinfo{person}{Chulin Xie}, \bibinfo{person}{Bo Li}, \bibinfo{person}{Bingsheng He}, {and} \bibinfo{person}{Dawn Song}.} \bibinfo{year}{2023}\natexlab{}.
\newblock \bibinfo{title}{Tree-as-a-prompt: boosting black-box large language models on few-shot classification of tabular data}.
\newblock \bibinfo{howpublished}{https://openreview.net/forum?id=SJTSvRtGsN}.
\newblock
\newblock
\shownote{Accessed: 2024-07-01}.


\bibitem[Liu et~al\mbox{.}(2025)]%
        {liu2025decision}
\bibfield{author}{\bibinfo{person}{Tennison Liu}, \bibinfo{person}{Nicolas Huynh}, {and} \bibinfo{person}{Mihaela van~der Schaar}.} \bibinfo{year}{2025}\natexlab{}.
\newblock \showarticletitle{Decision Tree Induction Through LLMs via Semantically-Aware Evolution}. In \bibinfo{booktitle}{\emph{Proceedings of the 13th International Conference on Learning Representations (ICLR)}}. \bibinfo{publisher}{International Conference on Learning Representations (ICLR)}, \bibinfo{address}{Singapore}.
\newblock


\bibitem[Longo et~al\mbox{.}(2024)]%
        {longo2024explainable}
\bibfield{author}{\bibinfo{person}{Luca Longo}, \bibinfo{person}{Mario Brcic}, \bibinfo{person}{Federico Cabitza}, \bibinfo{person}{Jaesik Choi}, \bibinfo{person}{Roberto Confalonieri}, \bibinfo{person}{Javier Del~Ser}, \bibinfo{person}{Riccardo Guidotti}, \bibinfo{person}{Yoichi Hayashi}, \bibinfo{person}{Francisco Herrera}, \bibinfo{person}{Andreas Holzinger}, {et~al\mbox{.}}} \bibinfo{year}{2024}\natexlab{}.
\newblock \showarticletitle{Explainable Artificial Intelligence (XAI) 2.0: A manifesto of open challenges and interdisciplinary research directions}.
\newblock \bibinfo{journal}{\emph{Information Fusion}}  \bibinfo{volume}{106} (\bibinfo{year}{2024}), \bibinfo{pages}{102301}.
\newblock
\href{https://doi.org/10.1016/j.inffus.2024.102301}{doi:\nolinkurl{10.1016/j.inffus.2024.102301}}


\bibitem[Matias and Gupta(2023)]%
        {matias2023medlm}
\bibfield{author}{\bibinfo{person}{Yossi Matias} {and} \bibinfo{person}{Aashima Gupta}.} \bibinfo{year}{2023}\natexlab{}.
\newblock \bibinfo{title}{MedLM: generative AI fine-tuned for the healthcare industry}.
\newblock \bibinfo{howpublished}{https://cloud.google.com/blog/topics/healthcare-life-sciences/introducing-medlm-for-the-healthcare-industry}.
\newblock
\newblock
\shownote{Accessed: 2024-07-01}.


\bibitem[McElfresh et~al\mbox{.}(2023)]%
        {mcelfresh2023neural}
\bibfield{author}{\bibinfo{person}{Duncan McElfresh}, \bibinfo{person}{Sujay Khandagale}, \bibinfo{person}{Jonathan Valverde}, \bibinfo{person}{Vishak Prasad~C}, \bibinfo{person}{Ganesh Ramakrishnan}, \bibinfo{person}{Micah Goldblum}, {and} \bibinfo{person}{Colin White}.} \bibinfo{year}{2023}\natexlab{}.
\newblock \showarticletitle{When do neural nets outperform boosted trees on tabular data?}
\newblock \bibinfo{journal}{\emph{Advances in neural information processing systems}}  \bibinfo{volume}{36} (\bibinfo{year}{2023}).
\newblock
\href{https://doi.org/10.48550/arXiv.2305.02997}{doi:\nolinkurl{10.48550/arXiv.2305.02997}}


\bibitem[Molnar(2022)]%
        {molnar2022}
\bibfield{author}{\bibinfo{person}{Christoph Molnar}.} \bibinfo{year}{2022}\natexlab{}.
\newblock \bibinfo{booktitle}{\emph{Interpretable Machine Learning} (\bibinfo{edition}{2} ed.)}.
\newblock \bibinfo{publisher}{Lulu.com}.
\newblock
\urldef\tempurl%
\url{https://christophm.github.io/interpretable-ml-book}
\showURL{%
\tempurl}


\bibitem[Moons et~al\mbox{.}(2015)]%
        {moons2015transparent}
\bibfield{author}{\bibinfo{person}{Karel G.~M. Moons}, \bibinfo{person}{Douglas~G. Altman}, \bibinfo{person}{Johannes~B. Reitsma}, \bibinfo{person}{John P.~A. Ioannidis}, \bibinfo{person}{Petra Macaskill}, \bibinfo{person}{Ewout~W. Steyerberg}, \bibinfo{person}{Andrew~J. Vickers}, \bibinfo{person}{David~F. Ransohoff}, {and} \bibinfo{person}{Gary~S. Collins}.} \bibinfo{year}{2015}\natexlab{}.
\newblock \showarticletitle{Transparent Reporting of a multivariable prediction model for Individual Prognosis or Diagnosis (TRIPOD): explanation and elaboration}.
\newblock \bibinfo{journal}{\emph{Annals of internal medicine}} \bibinfo{volume}{162}, \bibinfo{number}{1} (\bibinfo{year}{2015}), \bibinfo{pages}{W1--W73}.
\newblock
\href{https://doi.org/10.7326/M14-0698}{doi:\nolinkurl{10.7326/M14-0698}}


\bibitem[Moons et~al\mbox{.}(2019)]%
        {moons2019probast}
\bibfield{author}{\bibinfo{person}{Karel G.~M. Moons}, \bibinfo{person}{Robert~F. Wolff}, \bibinfo{person}{Richard~D. Riley}, \bibinfo{person}{Penny~F. Whiting}, \bibinfo{person}{Marie Westwood}, \bibinfo{person}{Gary~S. Collins}, \bibinfo{person}{Johannes~B. Reitsma}, \bibinfo{person}{Jos Kleijnen}, {and} \bibinfo{person}{Sue Mallett}.} \bibinfo{year}{2019}\natexlab{}.
\newblock \showarticletitle{PROBAST: a tool to assess risk of bias and applicability of prediction model studies: explanation and elaboration}.
\newblock \bibinfo{journal}{\emph{Annals of internal medicine}} \bibinfo{volume}{170}, \bibinfo{number}{1} (\bibinfo{year}{2019}), \bibinfo{pages}{W1--W33}.
\newblock
\href{https://doi.org/10.7326/M18-1377}{doi:\nolinkurl{10.7326/M18-1377}}


\bibitem[Moosmann et~al\mbox{.}(2006)]%
        {moosmann2006fast}
\bibfield{author}{\bibinfo{person}{Frank Moosmann}, \bibinfo{person}{Bill Triggs}, {and} \bibinfo{person}{Frederic Jurie}.} \bibinfo{year}{2006}\natexlab{}.
\newblock \showarticletitle{Fast discriminative visual codebooks using randomized clustering forests}.
\newblock \bibinfo{journal}{\emph{Advances in neural information processing systems}}  \bibinfo{volume}{19} (\bibinfo{year}{2006}).
\newblock


\bibitem[Nam et~al\mbox{.}(2024)]%
        {nam2024featuregeneration}
\bibfield{author}{\bibinfo{person}{Jaehyun Nam}, \bibinfo{person}{Kyuyoung Kim}, \bibinfo{person}{Seunghyuk Oh}, \bibinfo{person}{Jihoon Tack}, \bibinfo{person}{Jaehyung Kim}, {and} \bibinfo{person}{Jinwoo Shin}.} \bibinfo{year}{2024}\natexlab{}.
\newblock \showarticletitle{Optimized Feature Generation for Tabular Data via LLMs with Decision Tree Reasoning}.
\newblock \bibinfo{journal}{\emph{Advances in neural information processing systems}}  \bibinfo{volume}{37} (\bibinfo{year}{2024}).
\newblock
\href{https://doi.org/10.48550/arXiv.2406.08527}{doi:\nolinkurl{10.48550/arXiv.2406.08527}}


\bibitem[Nori et~al\mbox{.}(2023)]%
        {nori2023generalist}
\bibfield{author}{\bibinfo{person}{Harsha Nori}, \bibinfo{person}{Yin~Tat Lee}, \bibinfo{person}{Sheng Zhang}, \bibinfo{person}{Dean Carignan}, \bibinfo{person}{Richard Edgar}, \bibinfo{person}{Nicolo Fusi}, \bibinfo{person}{Nicholas King}, \bibinfo{person}{Jonathan Larson}, \bibinfo{person}{Yuanzhi Li}, \bibinfo{person}{Weishung Liu}, \bibinfo{person}{Renqian Luo}, \bibinfo{person}{Scott~Mayer McKinney}, \bibinfo{person}{Robert~Osazuwa Ness}, \bibinfo{person}{Hoifung Poon}, \bibinfo{person}{Tao Qin}, \bibinfo{person}{Naoto Usuyama}, \bibinfo{person}{Chris White}, {and} \bibinfo{person}{Eric Horvitz}.} \bibinfo{year}{2023}\natexlab{}.
\newblock \bibinfo{title}{Can Generalist Foundation Models Outcompete Special-Purpose Tuning? Case Study in Medicine}.
\newblock
\showeprint[arxiv]{2311.16452}~[cs.CL]
\urldef\tempurl%
\url{https://arxiv.org/abs/2311.16452}
\showURL{%
\tempurl}


\bibitem[Olson et~al\mbox{.}(2017)]%
        {olson2017pmlb}
\bibfield{author}{\bibinfo{person}{Randal~S. Olson}, \bibinfo{person}{William La~Cava}, \bibinfo{person}{Patryk Orzechowski}, \bibinfo{person}{Ryan~J. Urbanowicz}, {and} \bibinfo{person}{Jason~H. Moore}.} \bibinfo{year}{2017}\natexlab{}.
\newblock \showarticletitle{PMLB: a large benchmark suite for machine learning evaluation and comparison}.
\newblock \bibinfo{journal}{\emph{BioData mining}}  \bibinfo{volume}{10} (\bibinfo{year}{2017}), \bibinfo{pages}{1--13}.
\newblock
\href{https://doi.org/10.1186/s13040-017-0154-4}{doi:\nolinkurl{10.1186/s13040-017-0154-4}}


\bibitem[OpenAI(2024a)]%
        {openai2024sensitive}
\bibfield{author}{\bibinfo{person}{OpenAI}.} \bibinfo{year}{2024}\natexlab{a}.
\newblock \bibinfo{title}{Data Processing Addendum}.
\newblock \bibinfo{howpublished}{https://openai.com/policies/data-processing-addendum/}.
\newblock
\newblock
\shownote{Accessed: 2024-07-01}.


\bibitem[OpenAI(2024b)]%
        {openai2024gpt4o}
\bibfield{author}{\bibinfo{person}{OpenAI}.} \bibinfo{year}{2024}\natexlab{b}.
\newblock \bibinfo{title}{GPT-4o System Card}.
\newblock \bibinfo{howpublished}{https://cdn.openai.com/gpt-4o-system-card.pdf}.
\newblock
\newblock
\shownote{Accessed: 2024-08-08}.


\bibitem[OpenAI(2024c)]%
        {openai2024pro}
\bibfield{author}{\bibinfo{person}{OpenAI}.} \bibinfo{year}{2024}\natexlab{c}.
\newblock \bibinfo{title}{Introducing ChatGPT Pro}.
\newblock \bibinfo{howpublished}{https://openai.com/index/introducing-chatgpt-pro/}.
\newblock
\newblock
\shownote{Accessed: 2024-12-05}.


\bibitem[OpenAI(2024d)]%
        {openai2024reasoning}
\bibfield{author}{\bibinfo{person}{OpenAI}.} \bibinfo{year}{2024}\natexlab{d}.
\newblock \bibinfo{title}{OpenAI o1 System Card}.
\newblock \bibinfo{howpublished}{https://cdn.openai.com/o1-system-card-20241205.pdf}.
\newblock
\newblock
\shownote{Accessed: 2024-12-05}.


\bibitem[Pal et~al\mbox{.}(2024)]%
        {pal2024medicalllm}
\bibfield{author}{\bibinfo{person}{Ankit Pal}, \bibinfo{person}{Pasquale Minervini}, \bibinfo{person}{Andreas~Geert Motzfeldt}, \bibinfo{person}{Aryo~Pradipta Gema}, {and} \bibinfo{person}{Beatrice Alex}.} \bibinfo{year}{2024}\natexlab{}.
\newblock \bibinfo{title}{The Open Medical-LLM Leaderboard: Benchmarking Large Language Models in Healthcare}.
\newblock \bibinfo{howpublished}{https://huggingface.co/blog/leaderboard-medicalllm}.
\newblock
\newblock
\shownote{Accessed: 2024-07-01}.


\bibitem[Pan and Yang(2009)]%
        {pan2009survey}
\bibfield{author}{\bibinfo{person}{Sinno~Jialin Pan} {and} \bibinfo{person}{Qiang Yang}.} \bibinfo{year}{2009}\natexlab{}.
\newblock \showarticletitle{A survey on transfer learning}.
\newblock \bibinfo{journal}{\emph{IEEE Transactions on knowledge and data engineering}} \bibinfo{volume}{22}, \bibinfo{number}{10} (\bibinfo{year}{2009}), \bibinfo{pages}{1345--1359}.
\newblock
\href{https://doi.org/10.1109/TKDE.2009.191}{doi:\nolinkurl{10.1109/TKDE.2009.191}}


\bibitem[Peters et~al\mbox{.}(2018)]%
        {peters2018contextualized}
\bibfield{author}{\bibinfo{person}{Matthew~E. Peters}, \bibinfo{person}{Mark Neumann}, \bibinfo{person}{Mohit Iyyer}, \bibinfo{person}{Matt Gardner}, \bibinfo{person}{Christopher Clark}, \bibinfo{person}{Kenton Lee}, {and} \bibinfo{person}{Luke Zettlemoyer}.} \bibinfo{year}{2018}\natexlab{}.
\newblock \showarticletitle{Deep Contextualized Word Representations}. In \bibinfo{booktitle}{\emph{Proceedings of the 2018 Conference of the North {A}merican Chapter of the Association for Computational Linguistics (NAACL)}}. \bibinfo{publisher}{Association for Computational Linguistics (ACL)}, \bibinfo{address}{New Orleans, USA}, \bibinfo{pages}{2227--2237}.
\newblock


\bibitem[Ribeiro et~al\mbox{.}(2016)]%
        {ribeiro2016should}
\bibfield{author}{\bibinfo{person}{Marco~Tulio Ribeiro}, \bibinfo{person}{Sameer Singh}, {and} \bibinfo{person}{Carlos Guestrin}.} \bibinfo{year}{2016}\natexlab{}.
\newblock \showarticletitle{"Why should I trust you?" Explaining the predictions of any classifier}. In \bibinfo{booktitle}{\emph{Proceedings of the 22nd ACM SIGKDD international conference on knowledge discovery and data mining}}. \bibinfo{publisher}{Association for Computing Machinery (ACM)}, \bibinfo{address}{San Francisco, USA}, \bibinfo{pages}{1135--1144}.
\newblock


\bibitem[Ribeiro et~al\mbox{.}(2018)]%
        {ribeiro2018anchors}
\bibfield{author}{\bibinfo{person}{Marco~Tulio Ribeiro}, \bibinfo{person}{Sameer Singh}, {and} \bibinfo{person}{Carlos Guestrin}.} \bibinfo{year}{2018}\natexlab{}.
\newblock \showarticletitle{Anchors: High-precision model-agnostic explanations}. In \bibinfo{booktitle}{\emph{Proceedings of the 32nd AAAI conference on artificial intelligence}}. \bibinfo{publisher}{Association for the Advancement of Artificial Intelligence (AAAI)}, \bibinfo{address}{New Orleans, USA}, \bibinfo{pages}{1527--1535}.
\newblock


\bibitem[Romano et~al\mbox{.}(2022)]%
        {romano2022pmlb}
\bibfield{author}{\bibinfo{person}{Joseph~D Romano}, \bibinfo{person}{Trang~T Le}, \bibinfo{person}{William La~Cava}, \bibinfo{person}{John~T Gregg}, \bibinfo{person}{Daniel~J Goldberg}, \bibinfo{person}{Praneel Chakraborty}, \bibinfo{person}{Natasha~L Ray}, \bibinfo{person}{Daniel Himmelstein}, \bibinfo{person}{Weixuan Fu}, {and} \bibinfo{person}{Jason~H Moore}.} \bibinfo{year}{2022}\natexlab{}.
\newblock \showarticletitle{PMLB v1. 0: an open-source dataset collection for benchmarking machine learning methods}.
\newblock \bibinfo{journal}{\emph{Bioinformatics}} \bibinfo{volume}{38}, \bibinfo{number}{3} (\bibinfo{year}{2022}), \bibinfo{pages}{878--880}.
\newblock
\href{https://doi.org/10.1093/bioinformatics/btab727}{doi:\nolinkurl{10.1093/bioinformatics/btab727}}


\bibitem[Ryan et~al\mbox{.}(2018)]%
        {ryan2018handbook}
\bibfield{author}{\bibinfo{person}{Conor Ryan}, \bibinfo{person}{Michael O'Neill}, {and} \bibinfo{person}{J.~J. Collins}.} \bibinfo{year}{2018}\natexlab{}.
\newblock \bibinfo{booktitle}{\emph{Handbook of grammatical evolution}}. Vol.~\bibinfo{volume}{1}.
\newblock \bibinfo{publisher}{Springer}.
\newblock


\bibitem[Saab et~al\mbox{.}(2024)]%
        {saab2024capabilities}
\bibfield{author}{\bibinfo{person}{Khaled Saab}, \bibinfo{person}{Tao Tu}, \bibinfo{person}{Wei-Hung Weng}, \bibinfo{person}{Ryutaro Tanno}, \bibinfo{person}{David Stutz}, \bibinfo{person}{Ellery Wulczyn}, \bibinfo{person}{Fan Zhang}, \bibinfo{person}{Tim Strother}, \bibinfo{person}{Chunjong Park}, \bibinfo{person}{Elahe Vedadi}, \bibinfo{person}{Juanma~Zambrano Chaves}, \bibinfo{person}{Szu-Yeu Hu}, \bibinfo{person}{Mike Schaekermann}, \bibinfo{person}{Aishwarya Kamath}, \bibinfo{person}{Yong Cheng}, \bibinfo{person}{David G.~T. Barrett}, \bibinfo{person}{Cathy Cheung}, \bibinfo{person}{Basil Mustafa}, \bibinfo{person}{Anil Palepu}, \bibinfo{person}{Daniel McDuff}, \bibinfo{person}{Le Hou}, \bibinfo{person}{Tomer Golany}, \bibinfo{person}{Luyang Liu}, \bibinfo{person}{Jean baptiste Alayrac}, \bibinfo{person}{Neil Houlsby}, \bibinfo{person}{Nenad Tomasev}, \bibinfo{person}{Jan Freyberg}, \bibinfo{person}{Charles Lau}, \bibinfo{person}{Jonas Kemp}, \bibinfo{person}{Jeremy Lai}, \bibinfo{person}{Shekoofeh
  Azizi}, \bibinfo{person}{Kimberly Kanada}, \bibinfo{person}{SiWai Man}, \bibinfo{person}{Kavita Kulkarni}, \bibinfo{person}{Ruoxi Sun}, \bibinfo{person}{Siamak Shakeri}, \bibinfo{person}{Luheng He}, \bibinfo{person}{Ben Caine}, \bibinfo{person}{Albert Webson}, \bibinfo{person}{Natasha Latysheva}, \bibinfo{person}{Melvin Johnson}, \bibinfo{person}{Philip Mansfield}, \bibinfo{person}{Jian Lu}, \bibinfo{person}{Ehud Rivlin}, \bibinfo{person}{Jesper Anderson}, \bibinfo{person}{Bradley Green}, \bibinfo{person}{Renee Wong}, \bibinfo{person}{Jonathan Krause}, \bibinfo{person}{Jonathon Shlens}, \bibinfo{person}{Ewa Dominowska}, \bibinfo{person}{S.~M.~Ali Eslami}, \bibinfo{person}{Katherine Chou}, \bibinfo{person}{Claire Cui}, \bibinfo{person}{Oriol Vinyals}, \bibinfo{person}{Koray Kavukcuoglu}, \bibinfo{person}{James Manyika}, \bibinfo{person}{Jeff Dean}, \bibinfo{person}{Demis Hassabis}, \bibinfo{person}{Yossi Matias}, \bibinfo{person}{Dale Webster}, \bibinfo{person}{Joelle Barral}, \bibinfo{person}{Greg Corrado},
  \bibinfo{person}{Christopher Semturs}, \bibinfo{person}{S.~Sara Mahdavi}, \bibinfo{person}{Juraj Gottweis}, \bibinfo{person}{Alan Karthikesalingam}, {and} \bibinfo{person}{Vivek Natarajan}.} \bibinfo{year}{2024}\natexlab{}.
\newblock \bibinfo{title}{Capabilities of Gemini Models in Medicine}.
\newblock
\showeprint[arxiv]{2404.18416}~[cs.AI]
\urldef\tempurl%
\url{https://arxiv.org/abs/2404.18416}
\showURL{%
\tempurl}


\bibitem[Salinas and Erickson(2024)]%
        {salinas2024tabrepo}
\bibfield{author}{\bibinfo{person}{David Salinas} {and} \bibinfo{person}{Nick Erickson}.} \bibinfo{year}{2024}\natexlab{}.
\newblock \showarticletitle{TabRepo: A Large Scale Repository of Tabular Model Evaluations and its AutoML Applications}. In \bibinfo{booktitle}{\emph{Proceedings of the 3rd International Conference on Automated Machine Learning (AutoML)}}. \bibinfo{publisher}{PMLR}, \bibinfo{address}{Paris, France}, \bibinfo{pages}{19/1--30}.
\newblock


\bibitem[Schulhoff et~al\mbox{.}(2024)]%
        {schulhoff2024promptreport}
\bibfield{author}{\bibinfo{person}{Sander Schulhoff}, \bibinfo{person}{Michael Ilie}, \bibinfo{person}{Nishant Balepur}, \bibinfo{person}{Konstantine Kahadze}, \bibinfo{person}{Amanda Liu}, \bibinfo{person}{Chenglei Si}, \bibinfo{person}{Yinheng Li}, \bibinfo{person}{Aayush Gupta}, \bibinfo{person}{HyoJung Han}, \bibinfo{person}{Sevien Schulhoff}, \bibinfo{person}{Pranav~Sandeep Dulepet}, \bibinfo{person}{Saurav Vidyadhara}, \bibinfo{person}{Dayeon Ki}, \bibinfo{person}{Sweta Agrawal}, \bibinfo{person}{Chau Pham}, \bibinfo{person}{Gerson Kroiz}, \bibinfo{person}{Feileen Li}, \bibinfo{person}{Hudson Tao}, \bibinfo{person}{Ashay Srivastava}, \bibinfo{person}{Hevander~Da Costa}, \bibinfo{person}{Saloni Gupta}, \bibinfo{person}{Megan~L. Rogers}, \bibinfo{person}{Inna Goncearenco}, \bibinfo{person}{Giuseppe Sarli}, \bibinfo{person}{Igor Galynker}, \bibinfo{person}{Denis Peskoff}, \bibinfo{person}{Marine Carpuat}, \bibinfo{person}{Jules White}, \bibinfo{person}{Shyamal Anadkat}, \bibinfo{person}{Alexander Hoyle},
  {and} \bibinfo{person}{Philip Resnik}.} \bibinfo{year}{2024}\natexlab{}.
\newblock \bibinfo{title}{The Prompt Report: A Systematic Survey of Prompting Techniques}.
\newblock
\showeprint[arxiv]{2406.06608}~[cs.CL]
\urldef\tempurl%
\url{https://arxiv.org/abs/2406.06608}
\showURL{%
\tempurl}


\bibitem[Shen et~al\mbox{.}(2024)]%
        {shen2024pretrained}
\bibfield{author}{\bibinfo{person}{Lingfeng Shen}, \bibinfo{person}{Aayush Mishra}, {and} \bibinfo{person}{Daniel Khashabi}.} \bibinfo{year}{2024}\natexlab{}.
\newblock \showarticletitle{Do pretrained Transformers Learn In-Context by Gradient Descent?}. In \bibinfo{booktitle}{\emph{Proceedings of the 41st International Conference on Machine Learning (ICML)}}. \bibinfo{publisher}{PMLR}, \bibinfo{address}{Vienna, Austria}, \bibinfo{pages}{44712--44740}.
\newblock


\bibitem[Simonoff(2003)]%
        {simonoff2003analyzing}
\bibfield{author}{\bibinfo{person}{Jeffrey~S Simonoff}.} \bibinfo{year}{2003}\natexlab{}.
\newblock \bibinfo{booktitle}{\emph{Analyzing categorical data}}. Vol.~\bibinfo{volume}{496}.
\newblock \bibinfo{publisher}{Springer}.
\newblock


\bibitem[Snell et~al\mbox{.}(2025)]%
        {snell2025testtimecompute}
\bibfield{author}{\bibinfo{person}{Charlie Snell}, \bibinfo{person}{Jaehoon Lee}, \bibinfo{person}{Kelvin Xu}, {and} \bibinfo{person}{Aviral Kumar}.} \bibinfo{year}{2025}\natexlab{}.
\newblock \showarticletitle{Scaling LLM Test-Time Compute Optimally Can be More Effective than Scaling Parameters for Reasoning}. In \bibinfo{booktitle}{\emph{Proceedings of the 13th International Conference on Learning Representations (ICLR)}}. \bibinfo{publisher}{International Conference on Learning Representations (ICLR)}, \bibinfo{address}{Singapore}.
\newblock


\bibitem[Steyerberg(2019)]%
        {steyerberg2019clinical}
\bibfield{author}{\bibinfo{person}{Ewout~W. Steyerberg}.} \bibinfo{year}{2019}\natexlab{}.
\newblock \bibinfo{booktitle}{\emph{Clinical prediction models: a practical approach to development, validation, and updating}}.
\newblock \bibinfo{publisher}{Springer}.
\newblock


\bibitem[Sun et~al\mbox{.}(2022)]%
        {sun2022black}
\bibfield{author}{\bibinfo{person}{Tianxiang Sun}, \bibinfo{person}{Yunfan Shao}, \bibinfo{person}{Hong Qian}, \bibinfo{person}{Xuanjing Huang}, {and} \bibinfo{person}{Xipeng Qiu}.} \bibinfo{year}{2022}\natexlab{}.
\newblock \showarticletitle{Black-box tuning for language-model-as-a-service}. In \bibinfo{booktitle}{\emph{Proceedings of the 39th International Conference on Machine Learning (ICML)}}. \bibinfo{publisher}{PMLR}, \bibinfo{address}{Baltimore, USA}, \bibinfo{pages}{20841--20855}.
\newblock


\bibitem[Tam et~al\mbox{.}(2024)]%
        {tam2024letspeakfreely}
\bibfield{author}{\bibinfo{person}{Zhi~Rui Tam}, \bibinfo{person}{Cheng-Kuang Wu}, \bibinfo{person}{Yi-Lin Tsai}, \bibinfo{person}{Chieh-Yen Lin}, \bibinfo{person}{Hung yi Lee}, {and} \bibinfo{person}{Yun-Nung Chen}.} \bibinfo{year}{2024}\natexlab{}.
\newblock \showarticletitle{Let Me Speak Freely? A Study On The Impact Of Format Restrictions On Large Language Model Performance}. In \bibinfo{booktitle}{\emph{Proceedings of the 2024 Conference on Empirical Methods in Natural Language Processing (EMNLP)}}. \bibinfo{publisher}{Association for Computational Linguistics (ACL)}, \bibinfo{address}{Miami, USA}, \bibinfo{pages}{1218--1236}.
\newblock


\bibitem[Thoma et~al\mbox{.}(2019)]%
        {thoma2019coper}
\bibfield{author}{\bibinfo{person}{Louise~M Thoma}, \bibinfo{person}{Hege Grindem}, \bibinfo{person}{David Logerstedt}, \bibinfo{person}{Michael Axe}, \bibinfo{person}{Lars Engebretsen}, \bibinfo{person}{May~Arna Risberg}, {and} \bibinfo{person}{Lynn Snyder-Mackler}.} \bibinfo{year}{2019}\natexlab{}.
\newblock \showarticletitle{Coper classification early after anterior cruciate ligament rupture changes with progressive neuromuscular and strength training and is associated with 2-year success: the Delaware-Oslo ACL cohort study}.
\newblock \bibinfo{journal}{\emph{The American Journal of Sports Medicine}} \bibinfo{volume}{47}, \bibinfo{number}{4} (\bibinfo{year}{2019}), \bibinfo{pages}{807--814}.
\newblock
\href{https://doi.org/10.1177/0363546519825500}{doi:\nolinkurl{10.1177/0363546519825500}}


\bibitem[Van Der~Ploeg et~al\mbox{.}(2014)]%
        {van2014modern}
\bibfield{author}{\bibinfo{person}{Tjeerd Van Der~Ploeg}, \bibinfo{person}{Peter~C. Austin}, {and} \bibinfo{person}{Ewout~W. Steyerberg}.} \bibinfo{year}{2014}\natexlab{}.
\newblock \showarticletitle{Modern modelling techniques are data hungry: a simulation study for predicting dichotomous endpoints}.
\newblock \bibinfo{journal}{\emph{BMC medical research methodology}}  \bibinfo{volume}{14} (\bibinfo{year}{2014}), \bibinfo{pages}{1--13}.
\newblock
\href{https://doi.org/10.1186/1471-2288-14-137}{doi:\nolinkurl{10.1186/1471-2288-14-137}}


\bibitem[Wallsberger et~al\mbox{.}(2024)]%
        {wallsberger2024explainable}
\bibfield{author}{\bibinfo{person}{Raphael Wallsberger}, \bibinfo{person}{Ricardo Knauer}, {and} \bibinfo{person}{Stephan Matzka}.} \bibinfo{year}{2024}\natexlab{}.
\newblock \showarticletitle{Explainable Artificial Intelligence Beyond Feature Attributions: The Validity and Reliability of Feature Selection Explanations}. In \bibinfo{booktitle}{\emph{Proceedings of the 2nd World Conference on eXplainable Artificial Intelligence (xAI)}}. \bibinfo{publisher}{CEUR-WS}, \bibinfo{address}{Valletta, Malta}, \bibinfo{pages}{1--8}.
\newblock


\bibitem[Wang et~al\mbox{.}(2025)]%
        {wang2025largelanguagemodels}
\bibfield{author}{\bibinfo{person}{Ruochen Wang}, \bibinfo{person}{Si Si}, \bibinfo{person}{Felix Yu}, \bibinfo{person}{Dorothea~Wiesmann Rothuizen}, \bibinfo{person}{Cho-Jui Hsieh}, {and} \bibinfo{person}{Inderjit~S. Dhillon}.} \bibinfo{year}{2025}\natexlab{}.
\newblock \showarticletitle{Large Language Models are Interpretable Learners}. In \bibinfo{booktitle}{\emph{Proceedings of the 13th International Conference on Learning Representations (ICLR)}}. \bibinfo{publisher}{International Conference on Learning Representations (ICLR)}, \bibinfo{address}{Singapore}.
\newblock


\bibitem[Wei et~al\mbox{.}(2022)]%
        {wei2022chain}
\bibfield{author}{\bibinfo{person}{Jason Wei}, \bibinfo{person}{Xuezhi Wang}, \bibinfo{person}{Dale Schuurmans}, \bibinfo{person}{Maarten Bosma}, \bibinfo{person}{Fei Xia}, \bibinfo{person}{Ed Chi}, \bibinfo{person}{Quoc~V Le}, \bibinfo{person}{Denny Zhou}, {et~al\mbox{.}}} \bibinfo{year}{2022}\natexlab{}.
\newblock \showarticletitle{Chain-of-thought prompting elicits reasoning in large language models}.
\newblock \bibinfo{journal}{\emph{Advances in neural information processing systems}}  \bibinfo{volume}{35} (\bibinfo{year}{2022}), \bibinfo{pages}{24824--24837}.
\newblock
\href{https://doi.org/10.48550/arXiv.2201.11903}{doi:\nolinkurl{10.48550/arXiv.2201.11903}}


\bibitem[Winter(2009)]%
        {winter2009biomechanics}
\bibfield{author}{\bibinfo{person}{David~A. Winter}.} \bibinfo{year}{2009}\natexlab{}.
\newblock \bibinfo{booktitle}{\emph{Biomechanics and motor control of human movement}}.
\newblock \bibinfo{publisher}{John wiley \& sons}.
\newblock


\bibitem[Wirth and Eckstein(2008)]%
        {wirth2008technique}
\bibfield{author}{\bibinfo{person}{Wolfgang Wirth} {and} \bibinfo{person}{Felix Eckstein}.} \bibinfo{year}{2008}\natexlab{}.
\newblock \showarticletitle{A technique for regional analysis of femorotibial cartilage thickness based on quantitative magnetic resonance imaging}.
\newblock \bibinfo{journal}{\emph{IEEE transactions on medical imaging}} \bibinfo{volume}{27}, \bibinfo{number}{6} (\bibinfo{year}{2008}), \bibinfo{pages}{737--744}.
\newblock
\href{https://doi.org/10.1109/TMI.2007.907323}{doi:\nolinkurl{10.1109/TMI.2007.907323}}


\bibitem[Wirth et~al\mbox{.}(2017)]%
        {wirth2017sex}
\bibfield{author}{\bibinfo{person}{Wolfgang Wirth}, \bibinfo{person}{Susanne Maschek}, {and} \bibinfo{person}{Felix Eckstein}.} \bibinfo{year}{2017}\natexlab{}.
\newblock \showarticletitle{Sex-and age-dependence of region-and layer-specific knee cartilage composition (spin--spin--relaxation time) in healthy reference subjects}.
\newblock \bibinfo{journal}{\emph{Annals of Anatomy-Anatomischer Anzeiger}}  \bibinfo{volume}{210} (\bibinfo{year}{2017}), \bibinfo{pages}{1--8}.
\newblock
\href{https://doi.org/10.1016/j.aanat.2016.10.010}{doi:\nolinkurl{10.1016/j.aanat.2016.10.010}}


\bibitem[Wu and Cavanagh(1995)]%
        {wu1995isb}
\bibfield{author}{\bibinfo{person}{Ge Wu} {and} \bibinfo{person}{Peter~R. Cavanagh}.} \bibinfo{year}{1995}\natexlab{}.
\newblock \showarticletitle{ISB recommendations for standardization in the reporting of kinematic data}.
\newblock \bibinfo{journal}{\emph{Journal of biomechanics}} \bibinfo{volume}{28}, \bibinfo{number}{10} (\bibinfo{year}{1995}), \bibinfo{pages}{1257--1262}.
\newblock
\href{https://doi.org/10.1016/0021-9290(95)00017-C}{doi:\nolinkurl{10.1016/0021-9290(95)00017-C}}


\bibitem[Xie et~al\mbox{.}(2022)]%
        {xie2022explanation}
\bibfield{author}{\bibinfo{person}{Sang~Michael Xie}, \bibinfo{person}{Aditi Raghunathan}, \bibinfo{person}{Percy Liang}, {and} \bibinfo{person}{Tengyu Ma}.} \bibinfo{year}{2022}\natexlab{}.
\newblock \showarticletitle{An Explanation of In-context Learning as Implicit Bayesian Inference}. In \bibinfo{booktitle}{\emph{Proceedings of the 10th International Conference on Learning Representations (ICLR)}}. \bibinfo{publisher}{International Conference on Learning Representations (ICLR)}, \bibinfo{address}{Virtual}.
\newblock


\bibitem[Yang et~al\mbox{.}(2024)]%
        {yang2024advancing}
\bibfield{author}{\bibinfo{person}{Lin Yang}, \bibinfo{person}{Shawn Xu}, \bibinfo{person}{Andrew Sellergren}, \bibinfo{person}{Timo Kohlberger}, \bibinfo{person}{Yuchen Zhou}, \bibinfo{person}{Ira Ktena}, \bibinfo{person}{Atilla Kiraly}, \bibinfo{person}{Faruk Ahmed}, \bibinfo{person}{Farhad Hormozdiari}, \bibinfo{person}{Tiam Jaroensri}, \bibinfo{person}{Eric Wang}, \bibinfo{person}{Ellery Wulczyn}, \bibinfo{person}{Fayaz Jamil}, \bibinfo{person}{Theo Guidroz}, \bibinfo{person}{Chuck Lau}, \bibinfo{person}{Siyuan Qiao}, \bibinfo{person}{Yun Liu}, \bibinfo{person}{Akshay Goel}, \bibinfo{person}{Kendall Park}, \bibinfo{person}{Arnav Agharwal}, \bibinfo{person}{Nick George}, \bibinfo{person}{Yang Wang}, \bibinfo{person}{Ryutaro Tanno}, \bibinfo{person}{David G.~T. Barrett}, \bibinfo{person}{Wei-Hung Weng}, \bibinfo{person}{S.~Sara Mahdavi}, \bibinfo{person}{Khaled Saab}, \bibinfo{person}{Tao Tu}, \bibinfo{person}{Sreenivasa~Raju Kalidindi}, \bibinfo{person}{Mozziyar Etemadi}, \bibinfo{person}{Jorge Cuadros},
  \bibinfo{person}{Gregory Sorensen}, \bibinfo{person}{Yossi Matias}, \bibinfo{person}{Katherine Chou}, \bibinfo{person}{Greg Corrado}, \bibinfo{person}{Joelle Barral}, \bibinfo{person}{Shravya Shetty}, \bibinfo{person}{David Fleet}, \bibinfo{person}{S.~M.~Ali Eslami}, \bibinfo{person}{Daniel Tse}, \bibinfo{person}{Shruthi Prabhakara}, \bibinfo{person}{Cory McLean}, \bibinfo{person}{Dave Steiner}, \bibinfo{person}{Rory Pilgrim}, \bibinfo{person}{Christopher Kelly}, \bibinfo{person}{Shekoofeh Azizi}, {and} \bibinfo{person}{Daniel Golden}.} \bibinfo{year}{2024}\natexlab{}.
\newblock \bibinfo{title}{Advancing Multimodal Medical Capabilities of Gemini}.
\newblock
\showeprint[arxiv]{2405.03162}~[cs.CV]
\urldef\tempurl%
\url{https://arxiv.org/abs/2405.03162}
\showURL{%
\tempurl}


\bibitem[Ye et~al\mbox{.}(2024)]%
        {ye2024closerlook}
\bibfield{author}{\bibinfo{person}{Han-Jia Ye}, \bibinfo{person}{Si-Yang Liu}, \bibinfo{person}{Hao-Run Cai}, \bibinfo{person}{Qi-Le Zhou}, {and} \bibinfo{person}{De-Chuan Zhan}.} \bibinfo{year}{2024}\natexlab{}.
\newblock \bibinfo{title}{A Closer Look at Deep Learning on Tabular Data}.
\newblock
\showeprint[arxiv]{2407.00956}~[cs.LG]
\urldef\tempurl%
\url{https://arxiv.org/abs/2407.00956}
\showURL{%
\tempurl}


\bibitem[Yuksekgonul et~al\mbox{.}(2025)]%
        {yuksekgonul2025optimizing}
\bibfield{author}{\bibinfo{person}{Mert Yuksekgonul}, \bibinfo{person}{Federico Bianchi}, \bibinfo{person}{Joseph Boen}, \bibinfo{person}{Sheng Liu}, \bibinfo{person}{Pan Lu}, \bibinfo{person}{Zhi Huang}, \bibinfo{person}{Carlos Guestrin}, {and} \bibinfo{person}{James Zou}.} \bibinfo{year}{2025}\natexlab{}.
\newblock \showarticletitle{Optimizing generative AI by backpropagating language model feedback}.
\newblock \bibinfo{journal}{\emph{Nature}} \bibinfo{volume}{639}, \bibinfo{number}{8055} (\bibinfo{year}{2025}), \bibinfo{pages}{609--616}.
\newblock


\end{thebibliography}

\clearpage
\appendix

\section{Additional Dataset Details} \label{sec:appendix_dataset}

\subsection{Public Datasets}

For our 13 public datasets, we used tabmemcheck 0.1.5 to evaluate whether state-of-the-art LLMs know or have memorized them, \textit{i.e.}, we assessed the LLMs' knowledge of the feature names with the feature names test, the memorization of the table header with the header test, the memorization of 25 random rows with the row completion test, the memorization of 25 unique feature values with the feature completion test, and the memorization of the first tokens from 25 random rows with the first token test \citep{bordt2024elephants1,bordt2024elephants2}. The test results are shown in Table~\ref{table:datasets}, together with the sample sizes, feature set sizes, and number of classes. We found evidence of knowledge (54\%) or partial memorization (62\%) of the public datasets by at least 1 state-of-the-art LLM. Furthermore, negative test results do not imply the LLMs' have not seen or memorized the datasets, it might just not be possible to extract the relevant information via prompting. Our LLM-based decision tree induction and embedding approaches may therefore show overoptistmic performance results on the public datasets.

\subsection{Private Datasets}

To obtain unbiased performance estimates, we additionally included 2 private datasets that are not publicly available and hence cannot have been seen during the LLMs' pretraining \citep{bordt2024elephants1,bordt2024elephants2}.

\subsubsection{Post-Trauma Pain}

The first private dataset is from a recent prospective cohort study in 124 post-trauma patients \citep{evans2022estimating}. The prediction target $p$ was to classify ``whether the pain and disability outcome at 6 months following musculoskeletal trauma is good or bad''. At 6 months, there were 82 follow-up responders. The baseline set of 40 predictors included surrogates for pain mechanisms, qualitative sensory testing, and psychosocial factors, among others.\footnote{Please see \texttt{https://github.com/ml-lab-htw/llm-trees/\allowbreak blob/main/data\_sets/posttrauma/feature\_description.txt} for the full list of feature names $[f_1, ..., f_k]$.} We regarded the gender at birth, ethnicity, education age, work status, and penetrating injury as nominal, and the smoke status and injury severity score category as ordinal. Ordinal predictors were treated as numerical in our experiments \citep{steyerberg2019clinical}. Missing values ranged from 0\% to 30\% per feature, with a mean of 5\% and a median of 1\%. For more details on the dataset, please refer to the original publication.

\subsubsection{Anterior Cruciate Ligament Injury}

The second private dataset is from a to-be-published prospective cohort study in 34 patients with an ACL rupture \citep{brisson2020cartilage,brisson2024cartilage}. The prediction target $p$ was to classify ``whether the cartilage thickness change in the central medial femur over 12 months following anterior cruciate ligament injury is normal or abnormal''. Patients were assessed at a baseline of 5 months post-injury (± unbiased standard deviation estimate (SD) 3 months) and at a follow-up of 12 months (±SD 1 month). At baseline, we collected 3-dimensional gait data using a 10-camera motion capture system (Vicon) and 2 force plates (AMTI) during barefoot, overground walking at a natural, self-selected speed. Using the gait data and inverse dynamics \citep{winter2009biomechanics}, we computed external peak knee moments in a 3-dimensional floating axis coordinate system \citep{wu1995isb}. At both baseline and 12-months follow-up, we performed magnetic resonance imaging (MRI) with a 1.5 Tesla scanner (Avanto, Siemens; sagittal multi-echo spin-echo (MESE); 3.50mm slice spacing; 0.31mm in-plane resolution) to compute tibiofemoral, subregional mean cartilage thickness \citep{wirth2008technique} and measures of cartilage composition, \textit{i.e.}, mean superficial and deep cartilage transverse relaxation times (T2), 50\% each \citep{wirth2017sex}. We were primarily interested in excessive cartilage thinning or thickening in the central medial femur over 12 months \citep{biswal2002risk,frobell2009acutely} as a potential precursor to osteoarthritis \citep{iriondo2021towards}. We considered a cartilage thickness change abnormal in our experiments if it exceeded 1 SD of 13 healthy controls (0.03mm ±SD 0.07mm). The set of 20 baseline predictors included external peak knee moments, baseline mean cartilage thickness in the central medial femur, measures of cartilage composition in the central medial femur, patient-reported outcome measures, and a group assignment into copers (treated conservatively and clinically deemed to have mechanically stable knees), noncopers (treated conservatively and clinically deemed to have persistent mechanical knee instability), and surgical reconstruction (ipsilateral four-strand semitendinosus reconstruction, 4 months post-injury ±SD 3 months) \citep{kaplan2011identifying,thoma2019coper}, among others.\footnote{Please see \texttt{https://github.com/ml-lab-htw/llm-trees/\allowbreak blob/main/data\_sets/acl/feature\_description.txt} for the full list of feature names $[f_1, ..., f_k]$.} We regarded the group assignment, sex, and dominant leg as nominal, and the Tegner score as ordinal. The Tegner score was treated as numerical in our experiments \citep{steyerberg2019clinical}. There were no missing values in this dataset.

\section{Additional Results} \label{sec:appendix_results}

\begin{figure*}[t]
\centering
\begin{subfigure}{\textwidth}
\includegraphics[width=0.45\textwidth]{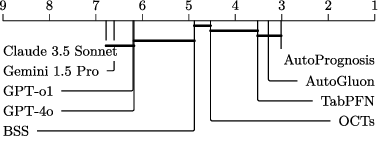}
\hspace{10mm}
\includegraphics[width=0.45\textwidth]{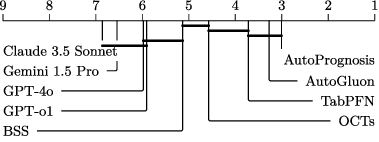}
\caption{Test F1-score ranks (left) and test balanced accuracy ranks (right) for our zero-shot decision tree induction experiments.}
\label{fig:results_induction_cdd}
\vspace{3mm}
\end{subfigure}
\begin{subfigure}{\textwidth}
\includegraphics[width=0.45\textwidth]{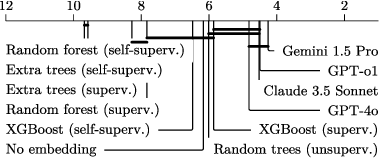}
\hspace{10mm}
\includegraphics[width=0.45\textwidth]{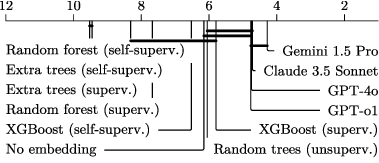}
\caption{Test F1-score ranks (left) and test balanced accuracy ranks (right) for our zero-shot decision tree embedding experiments.}
\label{fig:results_embedding_cdd}
\end{subfigure}
\caption{Critical difference diagrams to detect pairwise test score differences between our methods at 67\%/33\% train/test splits for the public and private datasets, based on the Holm-adjusted Wilcoxon signed-rank test \citep{benavoli2016should,demvsar2006statistical}. Approaches that are not statistically different at the 0.05 significance level are connected by a bold horizontal bar. Missing performance scores were imputed with 0 before running the tests.}
\Description{Critical difference diagrams for the test F1-scores and balanced accuracies for our induction and embedding experiments.}
\label{fig:results_cdd}
\end{figure*}

\begin{figure*}[t]
\centering
\begin{subfigure}{\textwidth}
    \centering
    \includegraphics[width=0.7\textwidth]{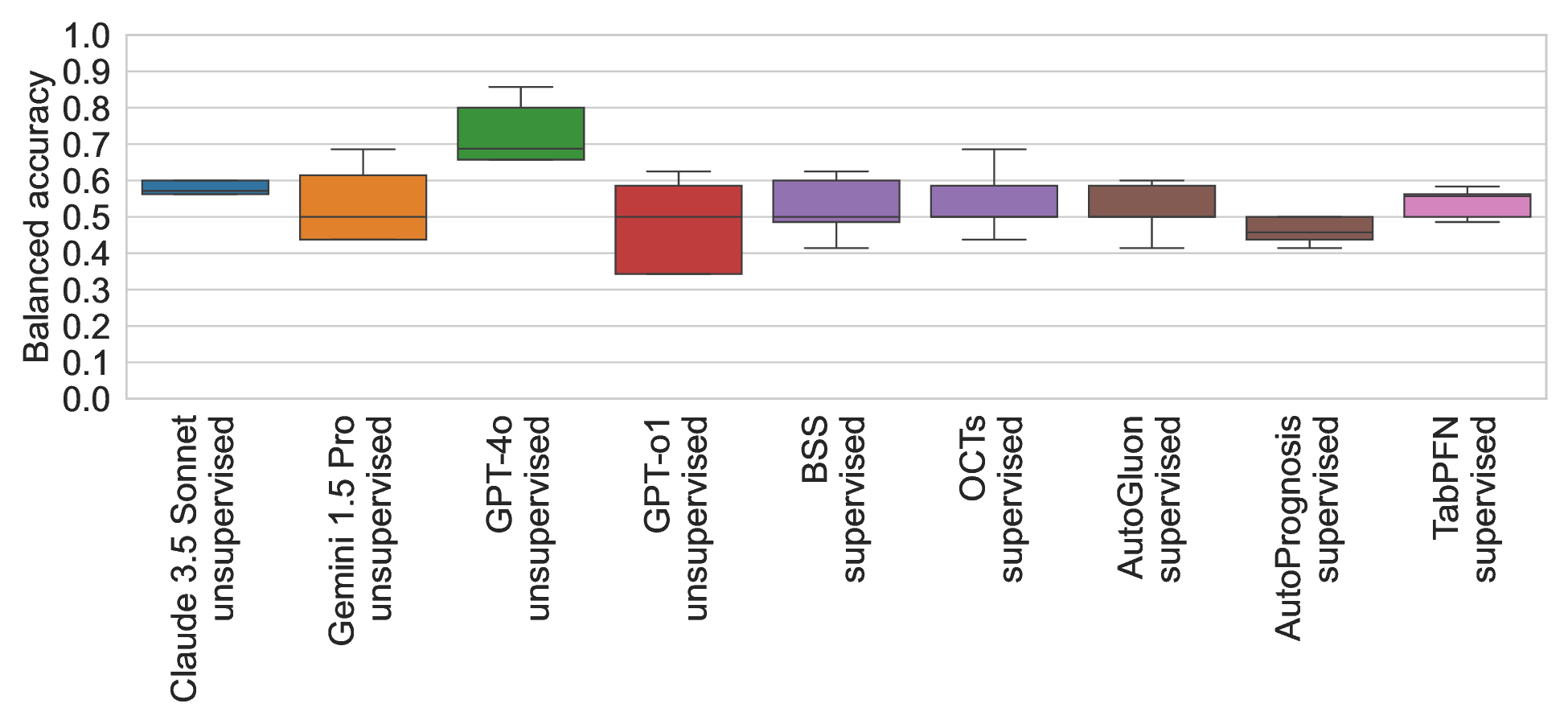}
    \caption{Test balanced accuracy on ACL injury data.}
    \label{fig:boxplot_induction_acl_acc}
\end{subfigure}
\vfill
\begin{subfigure}{\textwidth}
    \centering
    \includegraphics[width=0.7\textwidth]{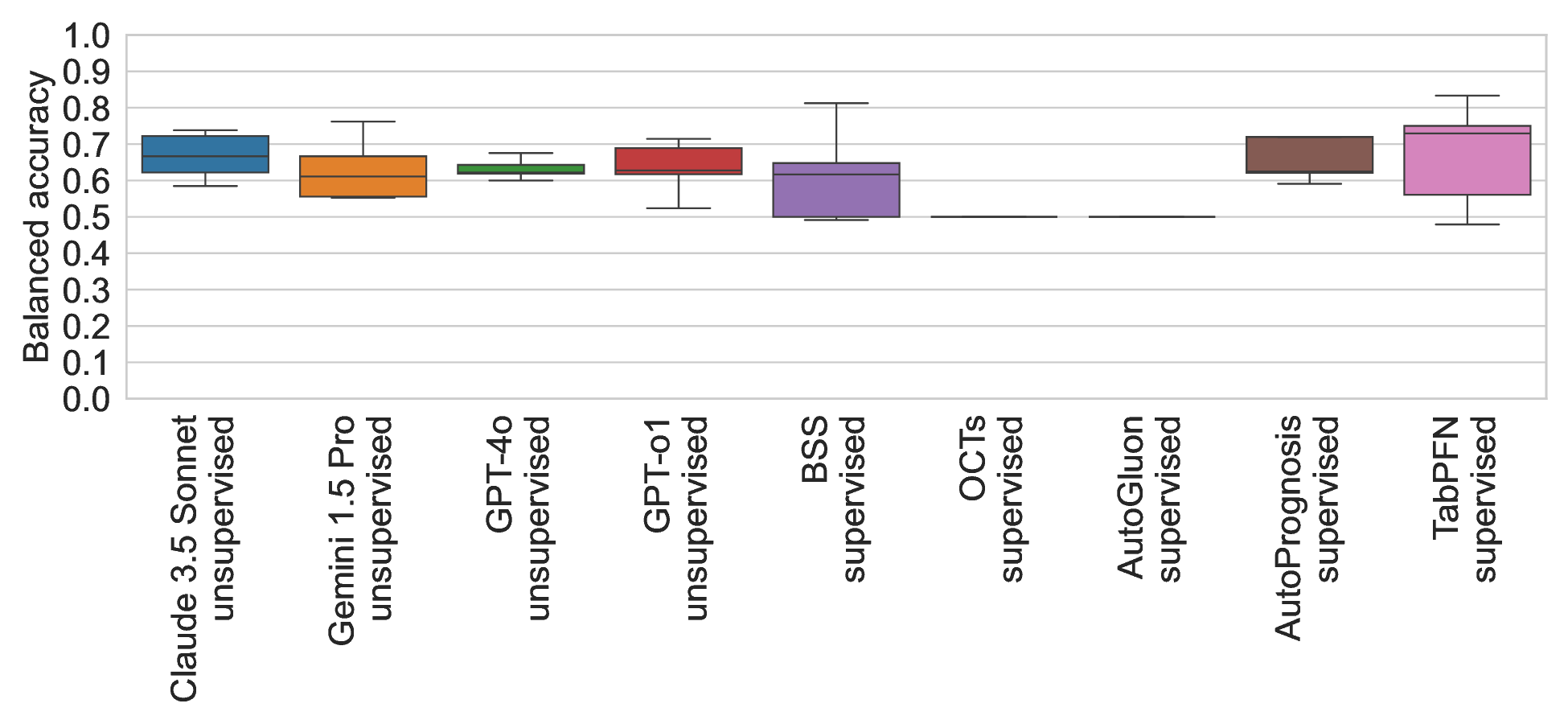}
    \caption{Test balanced accuracy on post-trauma pain data.}
    \label{fig:boxplot_induction_posttrauma_acc}
\end{subfigure}
\caption{Test balanced accuracy at 67\%/33\% train/test splits for our LLM-based zero-shot decision tree induction approach compared to the machine learning baselines on our private (a) ACL injury data and (b) post-trauma pain data.}
\Description{Bar charts showing test balanced accuracies for our induction experiments for the private data.}
\label{fig:boxplot_induction_acl_posttrauma_acc}
\end{figure*}

\begin{figure*}[t]
\centering
\begin{subfigure}{\textwidth}
    \centering
    \includegraphics[width=0.7\textwidth]{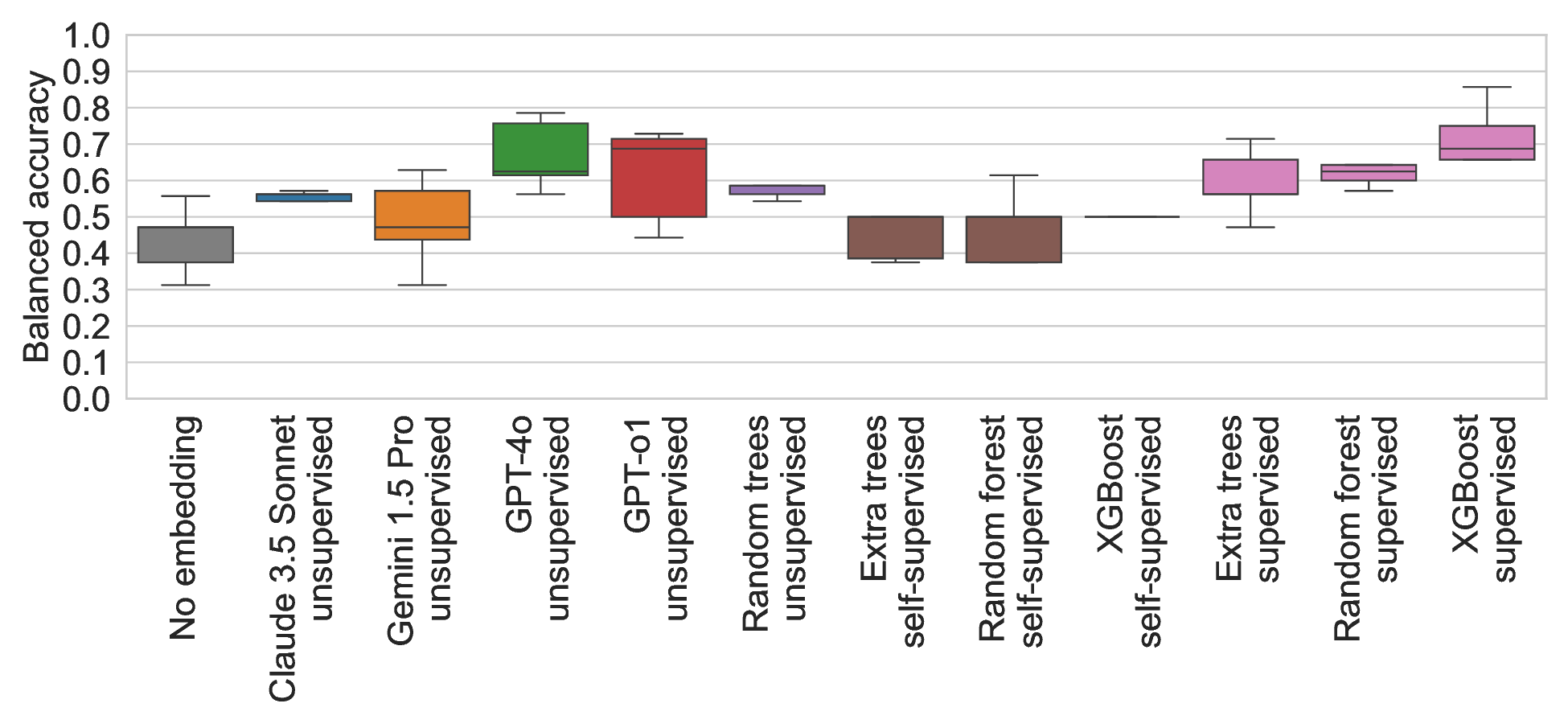}
    \caption{Test balanced accuracy on ACL injury data.}
    \label{fig:boxplot_embedding_acl_acc}
\end{subfigure}
\vfill
\begin{subfigure}{\textwidth}
    \centering
    \includegraphics[width=0.7\textwidth]{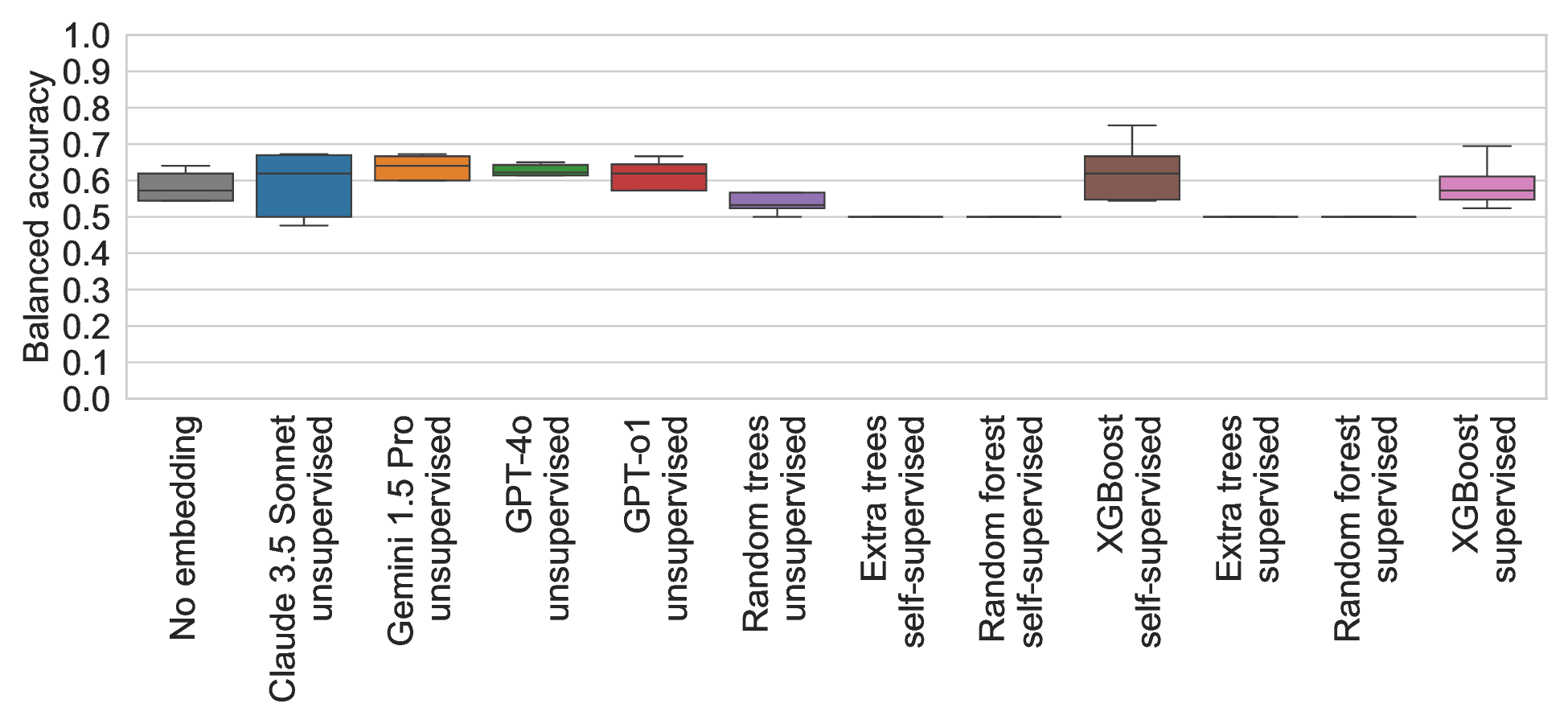}
    \caption{Test balanced accuracy on post-trauma pain data.}
    \label{fig:boxplot_embedding_posttrauma_acc}
\end{subfigure}
\caption{Test balanced accuracy at 67\%/33\% train/test splits of a multi-layer perceptron without embeddings, with our LLM-based zero-shot decision tree embeddings, as well as with unsupervised, self-supervised, and supervised embedding baselines on our private (a) ACL injury data and (b) post-trauma pain data.}
\Description{Bar charts showing test balanced accuracies for our embedding experiments for the private data.}
\label{fig:boxplot_embedding_acl_posttrauma_acc}
\end{figure*}

In this section, we present additional experimental results. Figure~\ref{fig:results_cdd} shows the statistical significance of our findings via critical difference diagrams \citep{benavoli2016should,demvsar2006statistical}. Table~\ref{table:results_induction_acc}, Table~\ref{tab:ablations_induction}, Figure~\ref{fig:boxplot_induction_acl_posttrauma_acc}, Figure~\ref{fig:grouped_boxplot_induction_acl_posttrauma_public_f1}, and Figure~\ref{fig:grouped_boxplot_induction_acl_posttrauma_public_acc} show further results for our zero-shot decision tree induction, Table~\ref{tab:results_embeddings_acc}, Table~\ref{tab:ablations_embedding}, and Figure~\ref{fig:boxplot_embedding_acl_posttrauma_acc} for our zero-shot decision tree embedding. Furthermore, we provide additional details on the best trees for the private datasets and our zero-shot embeddings below.

\subsection{Best Induced Trees on Private Datasets}

\begin{figure*}[t]
\centering

\begin{subfigure}{\textwidth}
\centering
\includegraphics[width=0.7\textwidth]{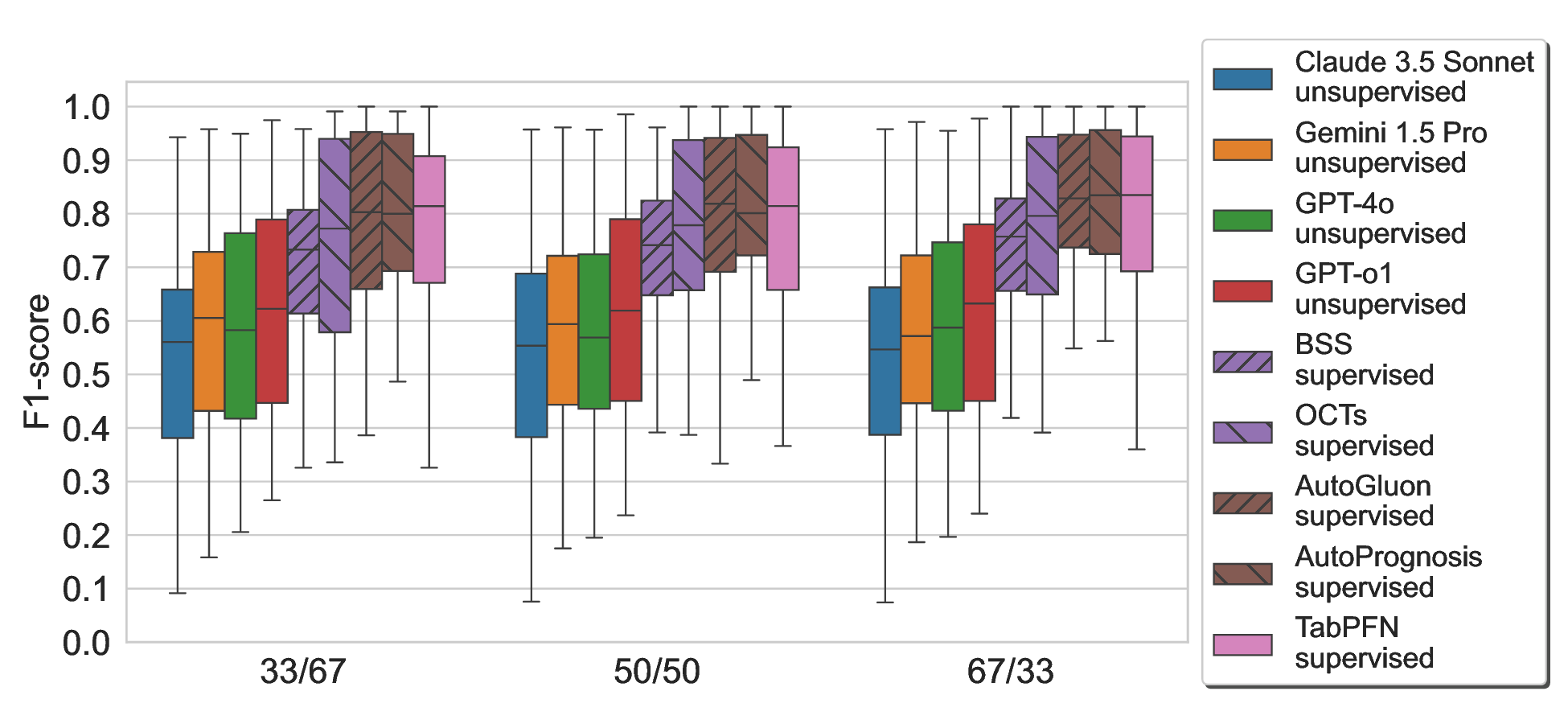}
\caption{Test F1-score on public data.}
\label{fig:grouped_boxplot_induction_public_f1}
\end{subfigure}
\vfill
\begin{subfigure}{\textwidth}
\centering
\includegraphics[width=0.7\textwidth]{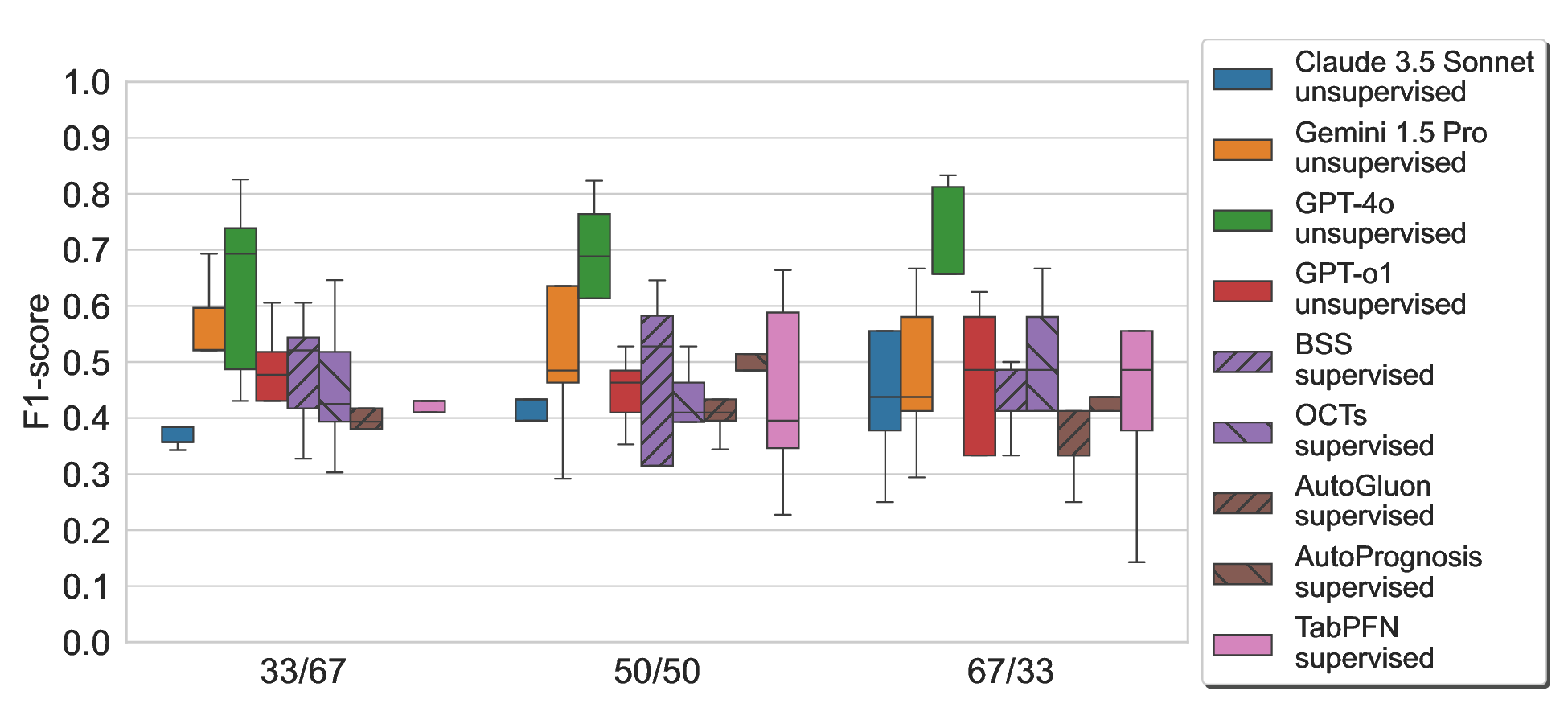}
\caption{Test F1-score on ACL injury data.}
\label{fig:grouped_boxplot_induction_acl_f1}
\end{subfigure}
\vfill
\begin{subfigure}{\textwidth}
\centering
\includegraphics[width=0.7\textwidth]{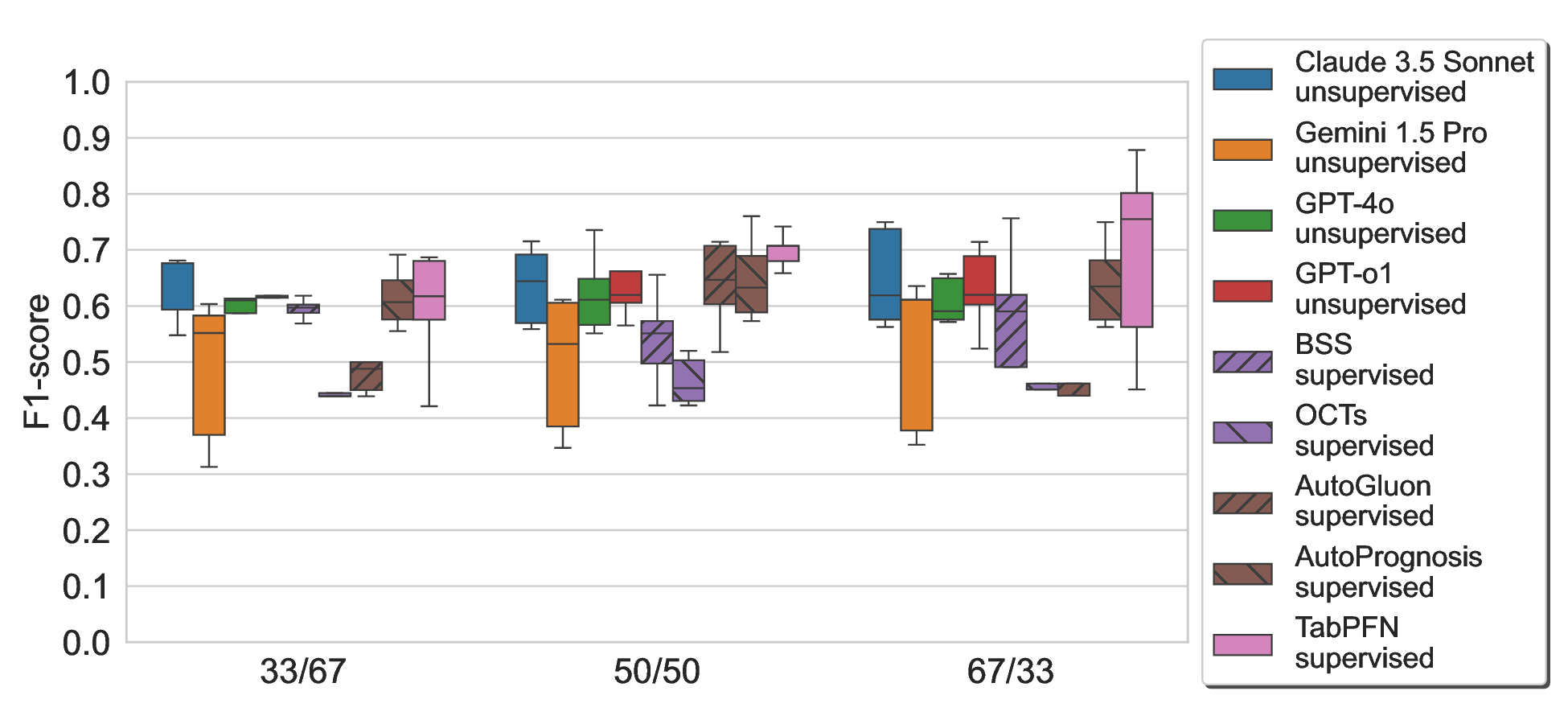}
\caption{Test F1-score on post-trauma pain data.}
\label{fig:grouped_boxplot_induction_posttrauma_f1}
\end{subfigure}

\caption{Test F1-scores at different train/test splits for our LLM-based zero-shot decision tree induction approach compared to the machine learning baselines on our (a) public data, (b) private ACL injury data, and (c) private post-trauma pain data.}
\Description{Bar charts showing test F1-scores across different train/test splits for our induction experiments.}
\label{fig:grouped_boxplot_induction_acl_posttrauma_public_f1}
\end{figure*}

\begin{figure*}[t]
\centering

\begin{subfigure}{\textwidth}
\centering
\includegraphics[width=0.7\textwidth]{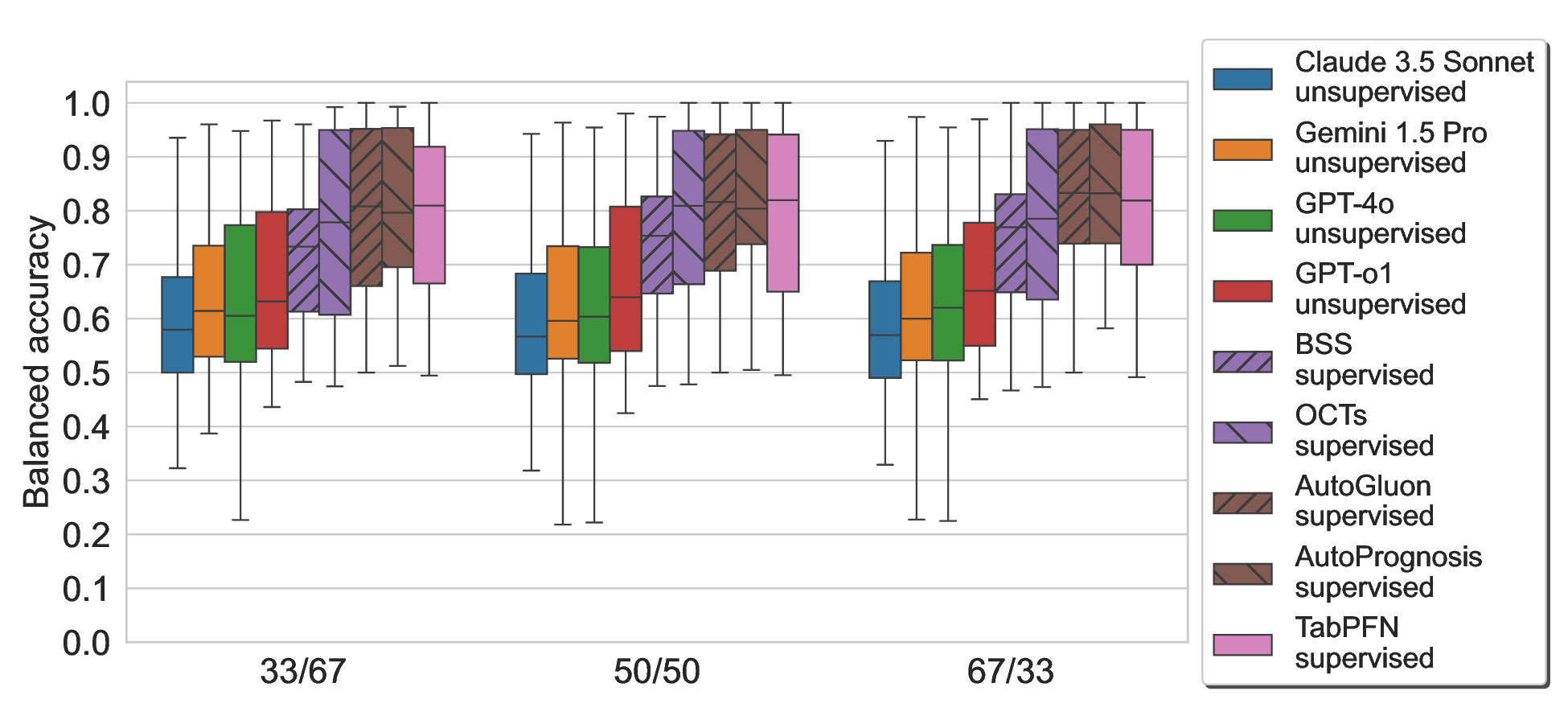}
\caption{Test balanced accuracy on public data.}
\label{fig:grouped_boxplot_induction_public_acc}
\end{subfigure}
\vfill
\begin{subfigure}{\textwidth}
\centering
\includegraphics[width=0.7\textwidth]{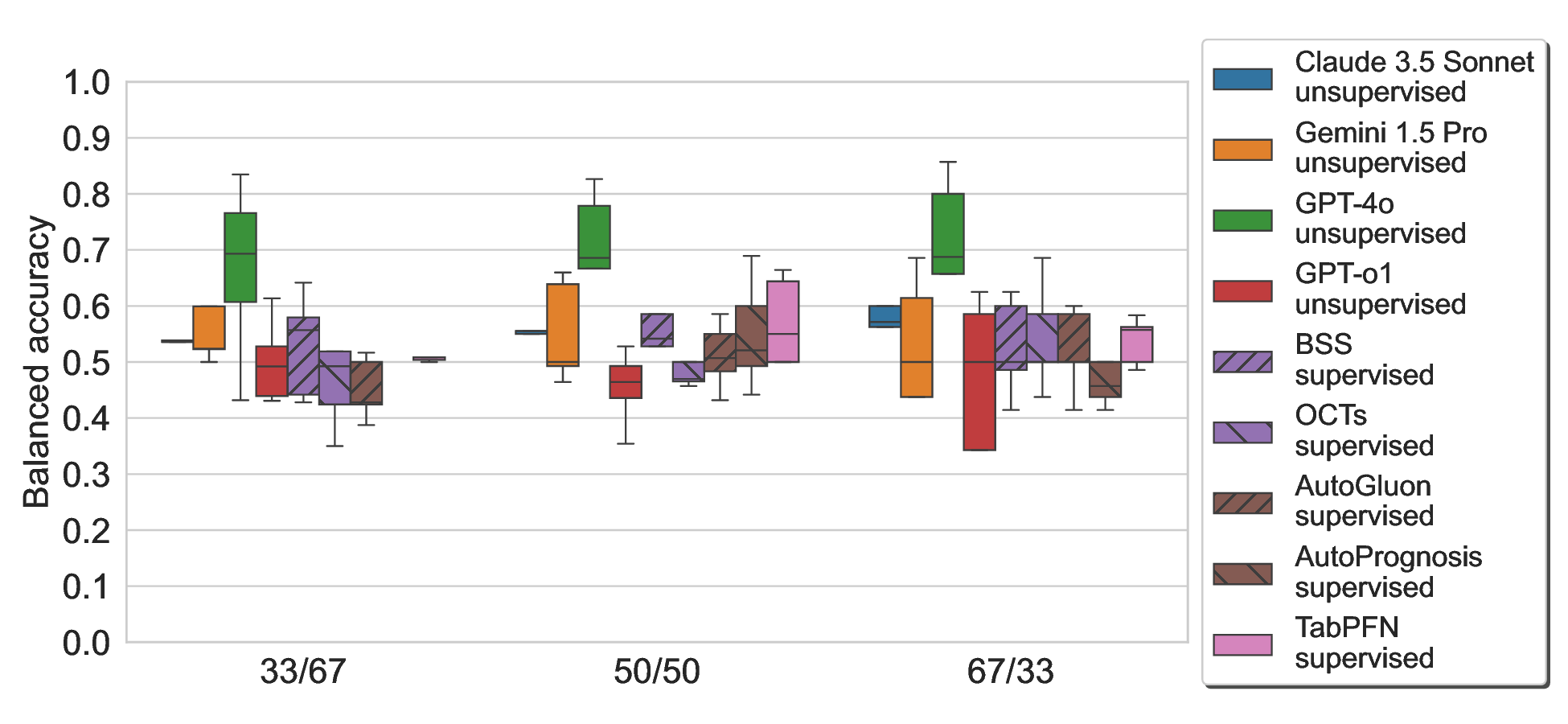}
\caption{Test balanced accuracy on ACL injury data.}
\label{fig:grouped_boxplot_induction_acl_acc}
\end{subfigure}
\vfill
\begin{subfigure}{\textwidth}
\centering
\includegraphics[width=0.7\textwidth]{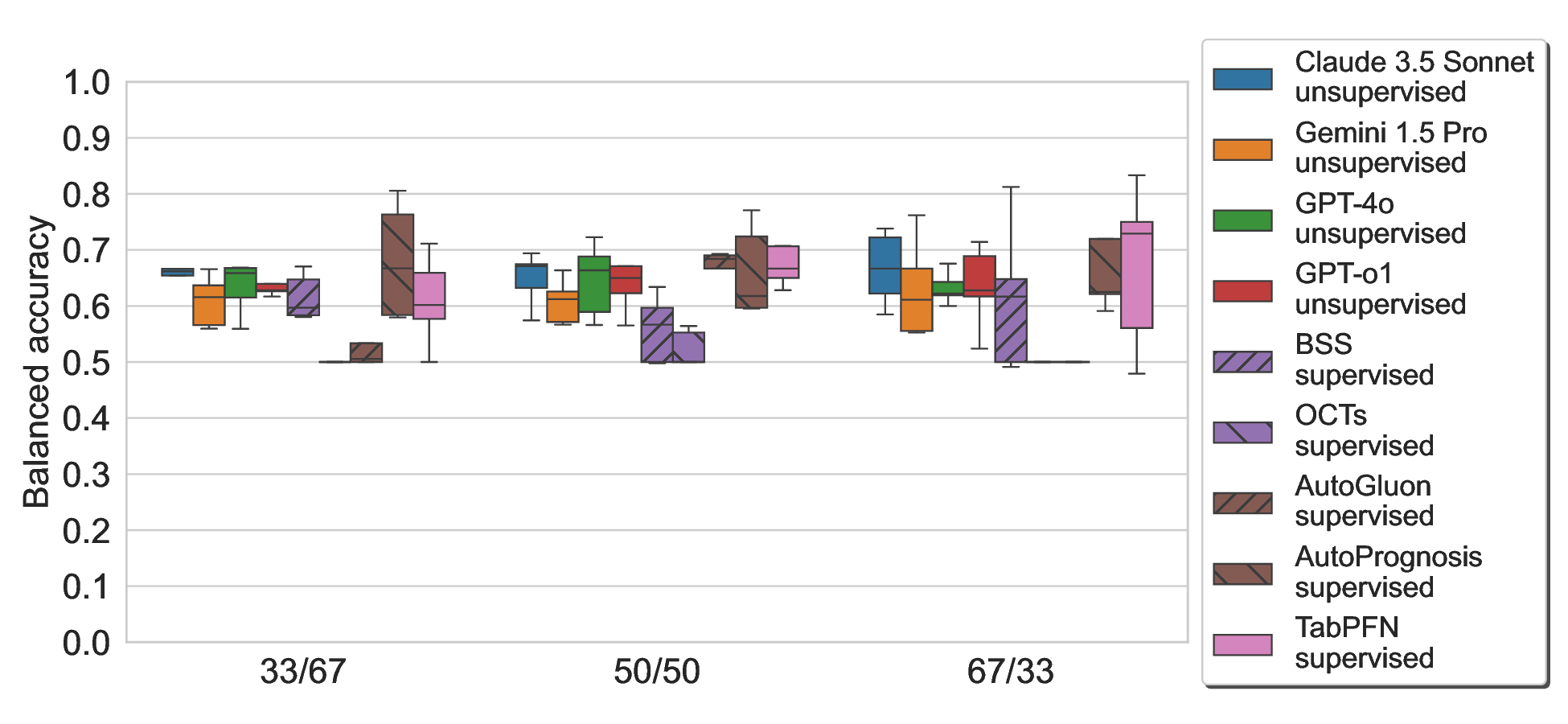}
\caption{Test balanced accuracy on post-trauma pain data.}
\label{fig:grouped_boxplot_induction_posttrauma_acc}
\end{subfigure}

\caption{Test balanced accuracies at different train/test splits for our LLM-based zero-shot decision tree induction approach compared to the machine learning baselines on our (a) public data, (b) private ACL injury data, and (c) private post-trauma pain data.}
\Description{Bar charts showing test balanced accuracies across different train/test splits for our induction experiments.}
\label{fig:grouped_boxplot_induction_acl_posttrauma_public_acc}
\end{figure*}

On our ACL injury data, the best tree was built by GPT-4o, with a test F1-score of 0.83 and a test balanced accuracy of 0.86. It included the mean cartilage thickness in the central medial femur, the mean T2 values for the deep cartilage layers in the central medial femur, and the group assignment into the coper group as features (Figure~\ref{fig:best_tree_knowledge_acl}). The best data-driven tree only included the height as a feature (Figure~\ref{fig:best_tree_data_acl}) and was therefore too simplistic to achieve good test F1-scores / balanced accuracies (0.67 / 0.69). The GPT-4o tree appreciably outperformed not only the best data-driven tree, but all other baselines across all train/test splits (Figure~\ref{fig:grouped_boxplot_induction_acl_f1} and Figure~\ref{fig:grouped_boxplot_induction_acl_acc}).

On our post-trauma pain data, the best tree was built by Claude 3.5 Sonnet, with a test F1-score of 0.75 and a test balanced accuracy of 0.74. It included the worst pain intensity and the pain self-efficacy questionnaire (PSEQ) score as features (Figure~\ref{fig:best_tree_knowledge_posttrauma}). The best data-driven tree included the number of fractures and pain extent as features (Figure~\ref{fig:best_tree_data_posttrauma}), but only achieved a test F1-score of 0.59 and a test balanced accuracy of 0.58. However, the best tree could vary depending on the split (Figure~\ref{fig:grouped_boxplot_induction_posttrauma_f1} and Figure~\ref{fig:grouped_boxplot_induction_posttrauma_acc}).

\subsection{Additional Embedding Insights}

Table~\ref{tab:results_embeddings_dimension} shows the dimensions of the zero-shot embeddings generated by the 4 LLMs. In general, the LLM-based embeddings increased or decreased the feature dimensionality, depending on the dataset and LLM. The embedding size varied significantly across datasets when no constraints were applied, suggesting that different maximum tree depths were chosen by the LLMs. When considering the results in Table~\ref{tab:results_embeddings_f1} and Table~\ref{tab:results_embeddings_acc}, though, no clear correlation between the embedding size and performance could be established. For instance, on the \textit{labor} dataset, Claude 3.5 Sonnet increased the feature dimensionality by +9, GPT-o1 did not change the dimensionality, and Gemini 1.5 Pro and GPT-4o decreased the dimensionality by -6 and -6. Nevertheless, Claude 3.5 Sonnet and GPT-o1 only showed modest test F1-score / balanced accuracy improvements of +0.05 / +0.07 and +0.04 / +0.08 compared to no embedding. On the other hand, Gemini 1.5 Pro exhibited larger improvements of +0.10 / +0.11, while GPT-4o showed a similar performance compared to no embedding (+0.01 / +0.03).

In terms of the number of selected features, we observed that the LLM-based embeddings leveraged a diverse feature set, providing a rich representation of the data. The embeddings resulted in trees that often included a comparable number of features, although GPT-4o and GPT-o1 sometimes selected a much higher number of features (Table~\ref{tab:results_embeddings_num_features}). This general consistency in feature usage across different models suggests that the observed performance differences were likely not only due to the selection of features, but also their combination and the thresholds applied.

\begin{table*}[t]
  \caption{Additional dataset details for our 13 public datasets, including the sample sizes, the feature set sizes, the number of classes, and tests evaluating whether state-of-the-art LLMs know or have memorized our public datasets \citep{bordt2024elephants1,bordt2024elephants2}. The knowledge and memorization test results for Anthropic Claude 3.5 Sonnet, Google Gemini 1.5 Pro, OpenAI GPT-4o, and OpenAI GPT-o1 are depicted in the table as * / * / * / *, with ``\checkmark'', ``$\times$'', ``?'', and ``-'' denoting ``evidence of knowledge or memorization'', ``no evidence of knowledge or memorization'', ``ambiguous result'', and ``test could be conducted'', respectively. Note that negative test results do not guarantee that an LLM has not seen or memorized a dataset.
  }
  \fontsize{8}{8}\selectfont
  \centering
  \begin{tabular}{p{1.7cm} p{0.4cm} p{0.4cm} p{0.4cm} p{1.25cm} p{1.25cm} p{1.25cm} p{1.25cm} p{1.25cm}}
  \toprule \\
    \textbf{Dataset} & \textbf{\# Samples} & \textbf{\# Features} & \textbf{\# Clas-ses} & \textbf{Feature names test} & \textbf{Header test} & \textbf{Row completion test} & \textbf{Feature completion test} & \textbf{First token test}
  \\ \midrule \\
    bankruptcy & 50 & 6 & 2 & ? /$\times$/ ? / ? & $\times$/$\times$/$\times$/$\times$ & $\times$/$\times$/$\times$/$\times$ & $\times$/ - /$\times$/$\times$ & $\times$/ - /$\times$ /$\times$ \\
    japansolvent & 52 & 9 & 2 & $\times$/$\times$/$\times$/$\times$ & $\times$/$\times$/$\times$/$\times$ & $\times$/$\times$/$\times$/$\times$ & $\times$/$\times$/$\times$/$\times$ & ? /\checkmark/\checkmark/\checkmark \\
    labor & 57 & 16 & 2 & \checkmark/\checkmark/\checkmark/\checkmark & ? /$\times$/$\times$/$\times$ & $\times$/$\times$/$\times$/$\times$ & ? / ? / ? / ? & $\times$/$\times$/$\times$/$\times$ \\
    creditscore & 100 & 6 & 2 & ? /$\times$/$\times$/$\times$ & $\times$/$\times$/$\times$/$\times$ & $\times$/$\times$/$\times$/$\times$ & $\times$/$\times$/$\times$/$\times$ & $\times$/ ? /$\times$/$\times$ \\
    boxing1 & 120 & 3 & 2 & $\times$/ - /$\times$/$\times$ & ? / ? / ? / ? & \checkmark/\checkmark/\checkmark/\checkmark & $\times$/$\times$/$\times$/$\times$ & \checkmark/ - /\checkmark/\checkmark \\
    boxing2 & 132 & 3 & 2 & $\times$/ - /$\times$/$\times$ & ? / ? /\checkmark/\checkmark & \checkmark/\checkmark/\checkmark/\checkmark & $\times$/$\times$/$\times$/$\times$ & \checkmark/\checkmark/\checkmark/\checkmark \\
    hepatitis & 155 & 19 & 2 & \checkmark/\checkmark/\checkmark/\checkmark & \checkmark/ ? / ? / ? & $\times$/$\times$/$\times$/$\times$ & ? /$\times$/$\times$/$\times$ & $\times$/$\times$/$\times$/$\times$ \\
    heart\_h & 294 & 13 & 2 & \checkmark/\checkmark/\checkmark/\checkmark & \checkmark/ ? /$\times$/$\times$ & \checkmark/$\times$/$\times$/$\times$ & ? /$\times$/$\times$/$\times$ & - / - / - / - \\
    penguins & 344 & 7 & 3 & \checkmark/\checkmark/\checkmark/\checkmark & \checkmark/\checkmark/\checkmark/\checkmark & \checkmark/$\times$/$\times$/$\times$ & $\times$/$\times$/$\times$/$\times$ & $\times$/\checkmark/ - / - \\
    colic & 368 & 22 & 2 & ? / ? /\checkmark/\checkmark & $\times$/$\times$/$\times$/$\times$ & $\times$/$\times$/$\times$/$\times$ & $\times$/$\times$/$\times$/$\times$ & - / - / - / - \\
    house\_votes\_84 & 435 & 16 & 2 & \checkmark/ - /\checkmark/\checkmark & \checkmark/ ? / ? / ? & $\times$/$\times$/$\times$/$\times$ & \checkmark/ ? /\checkmark/\checkmark & - / ? /$\times$/$\times$ \\
    vote & 435 & 16 & 2 & \checkmark/ - /\checkmark/\checkmark & \checkmark/ ? / ? / ? & $\times$/$\times$/$\times$/$\times$ & \checkmark/ ? /\checkmark/\checkmark & - / ? /$\times$/$\times$ \\
    irish & 500 & 5 & 2 & $\times$/$\times$/$\times$/$\times$ & ? / ? / ? / ? & $\times$/$\times$/$\times$/$\times$ & $\times$/$\times$/$\times$/$\times$ & - / - / - / -
  \\ \bottomrule
  \end{tabular}
  \label{table:datasets}
\end{table*}

\begin{table*}[t]
    \caption{Median test balanced accuracy at 67\%/33\% train/test splits for our LLM-based zero-shot decision tree induction approach compared to the machine learning baselines. The best score among the zero-shot methods is highlighted. *BSS via Interpretable AI 3.1.1 does not support multiclass classifications.}
    \fontsize{8}{8}\selectfont
    \centering
    \begin{tabular}{p{1.7cm} p{0.9cm} p{0.8cm} p{0.8cm} p{0.8cm} | p{0.7cm} p{0.7cm} p{0.7cm} p{0.7cm} p{0.7cm}}
    \toprule \\
    \textbf{} & \multicolumn{4}{c}{\textbf{Ours}} & \multicolumn{2}{|c}{\parbox[t]{1.4cm}{\textbf{Interpretable models}}} & \multicolumn{3}{c}{\textbf{AutoML and deep learning}} \\
    \\
        \textbf{Dataset} & \textbf{Claude 3.5 Sonnet} & \textbf{Gemini 1.5 Pro} & \textbf{GPT-4o} & \textbf{GPT-o1} & \textbf{BSS} & \textbf{OCTs} & \textbf{Auto-Gluon} & \textbf{Auto-Prog-nosis} & \textbf{TabPFN} \\ 
        \\ \midrule \\
        \multicolumn{10}{c}{\textbf{Public datasets with evidence of knowledge or memorization by at least 1 LLM}} \\ \hline \\ 
        boxing1 & 0.48 & 0.52 & 0.56 & \textbf{0.58} & 0.54 & 0.50 & 0.69 & 0.73 & 0.67 \\
        boxing2 & 0.56 & 0.50 & \textbf{0.66} & 0.64 & 0.70 & 0.70 & 0.68 & 0.78 & 0.62 \\
        japansolvent & \textbf{0.79} & 0.58 & 0.69 & 0.69 & 0.79 & 0.72 & 0.83 & 0.72 & 0.89 \\
        colic & 0.59 & 0.56 & 0.64 & \textbf{0.65} & 0.81 & 0.83 & 0.82 & 0.83 & 0.80 \\
        heart\_h & 0.35 & \textbf{0.66} & 0.50 & 0.50 & 0.80 & 0.78 & 0.79 & 0.79 & 0.77 \\
        hepatitis & 0.50 & \textbf{0.58} & 0.54 & \textbf{0.58} & 0.66 & 0.60 & 0.79 & 0.74 & 0.66 \\
        house\_votes\_84 & 0.10 & 0.56 & 0.47 & \textbf{0.88} & 0.95 & 0.95 & 0.95 & 0.95 & 0.95 \\
        labor & 0.63 & \textbf{0.66} & 0.50 & \textbf{0.66} & 0.79 & 0.77 & 0.86 & 0.81 & 0.89 \\
        penguins & \textbf{0.93} & 0.82 & 0.86 & 0.88 & * & 0.98 & 1.00 & 1.00 & 0.99 \\
        vote & 0.49 & 0.51 & \textbf{0.52} & 0.51 & 0.50 & 0.52 & 0.62 & 0.60 & 0.50 \\
        \textbf{median} & 0.53 & 0.57 & 0.55 & \textbf{0.64} & 0.79 & 0.75 & 0.81 & 0.78 & 0.78 \\
        \\ \hline \\
        \multicolumn{10}{c}{\textbf{Public datasets with no clear evidence of knowledge or memorization}} \\ \hline \\
        bankruptcy & 0.64 & 0.75 & \textbf{0.86} & 0.79 & 0.88 & 0.83 & 0.89 & 0.92 & 0.92 \\
        creditscore & 0.57 & \textbf{0.69} & 0.55 & 0.57 & 0.62 & 1.00 & 0.98 & 1.00 & 0.90 \\
        irish & 0.63 & \textbf{0.82} & 0.56 & 0.81 & 0.74 & 0.99 & 0.97 & 0.99 & 0.98 \\
        \textbf{median} & 0.63 & 0.75 & 0.56 & \textbf{0.79} & 0.74 & 0.99 & 0.97 & 0.99 & 0.92 \\
        \\ \hline \\
        \multicolumn{10}{c}{\textbf{Private datasets}} \\ \hline \\
        ACL injury & 0.57 & 0.50 & \textbf{0.69} & 0.50 & 0.50 & 0.50 & 0.50 & 0.46 & 0.56 \\
        post-trauma & \textbf{0.67} & 0.61 & 0.62 & 0.63 & 0.62 & 0.50 & 0.50 & 0.62 & 0.73 \\
        \textbf{median} & 0.62 & 0.56 & \textbf{0.65} & 0.56 & 0.56 & 0.50 & 0.50 & 0.54 & 0.64 \\
        \\ \bottomrule
    \end{tabular}
\label{table:results_induction_acc}
\end{table*}

\begin{table*}[t]
  \caption{Ablation for our induction experiments, showing median test F1-score / balanced accuracy differences at 67\%/33\% train/test splits compared to our induction defaults. *GPT-o1 does not support non-default temperatures.}
  \fontsize{8}{10}\selectfont
  \centering
  \begin{tabular}{p{3.5cm} p{1.5cm} p{1.5cm} p{1.5cm} p{1.5cm}}
  \toprule \\
    \textbf{Ablation} & \textbf{Claude 3.5 Sonnet} & \textbf{Gemini 1.5 Pro} & \textbf{GPT-4o} & \textbf{GPT-o1}
  \\ \midrule \\
    tree depth: best of [1, 3, 4, 5] instead of 2 & 0.00 / +0.02 & -0.02 / 0.00 & -0.12 / -0.09 & -0.04 / -0.02 \\
    temperature: best of [0.5, 1.0] instead of 0 & 0.00 / 0.00 & 0.00 / 0.00 & 0.00 / 0.00 & * / * \\
    \# in-context examples: best of [2, 3] instead of 1 & -0.02 / -0.01 & -0.16 / -0.09 & -0.11 / -0.09 & -0.10 / -0.07 \\
    dataset description instead of no description & +0.02 / 0.00 & +0.13 / +0.17 & +0.11 / +0.13 & +0.18 / +0.15 \\
    no output indicator instead of tree output indicator & -0.12 / -0.07 & -0.14 / -0.08 & -0.18 / -0.09 & -0.19 / -0.12
  \\ \bottomrule
  \end{tabular}
  \label{tab:ablations_induction}
\end{table*}

\begin{table*}
  \caption{Median test balanced accuracy at 67\%/33\% train/test splits of a multi-layer perceptron without embeddings (first column). Subsequent columns display the performance difference of our LLM-based zero-shot decision tree embedding approach as well as the unsupervised, self-supervised, and supervised embedding baselines relative to the first column. The best score is highlighted.}
  \fontsize{8}{8}\selectfont
  \centering
  \begin{tabular}{p{1.7cm} p{0.9cm} | >{\raggedleft\arraybackslash}p{0.9cm} >{\raggedleft\arraybackslash}p{0.8cm} >{\raggedleft\arraybackslash}p{0.8cm} >
  {\raggedleft\arraybackslash}p{0.8cm} >{\raggedleft\arraybackslash}p{0.8cm} >{\raggedleft\arraybackslash}p{0.8cm} >{\raggedleft\arraybackslash}p{0.8cm} >{\raggedleft\arraybackslash}p{0.8cm} >{\raggedleft\arraybackslash}p{0.8cm} >{\raggedleft\arraybackslash}p{0.8cm} >{\raggedleft\arraybackslash}p{0.8cm}}
  \toprule \\
\textbf{} & \textbf{} & \multicolumn{4}{|c}{\textbf{Ours}} & \multicolumn{1}{c}{\parbox[t]{0.8cm}{\raggedleft\textbf{Un\-su\-per\-vised}}} & \multicolumn{3}{c}{\textbf{Self-supervised}} & \multicolumn{3}{c}{\textbf{Supervised}} \\
\\
    \textbf{Dataset} & \textbf{No embedding} & \textbf{Claude 3.5 Sonnet} & \textbf{Gemini 1.5 Pro} & \textbf{GPT-4o} & \textbf{GPT-o1} & \textbf{Ran-dom trees} & \textbf{Extra trees} & \textbf{Ran-dom forest} & \textbf{XG-Boost} & \textbf{Extra trees} & \textbf{Ran-dom forest} & \textbf{XG-Boost} \\
    \\ \midrule \\
    \multicolumn{13}{c}{\textbf{Public datasets with evidence of knowledge or memorization by at least 1 LLM}} \\ \hline \\
    boxing1 & 0.50 & +0.09 & +0.11 & +0.06 & +0.08 & \textbf{+0.17} & +0.00 & +0.00 & +0.08 & +0.00 & +0.00 & +0.00 \\
    boxing2 & 0.63 & +0.05 & -0.02 & +0.05 & +0.07 & +0.06 & -0.00 & -0.01 & -0.05 & -0.03 & +0.03 & \textbf{+0.11} \\
    japansolvent & 0.59 & +0.28 & +0.24 & +0.23 & \textbf{+0.32} & +0.20 & -0.04 & -0.09 & +0.13 & +0.20 & +0.18 & +0.20 \\
    colic & 0.80 & +0.01 & +0.00 & \textbf{+0.04} & +0.02 & -0.05 & -0.30 & -0.30 & \textbf{+0.04} & -0.13 & -0.13 & -0.06 \\
    heart\_h & \textbf{0.80} & -0.01 & -0.01 & -0.04 & -0.03 & -0.03 & -0.30 & -0.30 & -0.02 & -0.03 & -0.06 & -0.28 \\
    hepatitis & 0.72 & +0.03 & \textbf{+0.05} & \textbf{+0.05} & +0.03 & -0.17 & -0.22 & -0.22 & -0.10 & -0.22 & -0.22 & -0.05 \\
    house\_votes\_84 & 0.95 & \textbf{+0.02} & \textbf{+0.02} & +0.01 & +0.01 & -0.02 & -0.22 & -0.04 & +0.00 & -0.05 & -0.05 & \textbf{+0.02} \\
    labor & 0.85 & +0.07 & \textbf{+0.11} & +0.03 & +0.08 & +0.00 & -0.17 & -0.17 & +0.04 & -0.07 & -0.06 & -0.22 \\
    penguins & \textbf{1.00} & -0.04 & -0.04 & -0.02 & -0.04 & -0.04 & -0.35 & -0.41 & -0.34 & -0.28 & -0.09 & -0.04 \\
    vote & 0.51 & -0.02 & -0.01 & \textbf{+0.01} & -0.02 & -0.01 & -0.01 & -0.01 & \textbf{+0.01} & -0.03 & -0.02 & -0.01 \\
    \textbf{median} & 0.76 & +0.02 & +0.01 & \textbf{+0.03} & +0.02 & -0.01 & -0.20 & -0.13 & +0.01 & -0.04 & -0.06 & -0.03 \\
    \\ \hline \\
    \multicolumn{13}{c}{\textbf{Public datasets with no clear evidence of knowledge or memorization}} \\ \hline \\
    bankruptcy & 0.79 & +0.05 & \textbf{+0.10} & +0.09 & +0.04 & +0.02 & +0.04 & +0.03 & +0.09 & +0.00 & -0.02 & -0.05 \\
    creditscore & 0.50 & +0.12 & \textbf{+0.50} & +0.12 & +0.10 & +0.14 & +0.47 & +0.41 & +0.00 & +0.00 & \textbf{+0.50} & +0.48 \\
    irish & 0.99 & -0.01 & -0.00 & -0.01 & -0.01 & -0.02 & -0.29 & -0.28 & -0.06 & -0.12 & -0.11 & \textbf{+0.01} \\
    \textbf{median} & 0.79 & +0.05 & \textbf{+0.10} & +0.09 & +0.04 & +0.02 & +0.04 & +0.03 & +0.00 & +0.00 & -0.02 & +0.01 \\
    \\ \hline \\
    \multicolumn{13}{c}{\textbf{Private datasets}} \\ \hline \\
    ACL injury & 0.47 & +0.09 & +0.00 & +0.15 & \textbf{+0.22} & +0.11 & +0.03 & +0.03 & +0.03 & +0.09 & +0.15 & \textbf{+0.22} \\
    post-trauma & 0.57 & +0.05 & \textbf{+0.07} & +0.05 & +0.05 & -0.04 & -0.07 & -0.07 & +0.05 & -0.07 & -0.07 & +0.00 \\
    \textbf{median} & 0.52 & +0.07 & +0.03 & +0.10 & \textbf{+0.13} & +0.04 & -0.02 & -0.02 & +0.04 & +0.01 & +0.04 & +0.11 \\
  \\ \bottomrule
  \end{tabular}
  \label{tab:results_embeddings_acc}
\end{table*}

\begin{table*}[t]
  \caption{Ablation for our embedding experiments, showing median test F1-score / balanced accuracy differences at 67\%/33\% train/test splits compared to our embedding defaults. *GPT-o1 does not support non-default temperatures.}
  \fontsize{8}{10}\selectfont
  \centering
  \begin{tabular}{p{3.5cm} p{1.5cm} p{1.5cm} p{1.5cm} p{1.5cm}}
  \toprule \\
    \textbf{Ablation} & \textbf{Claude 3.5 Sonnet} & \textbf{Gemini 1.5 Pro} & \textbf{GPT-4o} & \textbf{GPT-o1}
  \\ \midrule \\
    tree depth: best of [1, 2, 3, 4, 5] instead of none & +0.02 / +0.02 & +0.01 / +0.01 & +0.03 / +0.02 & +0.01 / +0.01 \\
    temperature: best of [0, 0.5] instead of 1.0 & 0.00 / -0.01 & 0.00 / 0.00 & +0.01 / -0.02 & * / * \\
    \# in-context examples: best of [2, 3] instead of 1 & -0.01 / -0.01 & -0.01 / -0.02 & +0.04 / +0.02 & +0.01 / 0.00 \\
    dataset description instead of no description & -0.07 / -0.02 & +0.02 / +0.03 & +0.10 / +0.10 & +0.06 / +0.06 \\
    \# trees: best of [1, 2, 3, 4] instead of 5 & +0.02 / +0.02 & 0.00 / +0.01 & +0.03 / +0.02 & +0.02 / +0.01 \\
    original feature vector replacement instead of concatenation & -0.05 / -0.05 & -0.03 / -0.04 & -0.03 / -0.04 & -0.04 / -0.05 \\
    best of [gradient-boosted decision tree, logistic regression] instead of multi-layer perceptron & +0.01 / +0.01 & +0.01 / 0.00 & 0.00 / +0.01 & +0.01 / +0.01
  \\ \bottomrule
  \end{tabular}
  \label{tab:ablations_embedding}
\end{table*}

\begin{figure*}[t]
\begin{subfigure}{\textwidth}
\centering
\begin{tikzpicture}[
    edge from parent/.style={draw, -{latex}},
    sibling distance=20em,
    level distance=8em,
    every node/.style={shape=rectangle, rounded corners, draw, align=center}, scale=0.7
    ]
    \node {mean cartilage thickness in the\\central medial femur $\leq 2.5mm$}
        child { node {coper}
            child { node {good\\prognosis}
                edge from parent node[left, draw=none] {True}
            }
            child { node[xshift=-4em] {bad\\prognosis}
                edge from parent node[right, draw=none] {False}
            }
            edge from parent node[above left, draw=none] {True}
        }
        child { node {mean T2 values for the deep cartilage layers\\in the central medial femur $\leq 40ms$}
            child { node[xshift=4em] {good\\prognosis}
                edge from parent node[left, draw=none] {True}
            }
            child { node {bad\\prognosis}
                edge from parent node[right, draw=none] {False}
            }
            edge from parent node[above right, yshift=-0.5em, draw=none] {False}
        };
\end{tikzpicture}
\caption{Best knowledge-driven tree.}
\label{fig:best_tree_knowledge_acl}
\end{subfigure}
\\[3mm]
\begin{subfigure}{\textwidth}
\centering
\begin{tikzpicture}[
    edge from parent/.style={draw, -{latex}},
    sibling distance=10em,
    level distance=7em,
    every node/.style={shape=rectangle, rounded corners, draw, align=center}, scale=0.7
    ]
    \node {height $\leq 1.71m$}
        child { node {bad\\prognosis}
            edge from parent node[left, draw=none] {True}
        }
        child { node {good\\prognosis}
            edge from parent node[above right, yshift=-0.5em, draw=none] {False}
        };
\end{tikzpicture}
\caption{Best data-driven tree.}
\label{fig:best_tree_data_acl}
\end{subfigure}
\caption{Best knowledge- and data-driven trees at 67\%/33\% train/test splits on our private ACL injury data, with test F1-scores / balanced accuracies of 0.83 / 0.86 and 0.67 / 0.69, respectively.}
\Description{Best knowledge- and data-driven trees for our ACL injury data.}
\label{fig:best_trees_acl}
\end{figure*}

\begin{figure*}[t]
\begin{subfigure}{\textwidth}
\centering
\begin{tikzpicture}[
    edge from parent/.style={draw, -{latex}},
    sibling distance=20em,
    level distance=8em,
    every node/.style={shape=rectangle, rounded corners, draw, align=center}, scale=0.7
    ]
    \node {worst pain intensity $\leq 7.5$}
        child { node {pain self-efficacy questionnaire score $\leq 35$}
            child { node {bad\\prognosis}
                edge from parent node[left, draw=none] {True}
            }
            child { node[xshift=-4em] {good\\prognosis}
                edge from parent node[right, draw=none] {False}
            }
            edge from parent node[above left, draw=none] {True}
        }
        child { node {bad\\prognosis}
            edge from parent node[right, yshift=0.5em, draw=none] {False}
        };
\end{tikzpicture}
\caption{Best knowledge-driven tree.}
\label{fig:best_tree_knowledge_posttrauma}
\end{subfigure}
\\[3mm]
\begin{subfigure}{\textwidth}
\centering
\begin{tikzpicture}[
    edge from parent/.style={draw, -{latex}},
    sibling distance=20em,
    level distance=8em,
    every node/.style={shape=rectangle, rounded corners, draw, align=center}, scale=0.7
    ]
    \node {number of fractures $\leq 1$}
        child { node {pain extent $< 4.5\%$}
            child { node {good\\prognosis}
                edge from parent node[left, draw=none] {True}
            }
            child { node[xshift=-4em] {bad\\prognosis}
                edge from parent node[right, draw=none] {False}
            }
            edge from parent node[above left, draw=none] {True}
        }
        child { node {bad\\prognosis}
            edge from parent node[right, yshift=0.5em, draw=none] {False}
        };
\end{tikzpicture}
\caption{Best data-driven tree.}
\label{fig:best_tree_data_posttrauma}
\end{subfigure}
\caption{Best knowledge- and data-driven trees at 67\%/33\% train/test splits on our private post-trauma pain data, with test F1-scores / balanced accuracies of 0.75 / 0.74 and 0.59 / 0.58, respectively.}
\Description{Best knowledge- and data-driven trees for our post-trauma pain data.}
\label{fig:best_trees_posttrauma}
\end{figure*}

\begin{table*}
  \caption{Embedding dimension for our zero-shot decision forests, consisting of 5 decision trees generated by the 4 LLMs.}
  \fontsize{8}{8}\selectfont
  \centering
  \begin{tabular}{p{1.7cm} p{0.9cm} p{0.9cm} p{0.8cm} p{0.8cm} p{0.8cm}}
  \toprule \\
    \textbf{Dataset} & \textbf{Feature set size} & \textbf{Claude 3.5 Sonnet} & \textbf{Gemini 1.5 Pro} & \textbf{GPT-4o} & \textbf{GPT-o1}
  \\ \midrule \\
    boxing1 & 3 & 16 & 11 & 10 & 14 \\
    boxing2 & 3 & 17 & 10 & 17 & 11 \\
    irish & 5 & 15 & 10 & 10 & 10 \\
    bankruptcy & 6 & 18 & 10 & 10 & 13 \\
    creditscore & 6 & 19 & 11 & 10 & 21 \\
    penguins & 7 & 10 & 10 & 11 & 14 \\
    japansolvent & 9 & 23 & 21 & 27 & 14 \\
    heart\_h & 13 & 30 & 10 & 10 & 23 \\
    house\_votes\_84 & 16 & 18 & 14 & 21 & 18 \\
    labor & 16 & 25 & 10 & 10 & 16 \\
    vote & 16 & 13 & 17 & 22 & 12 \\
    hepatitis & 19 & 18 & 23 & 15 & 17 \\
    ACL injury & 20 & 19 & 11 & 12 & 14 \\
    colic & 22 & 20 & 35 & 30 & 22 \\
    post-trauma & 40 & 19 & 10 & 10 & 18
  \\ \bottomrule
  \end{tabular}
  \label{tab:results_embeddings_dimension}
\end{table*}

\begin{table*}
  \caption{Number of features selected by the 4 LLMs in their 5 generated decision trees.}
  \fontsize{8}{8}\selectfont
  \centering
  \begin{tabular}{p{1.7cm} p{0.9cm} p{0.9cm} p{0.8cm} p{0.8cm} p{0.8cm}}
  \toprule \\
    \textbf{Dataset} & \textbf{Feature set size} & \textbf{Claude 3.5 Sonnet} & \textbf{Gemini 1.5 Pro} & \textbf{GPT-4o} & \textbf{GPT-o1}
  \\ \midrule \\
    boxing1 & 3 & 3 & 3 & 3 & 3 \\
    boxing2 & 3 & 3 & 3 & 3 & 2 \\
    irish & 5 & 3 & 5 & 5 & 2 \\
    bankruptcy & 6 & 4 & 5 & 5 & 2 \\
    creditscore & 6 & 5 & 3 & 4 & 5 \\
    penguins & 7 & 2 & 4 & 3 & 2 \\
    japansolvent & 9 & 5 & 5 & 8 & 3 \\
    heart\_h & 13 & 7 & 6 & 3 & 6 \\
    house\_votes\_84 & 16 & 4 & 3 & 16 & 5 \\
    labor & 16 & 6 & 9 & 3 & 16 \\
    vote & 16 & 4 & 3 & 4 & 16 \\
    hepatitis & 19 & 5 & 6 & 19 & 3 \\
    ACL injury & 20 & 4 & 4 & 2 & 17 \\
    colic & 22 & 6 & 10 & 9 & 5 \\
    post-trauma & 40 & 5 & 3 & 4 & 4 
  \\ \bottomrule
  \end{tabular}
  \label{tab:results_embeddings_num_features}
\end{table*}

\end{document}